\documentclass[journal]{IEEEtran}

\usepackage{filecontents}
\usepackage{tikzscale}
\usepackage[utf8]{inputenc}
\usepackage{graphics}
\usepackage{graphicx}
\usepackage{tikz}
\usepackage{amsmath}
\DeclareMathOperator{\atantwo}{atan2}
\usepackage[export]{adjustbox}
\usepackage{color}

\usepackage{enumitem}
\usepackage{amsfonts}
\usepackage{subfig}
\usepackage{tabularx,booktabs}
\newcolumntype{C}{>{\centering\arraybackslash}X} 
\usepackage{acro}
\usepackage[percent,demo,abs]{overpic}
\usepackage[font=small]{caption}
\usepackage{hyperref}
\usepackage{tikz}
\usetikzlibrary{plotmarks}
\usepackage{pgfplots}
\usepgfplotslibrary{colorbrewer}
\usepackage{mathtools}
\usepackage{subfiles}
\usepackage{pgfplotstable}
\usepgfplotslibrary{statistics}
\pgfplotsset{compat=1.15}
\DeclarePairedDelimiter{\abs}{\lvert}{\rvert}
\DeclareAcronym{cnn}{short = CNN, long = Convolutional Neural Network}
\DeclareAcronym{gan}{short = GAN, long = Generative Adversarial Network}
\DeclareAcronym{cgan_long}{short = cGAN, long = Conditional Generative Adversarial Network}
\DeclareAcronym{cgan}{short = cGAN, long = conditional GAN}
\DeclareAcronym{ss}{short = SS, long = Semantic Segmentation}

\title{\LARGE \bf Empty Cities: a Dynamic-Object-Invariant Space for Visual SLAM}
\author{Berta Bescos, Cesar Cadena, Jose Neira}

\makeatletter
 \let\old@ps@headings\ps@headings
 \let\old@ps@IEEEtitlepagestyle\ps@IEEEtitlepagestyle
 \def\confheader#1{%
 \def\ps@headings{%
 \old@ps@headings%
 \def\@oddhead{\textit{#1}\hfill\strut}%
 \def\@evenhead{\textit{#1}\hfill\strut}%
 }%
 \def\ps@IEEEtitlepagestyle{%
 \old@ps@IEEEtitlepagestyle%
 \def\@oddhead{\textit{#1}\hfill\strut}%
 \def\@evenhead{\textit{#1}\hfill\strut}%
 }%
 \ps@headings%
 }
 \makeatother

\confheader{IEEE Transactions on Robotics}

\begin{document}

{
\onecolumn
\begin{center}
\fontsize{12pt}{12pt}\selectfont
\vspace*{1cm}
\center{\rm  This paper has been accepted for publication in IEEE Transactions on Robotics.}
\vspace*{0.7cm}
\center{\rm  DOI: 10.1109/TRO.2020.3031267}
\vspace*{0.7cm}
\center{\textcopyright 2020 IEEE. Personal use of this material is permitted. Permission from IEEE must be obtained for all other uses, in any
current or future media, including reprinting /republishing this material for advertising or promotional purposes, creating new
collective works, for resale or redistribution to servers or lists, or reuse of any copyrighted component of this work in other
works.}
\end{center}
}

\twocolumn
\newpage

\maketitle

\begin{abstract}
In this paper we present a data-driven approach to obtain the \emph{static} image of a scene, eliminating dynamic objects that might have been present at the time of traversing the scene with a camera.  The general objective is to improve vision-based localization and mapping tasks in dynamic environments, where the presence (or absence) of different dynamic objects in different moments makes these tasks less robust. 
We introduce an end-to-end deep learning framework to turn images of an urban environment that include dynamic content, such as vehicles or pedestrians, into realistic static frames suitable for localization and mapping. 
This objective faces two main challenges: detecting the dynamic objects, and inpainting the static occluded background. 
The first challenge is addressed by the use of a convolutional network that learns a multi-class semantic segmentation of the image. 
The second challenge is approached with a generative adversarial model that, taking as input the original dynamic image and the computed dynamic/static binary mask, is capable of generating the final static image. 
This framework makes use of two new losses, one based on image steganalysis techniques, useful to improve the inpainting quality, and another one based on ORB features, designed to enhance feature matching between real and hallucinated image regions.

To validate our approach, we perform an extensive evaluation on different tasks that are affected by dynamic entities, \textit{i.e.,} {\color{black} visual odometry}, place recognition and multi-view stereo, with the hallucinated images.
Code has been made available on \href{https://github.com/bertabescos/EmptyCities}{{\tt \small https://github.com/bertabescos/EmptyCities\_SLAM}}.
\end{abstract}

\section{Introduction}

Most vision-based localization systems are conceived to work in static environments~\cite{murTRO2015, engel2018direct, forster2014svo}. 
They can deal with small fractions of dynamic content, but tend to compute dynamic objects motion as camera ego-motion. 
Thus, their performance is compromised. 
Building stable maps is also of key importance for long-term autonomy. 
Mapping dynamic objects prevents vision-based robotic systems from recognizing already visited places and reusing pre-computed maps. 

\begin{figure} [t!]
\centering
\subfloat[\label{fig:teaser_input} Input of our system: urban images with dynamic content.]{\includegraphics[width=0.49\linewidth]{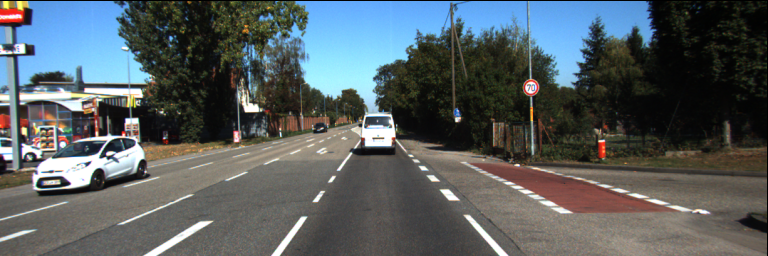}
\hspace{\fill}
\includegraphics[width=0.49\linewidth]{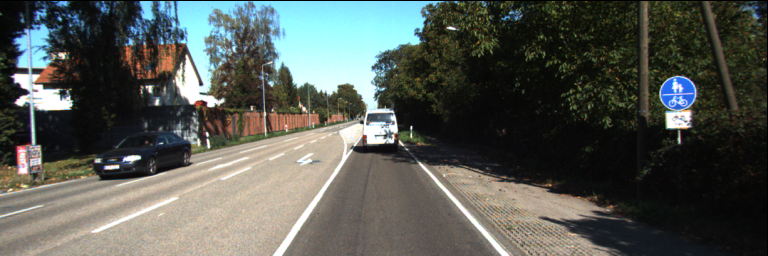}}
\\
\subfloat[\label{fig:teaser_output} Output of our system: dynamic objects have been removed.]{\includegraphics[width=0.49\linewidth]{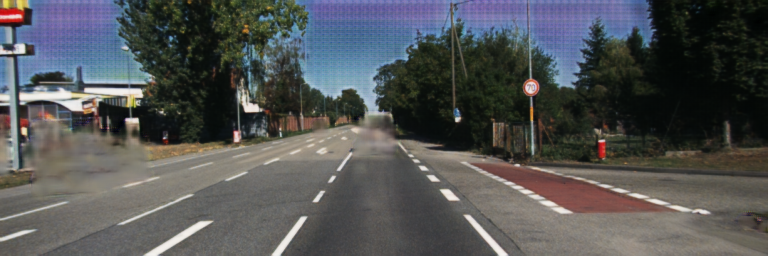}
\hspace{\fill}
\includegraphics[width=0.49\linewidth]{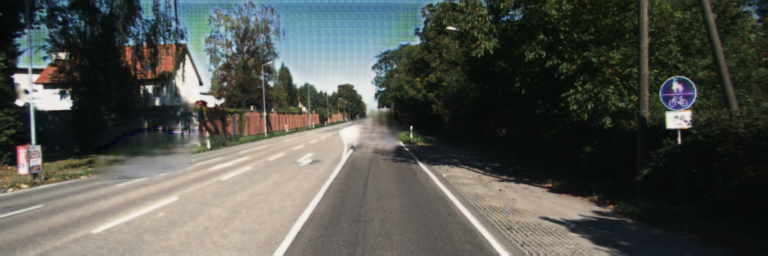}}
\\
\vspace{-0.3cm}
\subfloat[\label{fig:teaser_map} Static map built with the images pre-processed by our framework.]{
\includegraphics[width=1.0\linewidth]{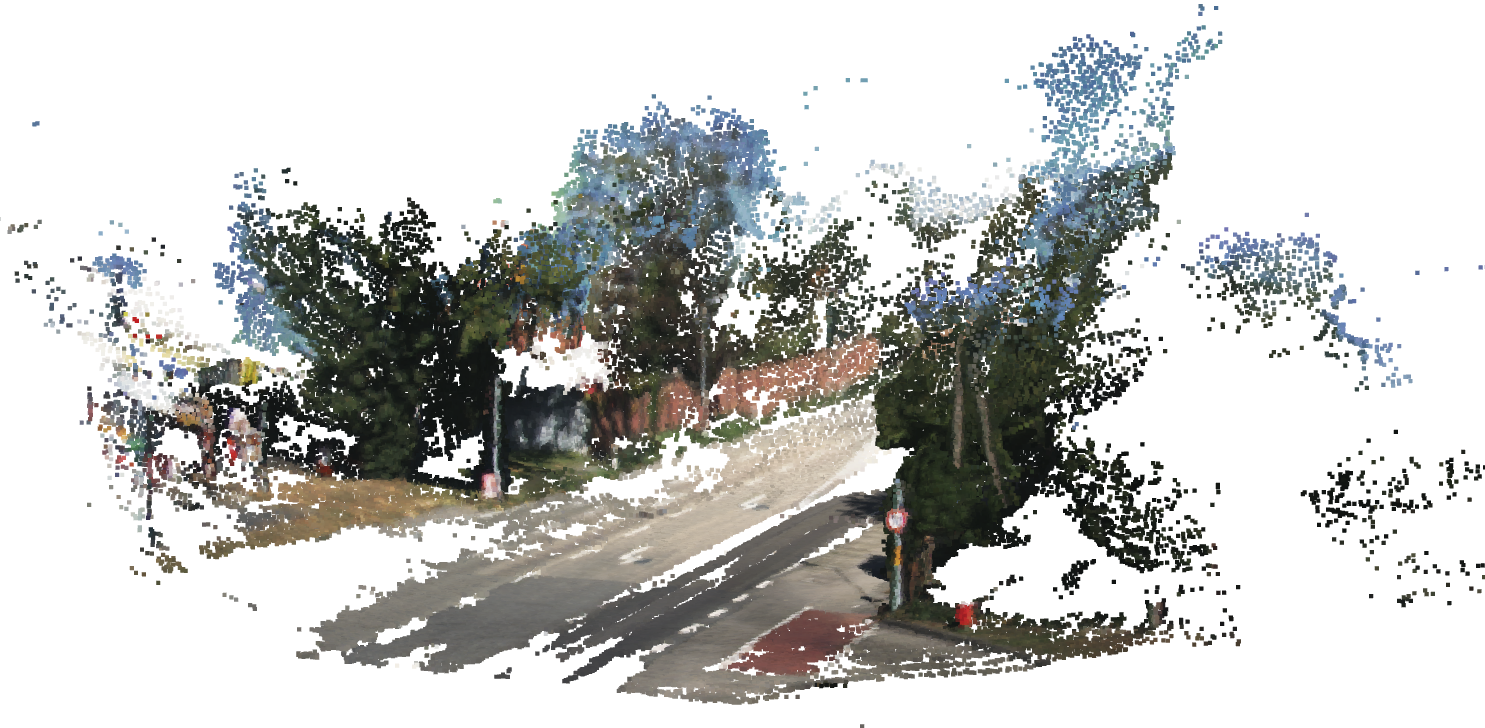}}
\caption{\label{fig:teaser} The dynamic images are firstly converted one by one into static with an end-to-end deep learning model. 
Such images allow us to compute an accurate camera trajectory estimation that is not damaged by the dynamic objects' motion, as well as to build dense static maps that are useful for long-term applications.}
\end{figure}

To deal with dynamic objects, some approaches include in their model the behavior of the observed dynamic content~\cite{agudo2015sequential, lamarca2019defslam}. 
Such strategy is needed when the majority of the observed scene is not rigid.
However, when scenes are mainly rigid, as in Fig.~\ref{fig:teaser_input}, the standard strategy consists of detecting the dynamic objects within the images and not to use them for localization and mapping~\cite{bescos2018dynaslam, alcantarilla2012combining, wang2014motion, tan2013robust}.
To address mainly rigid scenes, we propose to instead modify these images so that dynamic content is eliminated and the scene is converted realistically into static.
We consider that the combination of experience and context allows to hallucinate, \textit{i.e.}, inpaint, a geometrically and semantically consistent appearance of the rigid and static structure behind dynamic objects. 
This hallucinated structure can be used by the SLAM system to provide it with robustness against dynamic objects.

Turning images that contain dynamic objects into realistic static frames reveals several challenges:

\begin{enumerate}
\item{Detecting such dynamic content in the image. By this, we mean to detect not only those objects that are known to move such as vehicles, people and animals, but also the shadows and reflections that they might generate, since they also change the image appearance.}
\item{Inpainting the resulting space left by the detected dynamic content with plausible imagery. The resulting image would succeed in being realistic if the inpainted areas are both semantically and geometrically consistent with the static content of the image.}
\end{enumerate}

The first challenge can be addressed with geometrical approaches if an image sequence is available. 
This procedure usually consists in studying the optical flow consistency along the images~\cite{wang2014motion, alcantarilla2012combining}. 
In the case in which only one frame is available, deep learning is the approach that excels at this task by the use of \acp{cnn}~\cite{he2017mask,romera2018erfnet}. 
These frameworks are trained with the previous knowledge of what classes are dynamic and which ones are not. 
Recent works show that it is possible to acquire this knowledge in a self-supervised way~\cite{barnes2017driven, zhou2018dynamic}.

Regarding the second challenge, some recent image inpainting approaches use image statistics  of the remaining  image to fill in the holes~\cite{telea2004image, bertalmio2001navier}. 
The former work estimates the pixel value with the normalized weighted sum of all the known pixels in the neighbourhood. 
While this approach generally produces smooth results, it is limited by the available image statistics and has no concept of visual semantics. 
Neural networks learn semantic priors and meaningful hidden representations in an end-to-end fashion, which have been used for recent image inpainting efforts~\cite{liu2018image,yu2018generative,iizuka2017globally,pathak2016context}. 
These networks employ convolutional filters on images, replacing the removed content with inpainted areas that have geometrical and semantic consistency with the whole image.    

Both challenges can also be seen as one single task: translating a dynamic image into a corresponding static image. 
In this direction, Isola \textit{et~al.}~\cite{isola2017image} propose a general-purpose solution for image-to-image translation. 
Our previous work~\cite{bescos2019empty} builds on top of this idea and re-formulates the framework objectives to take advantage of a pre-computed dynamic object mask, seeking a more inpainting-oriented framework.
In this work, we follow this idea of transforming images with dynamic content into realistic static frames, while optimizing for localization and mapping performance. 
For such task, we introduce a new loss that, combined with the integration of a semantic segmentation network achieves the final objective of creating a dynamic-object-invariant space. 
This loss is based on steganalysis techniques and on ORB features detection, orientation and descriptor maps~\cite{rublee2011orb}. 
Such loss allows the inpainted images to be realistic and suitable for localization and mapping. 
These images provide a richer understanding of the stationary scene, and could also be of interest for the creation of high-detail road maps (Fig.~\ref{fig:teaser}), as well as for street-view imagery suppliers as a privacy measure to replace faces and license plates blurring.
We provide an extensive evaluation on robotic applications such as {\color{black}visual odometry}, place recognition, and mapping to prove the validity of our framework.

\section{Related Work}

\subsection{Dynamic Objects Detection}

The vast majority of SLAM systems assume a static environment.  
As a consequence, they can only manage small fractions of dynamic content by classifying them as spurious data or outliers to such static model. 
The most typical outlier rejection algorithms are RANSAC (\textit{e.g.}, in ORB-SLAM \cite{murTRO2015, mur2017orb}) and robust cost functions (\textit{e.g.}, in PTAM \cite{klein2007parallel}).

There are several SLAM systems that address more specifically the dynamic scene content. 
Tan \textit{et al.} \cite{tan2013robust} detect changes that take place in the scene by projecting the map features into the current frame for appearance and structure validation. 
Wang and Huang \cite{wang2014motion} segment the dynamic objects in the scene using RGB optical flow. 
Alcantarilla \textit{et al.} \cite{alcantarilla2012combining} detect moving objects by means of a scene flow representation with stereo cameras. 
More recently, thanks to the boost of deep learning, integrating semantics information into SLAM has allowed to deal with dynamic content in a different manner~\cite{runz2018maskfusion, bescos2018dynaslam}. 
This idea allows the clustering of map points belonging to independent objects with different dynamics, as well as the possibility of detecting dynamic objects in just one shot.

\subsection{Sequence-based Inpainting}

Previous works on SLAM in dynamic scenes have attempted to reconstruct the background occluded by dynamic objects in the images with information from previous frames~\cite{bescos2018dynaslam, scona2018staticfusion}. 
Such works need per-pixel depth information and only make use of the static content of the pre-built map to create the inpainted frames, but do not add semantic consistency. 
The work by Granados \textit{et~al.}~\cite{granados2012background} removes marked dynamic objects from videos by aligning other candidate frames in which parts of the missing region are visible, assuming that the scene can be approximated using piecewise planar geometry. 
The recent work by Uittenbogaard \textit{et~al.}~\cite{uittenbogaard2018moving} utilizes a \ac{gan} to learn to use information from different viewpoints and select imagery information from those views to generate a plausible inpainting which is similar to the ground-truth static background.
Eventually, if only one frame is available, the static occluded background can only be reconstructed by utilizing image-based inpainting techniques.

\subsection{Image-based Inpainting}

Among the non-learning approaches to image inpainting, propagating appearance information from neighboring pixels to the target region is the usual procedure~\cite{telea2004image}.
Accordingly, these methods succeed in dealing with narrow holes, where color and texture vary smoothly, but fail when handling big holes, resulting in over-smoothing. 
Differently, patch-based methods iteratively search for relevant patches from the rest of the image~\cite{efros2001image}. 
These approaches are computationally expensive and hence not fast enough for real-time applications. 
Yet, they do not make semantically-aware patch selections.

Deep learning based methods usually initialize the image holes with a constant value, and further pass it through a \ac{cnn}. 
Context Encoders \cite{pathak2016context} were among the first ones to successfully use a standard pixel-wise reconstruction loss, as well as an adversarial loss for image inpainting tasks. 
Due to the resulting artifacts, Yang \textit{et~al.}~\cite{yang2017high} take their results as input and then propagate the texture information from non-hole regions to fill the hole regions as post-processing.
Song \textit{et~al.}~\cite{song2017image} use a refinement network in which a blurry initial hole-filling result is used as the input, then iteratively replaced with patches from the closest non-hole regions in the feature space. 
Iizuka \textit{et~al.}~\cite{iizuka2017globally} extend Content Encoders by defining global and local discriminators, then apply a post-processing. 
Following this work, Yu \textit{et~al.}~\cite{yu2018generative} replaced the post-processing with a refinement network powered by the contextual attention layers. 
The recent work of Liu \textit{et~al.}~\cite{liu2018image} obtains excellent results by using partial convolutions. 

In contrast, the work by Ulyanov \textit{et~al.}~\cite{ulyanov2017deep} proves that there is no need for external dataset training. 
The generative network itself can rely on its structure to complete the corrupted image. 
However, this approach usually applies several iterations ($\sim$50000) to get good and detailed results. 

\subsection{Image Inpainting for a Dynamic-Object-Invariant Space}

This work builds on our previous work Empty Cities~\cite{bescos2019empty}, which bins the image sequences and treats the frames independently. 
It makes use of deep learning to segment out the \textit{a priori} moving objects: vehicles, animals and pedestrians, and also of image-based ``inpainting''. 
It does not perform pure inpainting but image-to-image translation with the help of a dynamic objects' mask, outcome of a semantic segmentation network. 
This choice is justified by the fact that the dynamic objects' mask might be inaccurate or may not include their shadows.  The adoption of an image-to-image translation framework allows  to slightly modify the image non-hole regions for better accommodation of the reconstructed areas. 
Differently to inpainting methods, the ``holes'' cannot be initialized with any placeholder values because we do not want the framework to only modify those values and hence, our inpainting network input consists of the dynamic original image with the dynamic/static mask concatenated. 
Concisely, utilizing an image-to-image translation approach allows us to have the image hole regions inpainted, and the non-hole regions slightly modified for better accommodation of the reconstructed areas to cope with imprecise masks, or with dynamic objects possible shadows and reflections.

\section{Image-to-Image Translation}

Our work makes use of the successful image-to-image translation framework by Isola \textit{et~al.}~\cite{isola2017image}. 
For the sake of completeness, we summarize the basis of their approach.

\begin{figure}
    \centering
    \includegraphics[width=1.0\linewidth]{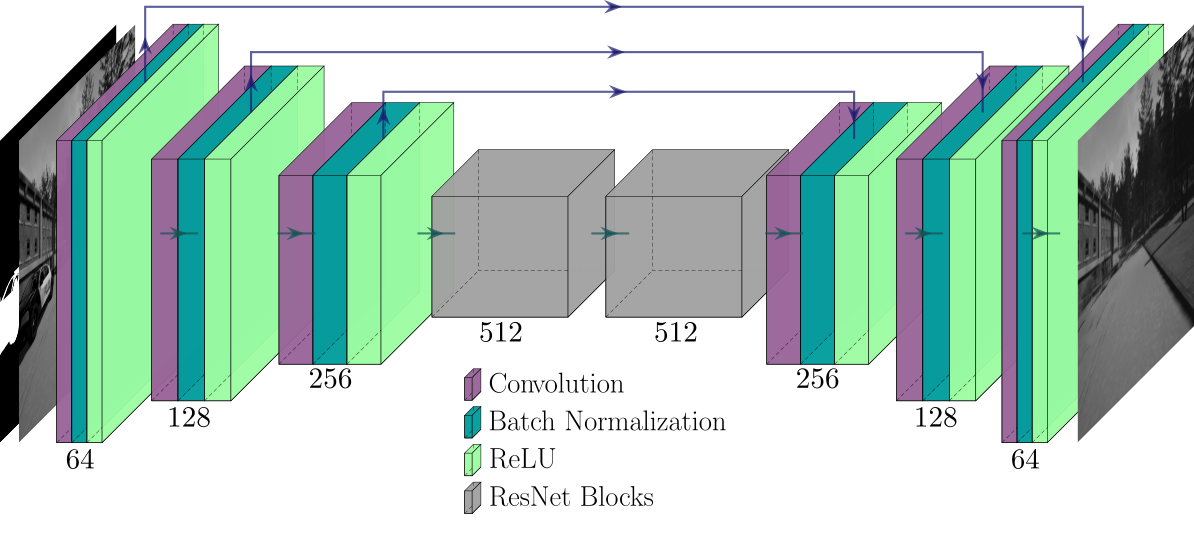}
    \caption{Our generator $G$ adopts a UResNet-like architecture. It employs 3 down-convolutional layers with a stride of 2, 6 ResNet blocks and 3 up-convolutional layers with a fractional stride of 1/2, with skip connections between corresponding down- and up-convolutional layers.
    Only 2 ResNet blocks are shown for simplicity.}
    \label{fig:architecture}
\end{figure}

A \ac{gan} is a generative model that learns a mapping from a random noise vector~$z$ to an output image~$y$, $G$: $z \rightarrow y$ \cite{goodfellow2014generative}. 
In contrast, a \ac{cgan} learns a mapping from observed image $x$ and optional random noise vector $z$, to $y$, \mbox{$G$ : $ \lbrace x,z \rbrace \rightarrow y $}~\cite{gauthier2014conditional}, or \mbox{$G$ : $ x \rightarrow y $}~\cite{isola2017image}. 
The generator $G$ is trained to produce outputs indistinguishable from the ``real'' images by an adversarially trained discriminator, $D$, which is trained to do as well as possible at detecting the generator's ``fakes''. 
The objective of a \ac{cgan} can be expressed as 
\begin{multline}\label{eq:cGAN}
\mathcal{L}_{cGAN}(G,D)  = \mathbb{E}_{x,y}[\log{D(x,y)}] + \\
\mathbb{E}_{x}[\log{(1-D(x,G(x)))}],
\end{multline}
where $G$ tries to minimize this objective against an adversarial $D$ that tries to maximize it. 
Previous approaches have found it beneficial to mix the \ac{gan} objective with a more traditional appearance loss, such as the $L1$ or $L2$ distance \cite{pathak2016context}. 
The discriminator's job remains unchanged, but the generator is tasked not only with fooling the discriminator, but also with being near the ground-truth in a $L1$ sense, as expressed in 
\begin{equation} \label{eq:L1}
G^*  = \text{arg}\,\min\limits_{G} \max\limits_{D}  \mathcal{L}_{cGAN}(G,D) + \lambda_{1} \cdot \mathcal{L}_{L1}(G),
\end{equation}
where $\mathcal{L}_{L1}(G)  = \mathbb{E}_{x,y}[||y-G(x)||_1]$. 
The recent work of Isola \textit{et~al.}~\cite{isola2017image} shows that \ac{cgan}s are suitable for image-to-image translation tasks, where the output image is conditioned on its corresponding input image, \textit{i.e.}, it translates an image from one space into another (semantic labels to RGB appearance, RGB appearance to drawings, day to night, \textit{etc.}). 
The realism of their results is also enhanced by their generator architecture. 
They employ a U-Net~\cite{ronneberger2015u}, which allows low-level information to shortcut across the network. 
In our previous work~\cite{bescos2019empty} we made use of this same architecture with $256 \times 256$ resolution images. 
However, visual localization systems see their accuracy degraded when working with low-resolution images. 
For this objective, we hereby employ as the architecture for our generator $G$ a UResNet~\cite{guerrero2018white}, see~Fig.~\ref{fig:architecture}. 
This architecture uses residual blocks~\cite{hinton2006reducing} and has shown impressive results for super-resolution images~\cite{kingma2013auto}.

It is well known that $L2$ and $L1$ losses produce blurry results on image generation problems, \textit{i.e.}, they can capture the low frequencies but fail to encourage high frequency crispness. 
This motivates restricting the \ac{gan} discriminator to only model high frequency structures. 
Following this idea, Isola \textit{et~al.}~\cite{isola2017image} adopt a discriminator architecture that classifies each $N \times N$ patch in an image as real or fake, rather than classifying the image as a whole. 
Due to their excellent results, we adopt this same architecture for our discriminator.

\section{Our Proposal}
\label{sec:system}

\begin{figure*}
    \centering
    \includegraphics[width=0.9\linewidth]{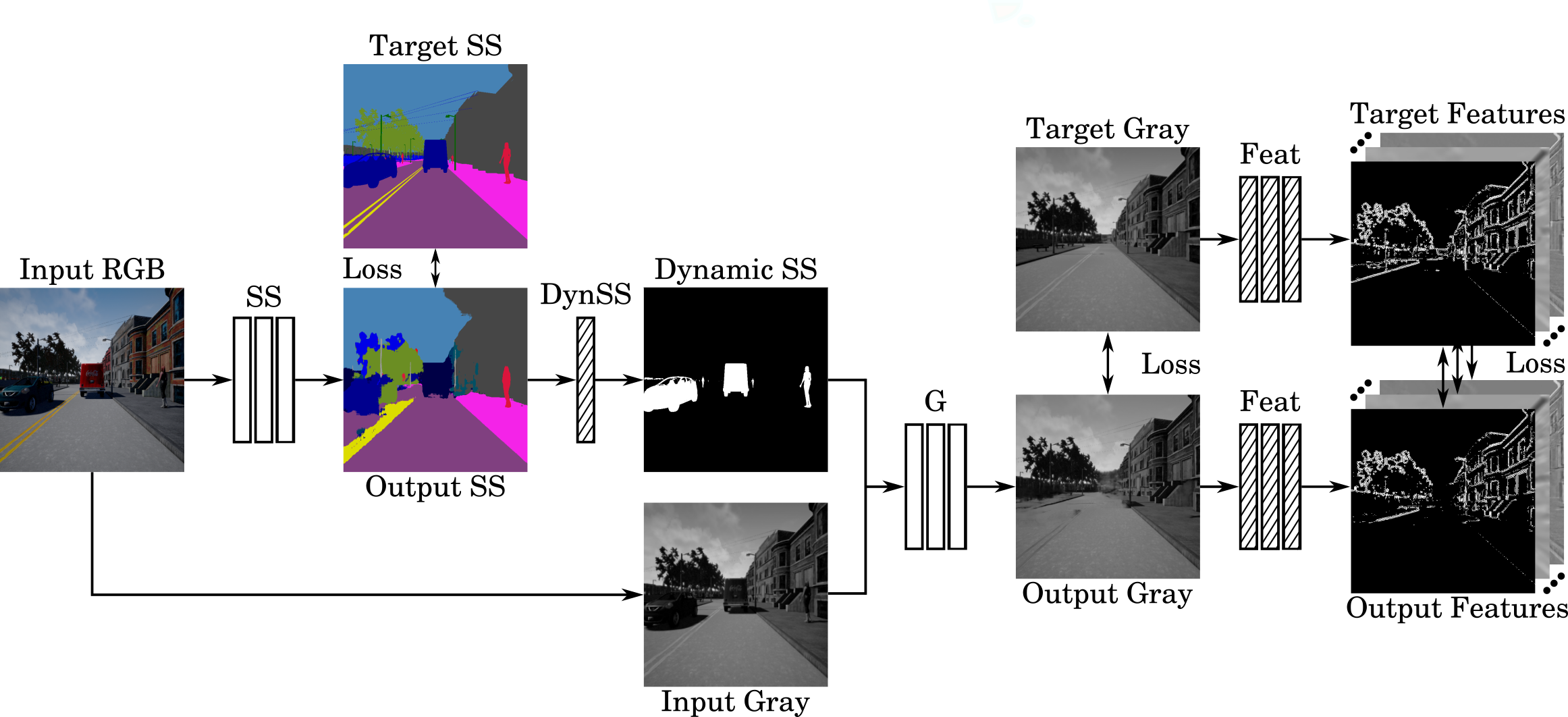}
    \caption{Block diagram of our proposal. 
    We first compute the segmentation of the RGB dynamic image, as well as its loss against its ground-truth. 
    Both the dynamic/static binary mask and dynamic image are used to obtain the static image. 
    A loss based on ORB features together with an appearance and an adversarial loss are obtained and back-propagated until the RGB dynamic image. 
    The striped blocks are differentiable layers that are fixed and hence not modified during training time.
    The adversarial discriminator is not shown here for simplicity.}
    \label{fig:Pipeline}
\end{figure*}

Our proposed system turns images of an urban environment that show dynamic content, such as vehicles or pedestrians, into realistic static frames which are suitable for localization and mapping. 
We first obtain the pixel-wise semantic segmentation of the RGB dynamic image (see Fig.~\ref{fig:Pipeline}). 
Then, the segmentation of only the dynamic objects is obtained with the convolutional network $DynSS$. 
Once we have this mask, we convert the RGB dynamic image to gray scale and we compute the static image, also in gray scale, with the use of the generator $G$, which has been trained in an adversarial way. 
For simplicity, the discriminator is not shown in this diagram. 
To fully exploit the capabilities of this framework for localization and mapping, inpainting is enriched with a loss based on ORB features detection, orientation and descriptors between the ground-truth and computed static images.
Another feature of our framework for localization and mapping is the fact that we perform the inpainting in gray-scale rather than in RGB. 
The motivation for this is that many visual localization applications only need the images grayscale information. 
The different stages are described in subsections \ref{subsec:image2image} to \ref{subsec:dynSS}.

\subsection{From Image-to-Image Translation to Inpainting}
\label{subsec:image2image}

For our objective, dynamic objects masks are specially considered to re-formulate the training objectives of the general purpose image-to-image translation work by Isola \textit{et~al.}~\cite{isola2017image}. 
We adopt a variant of the \ac{cgan} that learns a mapping from observed image $x$ and dynamic/static binary mask $m$, to $y$, \mbox{$G$ : $ \lbrace x,m \rbrace \rightarrow y $}. 
Also, the discriminator $D$ learns to classify $\hat{y} = G(x,m)$ patches as ``fake'' from $\hat{y}$, $m$ and $x$, and the patches of $y$ as ``real'' from $y$, $m$ and $x$, \mbox{$D$ : $ \lbrace x,y/\hat{y},m \rbrace \rightarrow real/fake $}.

In most of the training dataset images, the relationship between the static and dynamic regions sizes is unbalanced, \textit{i.e.}, static regions occupy usually a much bigger area. 
This leads us to believe that the influence of dynamic regions on the final loss is significantly reduced.
As a solution to this problem, we propose to re-formulate the $cGAN$ and $L1$ losses so that there is more emphasis on the main areas that have to be inpainted, according to Eqs.~\ref{eq:wL1} and \ref{eq:wcGAN}. 
The weights $w$ are computed as $w = \frac{N}{N_{dyn}} \text{ if } m=1$ (dynamic object), and as $w = \frac{N}{N - N_{dyn}} \text{ if } m=0$ (background). $N$ stands for the number of elements in the binary mask $m$, and $N_{dyn}$ means the number of pixels where $m=1$.
\begin{equation} \label{eq:wL1}
\mathcal{L}_{L1}(G)  = \mathbb{E}_{x,y}[w \cdot \abs{\abs{y - G(x,m)}}_1],
\end{equation}
\begin{multline}\label{eq:wcGAN}
\mathcal{L}_{cGAN}(G,D)  = \mathbb{E}_{x,y}[w \cdot \log{D(x,y,m)}] + \\
\mathbb{E}_{x}[w \cdot \log{(1-D(x,G(x,m),m))}],
\end{multline}

\begin{figure}
    \centering
    \subfloat[\label{fig:i}]{\includegraphics[width=.20\linewidth]{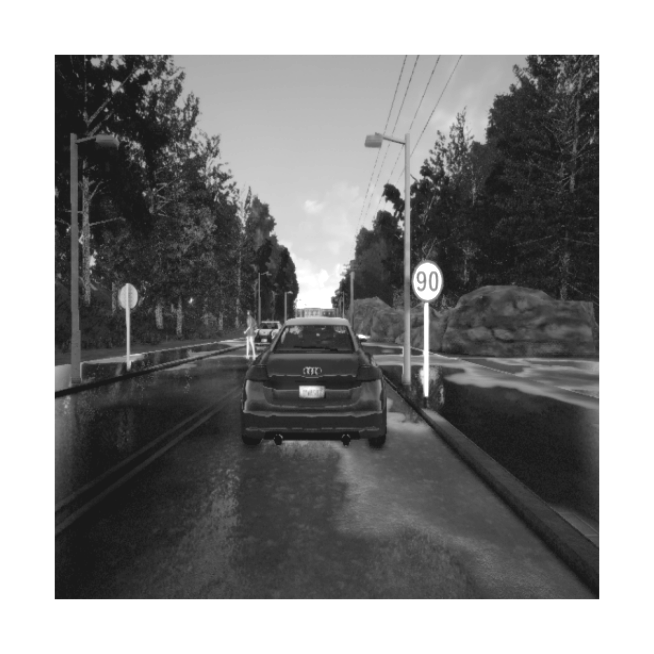}}
    \hfill
    \subfloat[\label{fig:o}]{\includegraphics[width=.20\linewidth]{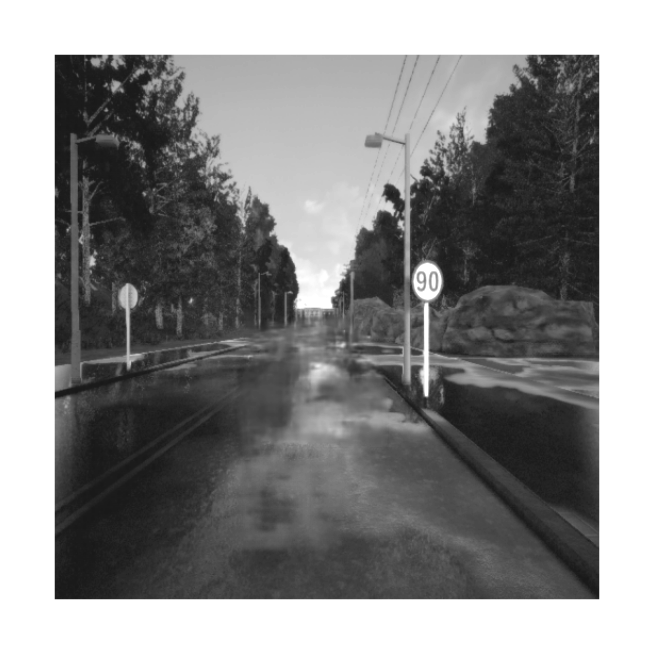}}
    \hfill
    \subfloat[\label{fig:on}]{\includegraphics[width=.20\linewidth]{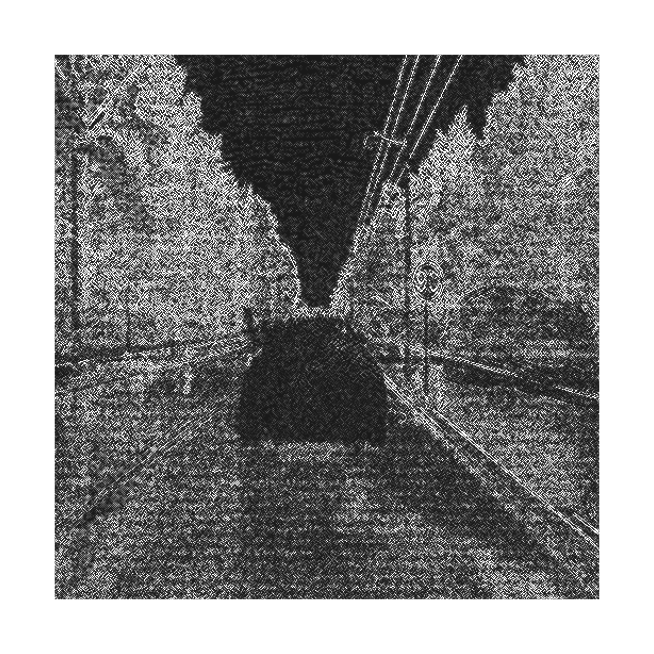}}
    \hfill
    \subfloat[\label{fig:t}]{\includegraphics[width=.20\linewidth]{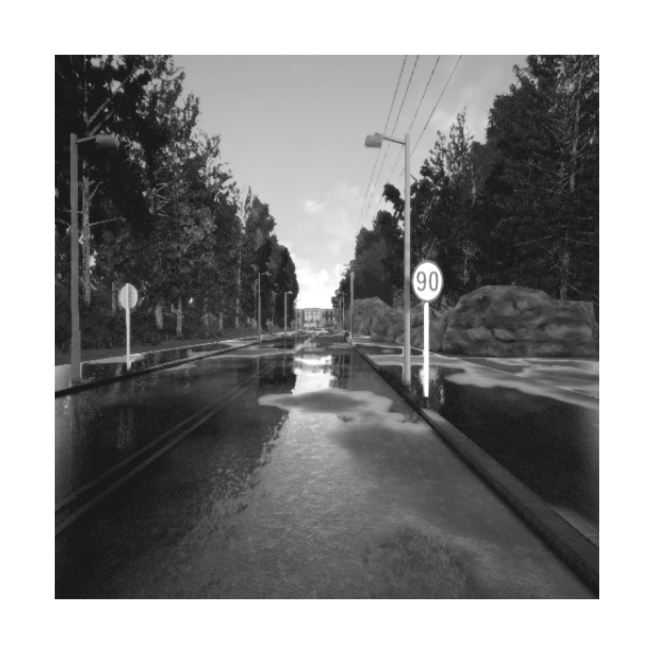}}
    \hfill
    \subfloat[\label{fig:tn}]{\includegraphics[width=.20\linewidth]{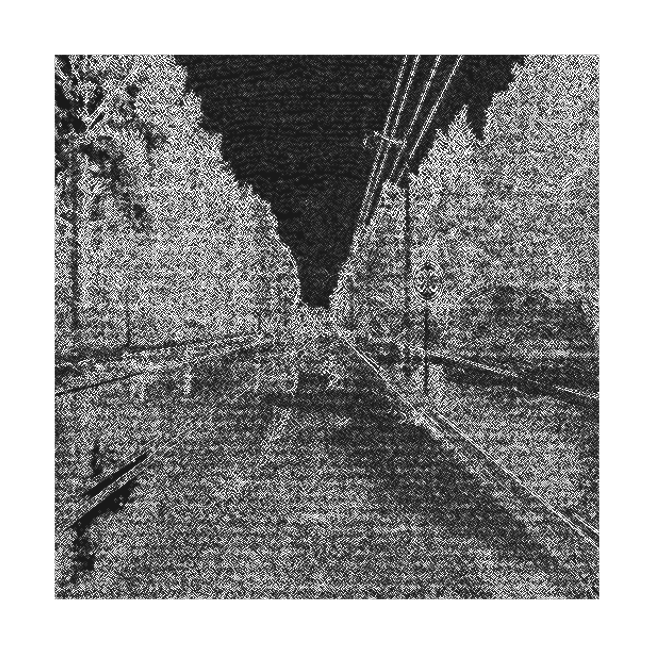}}
    \caption{\protect\subref{fig:o} shows the image generated by our framework when taking \protect\subref{fig:i} as input. 
    \protect\subref{fig:t} is the static objective image. 
    \protect\subref{fig:o} and \protect\subref{fig:t} are visually similar, but their computed noise (\protect\subref{fig:on} and \protect\subref{fig:tn} respectively) clearly show what image and what parts of it have been modified the most. 
    The noise magnitude has been amplified ($\times$10) for visualization.}
    \label{fig:noise}
\end{figure}

\begin{figure*} [h]
\centering
\subfloat{\includegraphics[height=.19\linewidth]{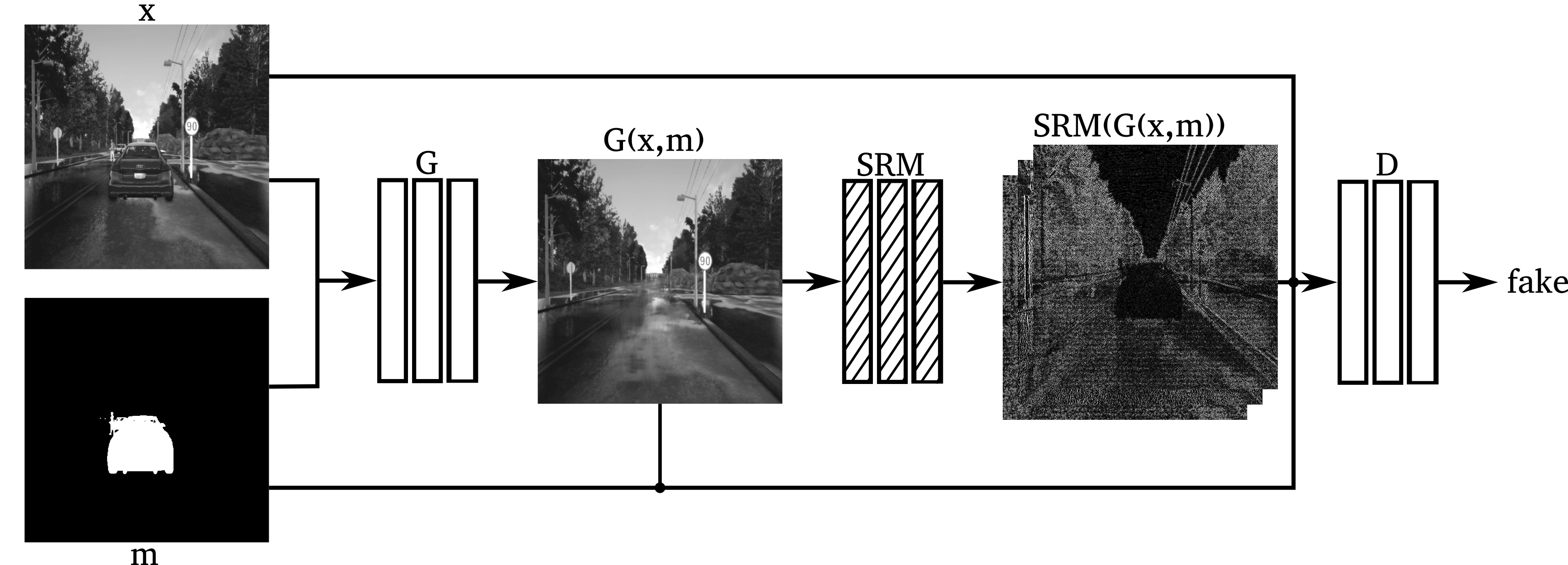}}
\hfill
\centering
\subfloat{\includegraphics[height=.19\linewidth]{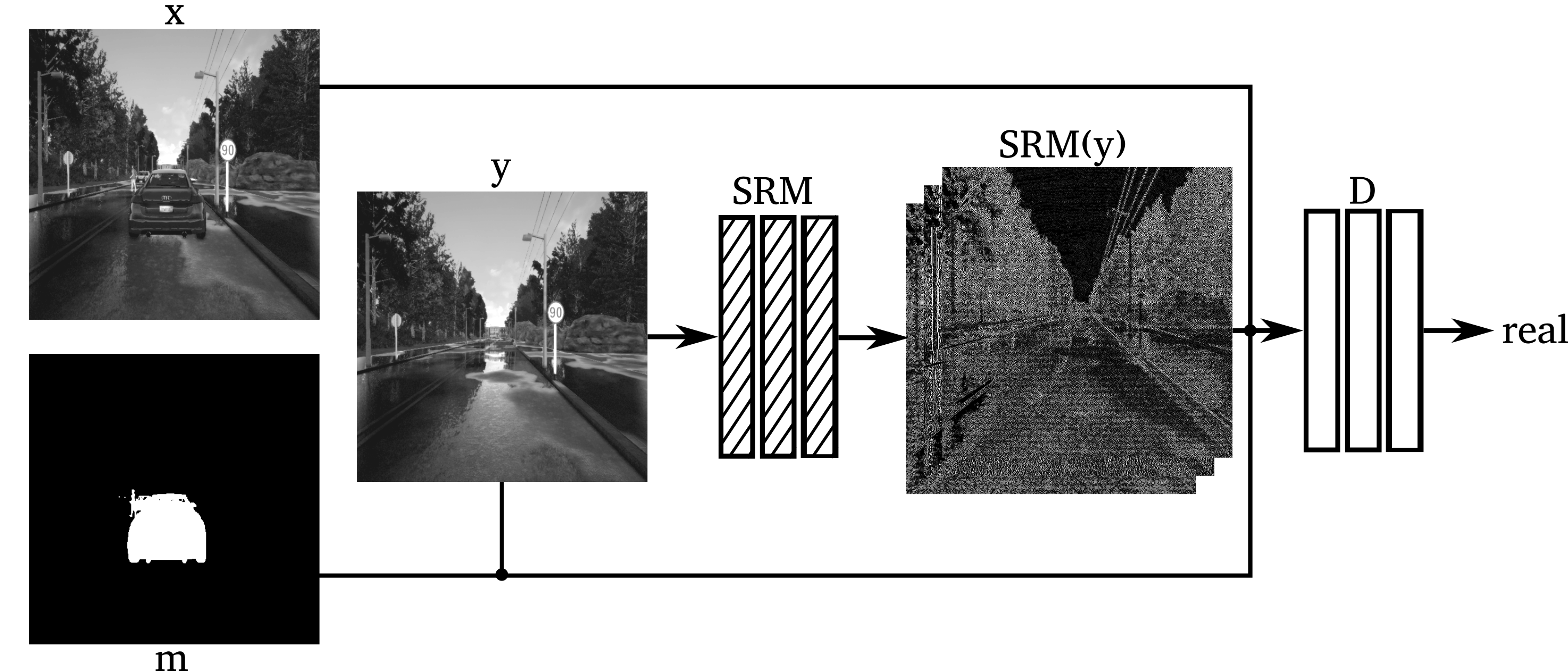}}
\caption{\label{fig:Disc} The discriminator $D$ has to learn to differ between the real images $y$ and the images generated by the generator $G(x,m)$. 
$D$ makes a better decision (\textit{real} / \textit{fake}) by seeing the inputs of the generator $x$ and $m$, and by seeing the SRM noise features of $G(x,m)$ and $y$.
The striped blocks are convolutional layers whose weights do not require upgrading during training time.}
\end{figure*}

An important feature that we have also incorporated to the framework is the computation of our output and target images ``noise''. 
This is motivated by the use of the noise domain for steganalysis to detect if an image has been tampered or not.
Fig.~\ref{fig:noise} shows an example of why working in the noise domain is helpful for detecting ``fake'' images. 
While the static generated image (Fig.~\ref{fig:o}) looks visually similar to its target image (Fig.~\ref{fig:t}), their computed noises (Figs.~\ref{fig:on} and \ref{fig:tn}) are very different. 
It would be very easy for us, humans, to tell what parts of the original image (Fig.~\ref{fig:i}) have been changed by analyzing their noise mapping. 
In the same way, the discriminator could more easily learn to distinguish ``real'' from ``fake'' images if it can take as input their noise.
This idea is explained more in depth in subsection~\ref{subsec:steganalysis}, and the whole training procedure is diagrammed in Fig.~\ref{fig:Disc}. 
To the best of our knowledge, steganalysis noise features have not been used before in the context of \ac{gan}s.

This \ac{gan} training setup leads to having good inpainting results. 
However, despite the efforts of the discriminator to catch the high frequency of the ``real'' images, the outputs of our framework are still slightly blurry. 
One of the objectives of this work is to use our images for localization tasks, therefore, if the inpainted regions are somewhat blurry, features would not be extracted in these areas. 
Image features are important for localization since many visual SLAM systems rely on them as their core (ORB-SLAM~\cite{mur2017orb}).
Having blurriness in inpainted areas could be seen as a good feature of our framework for navigation because it would allow feature-based localization systems to work with our images without any modification in their architecture, and ``fake'' features would not be introduced. 
This would be equivalent to modifying the utilized localization system to work with the raw images and the dynamic/static binary masks. 
We have proved with our localization experiments (subsection~\ref{subsec:localization}) that not utilizing moving objects' features leads to worse tracking results than working with fully static images. 
For that reason, we want to exploit our framework for obtaining both high-quality inpainting results, and to succeeding in generating reliable features for visual localization tasks. 
Fortunately, these two assignments are highly related. 
Solving one of them leads to having the other one tackled. 
Therefore, we have implemented a new loss based on ORB features~\cite{rublee2011orb}. 
That is, we want the output of our generator $G$ to have the same ORB features than its target image, while keeping it realistic and close to its target in a $L1$ sense. 
By the same ORB features we mean the same detected keypoints with their same orientation and descriptors, following ORB's implementation to the extent possible. 
This procedure is further described in subsection~\ref{subsec:ORB}.

\subsection{Steganalysis-Based Loss}
\label{subsec:steganalysis}
With the advances of image editing techniques, tampered or manipulated image generation processes have become widely available. %
As a result, distinguishing authentic images from tampered images has become increasingly challenging. 

What our framework is actually trying to achieve is to eliminate certain regions from an authentic image followed by inpainting, \textit{i.e.}, \textit{removal}, one of the most common image manipulation techniques. 
It is the discriminator's job to classify the generated image patches as tampered (fake) or real. 

Images have a low-frequency component dependent on their content, and a high-frequency component dependent on their source camera. 
These high-frequency components are known as noise features or noise residuals, and can be extracted using linear and non-linear high-pass filters.
Recent works on image forensics utilize noise features~\cite{fridrich2012rich, zhou2018learning} as clues to classify a specific patch or pixel in an image as tampered or not, and localize the tampered regions.
The intuition behind this idea is that when an object is removed from one image (source) and the gap is inpainted (target), the noise features between the source and target are unlikely to match. 

\begin{figure} [h]
\centering
\subfloat{\includegraphics[height=.19\linewidth]{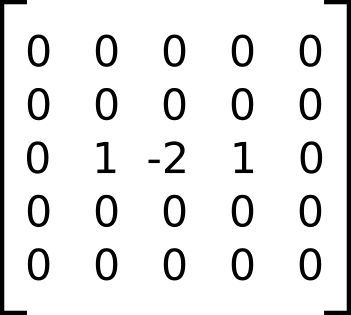}}
\hspace{0.3cm}
\subfloat{\includegraphics[height=.19\linewidth]{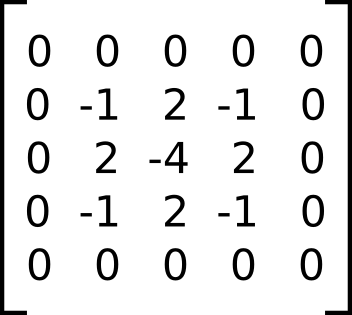}}
\hspace{0.3cm}
\subfloat{\includegraphics[height=.19\linewidth]{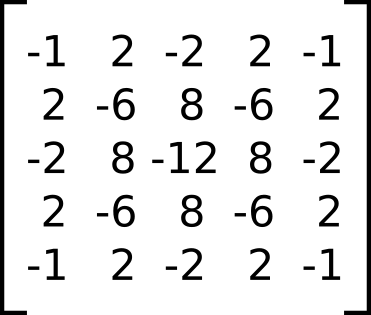}}
\caption{\label{fig:kernels} The three utilized SRM kernels to extract noise features. The left kernel is useful in regions with a strong gradient. The middle and the rightmost kernels  provide the layer with a high shift-invariance.}
\end{figure}

To provide the discriminator with better clues to distinguish real from fake inputs, we first extract the noise features from our images and concatenate them to the gray-scale images, as depicted in Fig.~\ref{fig:Disc}. 
The \ac{cgan} objective is re-formulated as
\begin{multline}\label{eq:noiseGAN}
\mathcal{L}_{cGAN}(G,D)  = \mathbb{E}_{x,y}[w \cdot \log{D(x,y,m,n)}] + \\
\mathbb{E}_{x}[w \cdot \log{(1-D(x,\hat{y},m,\hat{n}))}],
\end{multline}
where $n=SRM(y)$ and $\hat{n}=SRM(\hat{y})$. 
There are many ways to produce noise features from an image.
Inspired by recent progress on Steganalysis Rich Models (SRM) for image manipulation detection~\cite{fridrich2012rich}, we use SRM filter kernels to extract the local noise features from the static images as the input to our discriminator.
The SRM use statistics of neighboring noise residual samples as features to capture the dependency changes caused by embedding. 
Zhou \textit{et~al.}~\cite{zhou2018learning} use SRM residuals, together with the RGB image to detect and localize corrupted regions in images. 
They only use three SRM kernels, instead of thirty (as in the original Fridrich \textit{et~al.}'s work~\cite{fridrich2012rich}), and claim that they achieve comparable performance. 
Similarly, we use these same three filters (Fig.~\ref{fig:kernels}), setting the kernel size of the SRM filter layer to be $5 \times 5 \times 3$.

\subsection{ORB-Features-Based Loss}
\label{subsec:ORB}
ORB features allow real-time detection and description, and provide good invariance to changes in viewpoint and illumination. 
Furthermore, they are useful for visual SLAM and place recognition, as demonstrated in the popular ORB-SLAM~\cite{mur2017orb} and its binary bag-of-words~\cite{galvez2012bags}. 
The following sections summarize how the ORB features detector, descriptors and orientation are computed, and how we have adapted them into a new loss.

\subsubsection{Detector}
The ORB Detector is based on the FAST algorithm~\cite{rosten2006machine}. 
It takes one parameter, the intensity threshold $t$ between the center pixel $p$, $I_p$, and those in a circular ring around the center. 
If there exists a set of contiguous pixels in the circle which are all brighter than $I_p + t$, or all darker than $I_p -t$, the pixel $p$ would be a keypoint candidate.
Then the Harris corner measure is computed for each of these candidates, and the target $N$ keypoints with the highest Harris measure are finally selected. 
FAST does not produce multi-scale features, therefore, ORB uses a scale pyramid of the image and extracts FAST features at each level in the pyramid. 

\begin{figure}
    \centering
    \subfloat{\includegraphics[width=.12\linewidth]{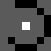}}
    \hfill
    \subfloat{\includegraphics[width=.12\linewidth]{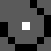}}
    \hfill
    \subfloat{\includegraphics[width=.12\linewidth]{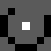}}
    \hfill
    \subfloat{\includegraphics[width=.12\linewidth]{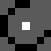}}
    \hfill
    \subfloat{\includegraphics[width=.12\linewidth]{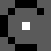}}
    \hfill
    \subfloat{\includegraphics[width=.12\linewidth]{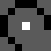}}
    \hfill
    \subfloat{\includegraphics[width=.12\linewidth]{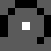}}
    \hfill
    \subfloat{\includegraphics[width=.12\linewidth]{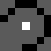}}
    \caption{A subset of the 16 kernels used to obtain corner responses in the images. 
    The $12$ black pixels have a value of $-1/12$, the gray pixels are set to $0$, and the white pixel is set to $1$. 
    A very positive or a very negative response will be obtained when convolving these kernels with a corner area in an image.
    }
    \label{fig:corners}
\end{figure}

To bring this to a differentiable solution, we have defined a convolution capable of detecting corners in an image in the same way that FAST does. 
We have approximated the FAST corner detection and have used instead a convolution with the kernels in Fig.~\ref{fig:corners}. 
These images show some of the kernels used for corners detection for a circular ring of 3 pixels around the center. 
By  convolving  the  image with these kernels for  different kernel sizes, we obtain its corner response for the different image pyramid levels. 
We keep the maximum score per pixel and per level and raise each element to its 2nd power to equally leverage positive and negative responses. 
We then subtract a value which is equivalent to the FAST threshold~$t$, and apply a sigmoid operation. 
Its output is the probability of a pixel of being a FAST feature, and could also be seen as the Harris corner measure.
Features for the output and target images are computed following this procedure. 
We define this network as $det$, and the corresponding loss $\mathcal{L}_{det}(G)$ can be expressed as
\begin{multline} \label{eq:det}
\mathcal{L}_{det}(G)  = -\mathbb{E}_{x,y}[w_{det} \cdot (det(y) \cdot \log(det(\hat{y})) + \\ 
(1 - det(y)) \cdot \log(1 - det(\hat{y})) )],
\end{multline}
where $\hat{y} = G(x,m)$ and $w_{det}$ is calculated following Eq.~\ref{eq:w_det}. 
This weights definition allows us to leverage the uneven distribution of non-features and features pixels, and to affect only those image regions with a wrong feature response. 
$N$ stands for the number of pixels in the features map, and $N_f$ represents the number of pixels in the response map $det(y)$ where \mbox{$det(y) > 0.5$}, \textit{i.e.}, the number of FAST features in the current objective frame.

\begin{equation}\label{eq:w_det}
    w_{det} = \left\{
     \begin{array}{@{}l@{\thinspace}l}
      \frac{N}{N_f}, & det(y) > 0.5 \quad \text{and} \quad det(\hat{y}) \leq 0.5\\
       \frac{N}{N - N_f}, & det(y) \leq 0.5 \quad \text{and} \quad det(\hat{y}) > 0.5\\
       0, & \text{otherwise}\\
     \end{array}
   \right.
\end{equation}

According to our results, the optimum number of image pyramid levels for this objective is 1. 
More levels lead to a greater training time and the results are barely influenced. 
This is coherent with the idea that we want to maximize the sharpness of small features rather that of the big corners. 
These  convolutions have been applied with a stride of 5, offering a good trade-off between computational training time and good-quality results.

%
%

Other approaches have tried before to include a similar loss inside a CycleGAN framework~\cite{porav2018adversarial}. 
The work by Porav \textit{et~al.} uses the SURF detector~\cite{bay2008speeded}, which is already differentiable, but does not compute a binary loss. 
They compute a more traditional L1 loss between the blob responses of both output and ground-truth images. 
Computing a binary loss as in Eq.~\ref{eq:w_det} allows us to have more emphasis on the high-gradient areas. 

\subsubsection{Orientation}
Once FAST features have been detected, the original ORB work extracts their orientation to provide them with rotation invariance. 
This is done by computing its orientation $\theta = \atantwo(m_{01},m_{10})$ and its intensity centroid $C = (\frac{m_{10}}{m_{00}}, \frac{m_{01}}{m_{00}})$, where $m_{pq} = \sum_{x,y}x^py^qI(x,y)$ are the moments of an image patch. 
More precisely, the three utilized patch moments are \mbox{$m_{10} = \sum_{x,y}x \cdot I(x,y)$}, \mbox{$m_{01} = \sum_{x,y}y \cdot I(x,y)$} and \mbox{$m_{00} = \sum_{x,y}I(x,y)$}.
We have created three 14-pixel-radius circular kernels with the values $x$, $y$ and $1$ respectively for $m_{10}$, $m_{01}$ and $m_{00}$ (centered in 0), so that when convolving the image with them, we obtain their respective patch moments $m_{01}$, $m_{10}$ and $m_{00}$.
We define this network as $ori$, and the objective of its corresponding loss is that the ``fake'' static image detected features, $det(G(x,m))$, have the same orientation parameters $m_{01}$, $m_{10}$ and $m_{00}$ than the ground-truth static image detected features, $det(y)$.
This loss can be expressed as

\begin{equation} \label{eq:ori}
\mathcal{L}_{ori}(G)  = -\mathbb{E}_{x,y}[w_{ori} \cdot \abs{\abs{ori(y) - ori(\hat{y})}}_1].
\end{equation}

Even though these convolutions are applied to the whole image with a stride of 5 as in the detection loss, the weighting term $w_{ori}$ in Eq.~\ref{eq:ori} has a value of $1$ if a FAST feature has been detected in either the ground-truth static image or the output image, \textit{i.e.}, if $det(y) > 0.5 \quad \text{or} \quad det(\hat{y}) > 0.5$.
Otherwise, the weighting term $w_{ori}$ is set to $0$.

\subsubsection{Descriptor}

The ORB descriptor is a bit string description of an image patch constructed from a set of binary intensity tests. 
Consider a smoothed image patch, $\textbf{p}$, a binary test $\tau$ is defined by
\begin{equation}\label{eq:binary_test}
    \tau(\textbf{p};x,y) = \left\{
     \begin{array}{@{}l@{\thinspace}l}
       1, & \quad \textbf{p}(x) < \textbf{p}(y)\\
       0, & \quad \textbf{p}(x) \geq \textbf{p}(y)
     \end{array}
   \right.
\end{equation}
where $\textbf{p}(x)$ is the intensity of $\textbf{p}$ at a point $x$. 
The feature is defined as a vector of $n$ binary tests
\begin{equation}\label{eq:descriptor}
    f_n(\textbf{p}) = \sum_{1\leq i \leq n}2^{i-1}\tau(\textbf{p};x_i,y_i)
\end{equation}

As in Rublee \textit{et~al.}'s work~\cite{rublee2011orb}, we use a Gaussian distribution around the center of the patch and a vector length $n$~=~256. 
This can be achieved in a differentiable and convolutional manner by creating $n$ kernels with all values set to 0 except for those in the positions $x$ and $y$:
\begin{equation} \label{eq:desc}
    \textbf{k}(z) = \left\{
    \begin{array}{@{}l@{\thinspace}l}
        1, & \quad z = x\\
        -1, & \quad z = y\\
        0, & \quad \text{otherwise}\\
    \end{array}
    \right.
\end{equation}
where $\textbf{k}(z)$ is the value of the kernel $\textbf{k}$ at a point $z$. 
Convolving an image with these $n$ kernels yields each pixel's ORB descriptor (a negative output corresponds to the bit value 0 and a positive one to 1). 
This convolution is followed by a sigmoid activation function.
We define this network as $desc$, and the corresponding loss $\mathcal{L}_{desc}(G)$ can be expressed as
\begin{multline} \label{eq:desc}
\mathcal{L}_{desc}(G)  = -\mathbb{E}_{x,y}[w_{desc} \cdot (desc(y) \cdot \log(desc(\hat{y})) +\\
(1 - desc(y)) \cdot \log(1 - desc(\hat{y})) )],
\end{multline}
where the weights $w_{desc}$ are defined in Eq.~\ref{eq:w_desc}. 
This descriptors loss is back-propagated to the whole image, whether a feature has been detected or not, as it helps keeping the image statistics.

\begin{equation} \label{eq:w_desc}
    w_{desc} = \left\{
     \begin{array}{@{}l@{\thinspace}l}
      1, & \enskip desc(y) > 0.5 \enskip \& \enskip desc(\hat{y}) \leq 0.5\\
       1, & \enskip desc(y) \leq 0.5 \enskip \& \enskip desc(\hat{y}) > 0.5\\
       0, & \enskip \text{otherwise}\\
     \end{array}
   \right.
\end{equation}

All these losses are combined into one loss $\mathcal{L}_{ORB}(G)$, that is computed as in Eq.~\ref{eq:feat}. The values for the weights of the different losses $\lambda _{det}$, $\lambda _{ori}$ and $\lambda _{desc}$ have been chosen empirically, and they are set to $10$, $0.1$ and $1$ respectively.

\begin{equation} \label{eq:feat}
    \mathcal{L}_{ORB}(G) = \lambda _{det} \mathcal{L}_{det}(G) + \lambda_{ori} \mathcal{L}_{ori}(G) + \lambda_{desc} \mathcal{L}_{desc}(G)
\end{equation}

The features detection, orientation and descriptor maps can be computed in a parallel way to decrease the training time, since their computation is not necessarily sequential.

Finally, the generator's job can be expressed as in Eq.~\ref{eq:GFeat}.
\begin{equation} \label{eq:GFeat}
G^*  = \text{arg}\,\min\limits_{G} \max\limits_{D}  \mathcal{L}_{cGAN}(G,D) + \lambda_{1} \cdot \mathcal{L}_{L1}(G) + \mathcal{L}_{ORB}(G)
\end{equation}

As an implementation detail, we have first trained the whole system without the ORB loss for 125 epochs, and then have fine-tuned including it for another 25 epochs.

\subsection{Semantic Segmentation}
Semantic Segmentation is a challenging task that addresses most of the perception needs of intelligent vehicles in a unified way. Deep neural networks excel at this task, as they can be trained end-to-end to accurately classify multiple object categories in an image at pixel level. However, very few architectures have a good trade-off between high quality and computational resources. The recent work of Romera \textit{et~al.}~\cite{romera2018erfnet} runs in real time while providing accurate semantic segmentation. The core of their architecture (ERFNet) uses residual connections and factorized convolutions to remain efficient while retaining remarkable accuracy.

Romera \textit{et~al.}~\cite{romera2018erfnet} have made public some of their trained models \cite{erfnet2017er}. 
{\color{black}As in our preliminary work}, we use for our approach the ERFNet model with encoder and decoder both trained from scratch on Cityscapes train set \cite{cordts2016cityscapes}. 
We have fine tuned their model to adjust it to our inpainting approach by back-propagating the loss of the semantic segmentation $\mathcal{L}_{CE}(SS)$, calculated with the cross entropy criterion using the class weights they suggest, $w_{SS}$, and the adversarial loss of our final inpainting model $\mathcal{L}_{cGAN}(G,D)$. 
The semantic segmentation network's job (SS) can be hence expressed as:
\begin{equation}
\label{eq:LossMask}
SS^{*} = \text{arg}\,\min\limits_{SS} \max\limits_{D}
\mathcal{L}_{cGAN}(G,D) + \lambda_2 \cdot \mathcal{L}_{CE}(SS),
\end{equation}
where $\mathcal{L}_{CE}(SS) = w_{SS}[class] \cdot (log(\sum_{j} \exp(y_{SS}[j])) - y_{SS}[class])$. 
Its objective is to produce an accurate semantic segmentation $y_{SS}$, but also to fool the discriminator $D$.
The latter objective might occasionally lead the network to not only recognize dynamic objects but also their shadows.

\subsection{Dynamic Objects Semantic Segmentation}
\label{subsec:dynSS}
Once the semantic segmentation of the RGB image is done, we can select those classes known to be dynamic (vehicles and pedestrians). 
This has been done by applying a \textit{SoftMax} layer, followed by a convolutional layer with a kernel of $n \times 1 \times 1$, where $n$ is the number of classes, and with the weights of those dynamic and static channels set to $w_{dyn}$ and $w_{stat}$ respectively. 
%
With $w_{dyn} = \frac{n-n_{dyn}}{n}$ and $w_{stat} = -\frac{n_{dyn}}{n}$, where $n_{dyn}$ stands for the number of dynamic existing classes. 
A positive output corresponds to a dynamic object, whereas a negative one corresponds to a static one.
The resulting output passes through a hyperbolic tangent-type activation function to obtain the desired dynamic/static mask. 
Note that the defined weights $w_{dyn}$ and $w_{stat}$ are not changed during training time.

{\color{black}This segmentation stage has been adopted from our preliminary work~\cite{bescos2019empty} without suffering any new modifications.}

\section{Image-Based Experiments}
\label{sec:image}
\subsection{Data Generation}
\label{subsec:data}
We have analyzed the performance of our method using CARLA~\cite{Dosovitskiy17}. 
CARLA is an open-source simulator for autonomous driving research, that provides open digital assets (urban layouts, buildings, vehicles, pedestrians, \textit{etc.}). 
The simulation platform supports flexible specification of sensor suites and environmental conditions. 
We have generated over 12000 image pairs consisting of a target image captured with neither vehicles nor pedestrians, and a corresponding input image captured at the same pose with the same illumination conditions, but with cars, tracks and people moving around. 
These images have been recorded using a front and a rear RGB camera mounted on a car. 
Their ground-truth semantic segmentation has also been captured. 
CARLA offers two different towns that we have used for training and testing, respectively. 
Our dataset, together with more information about our framework, is available on \href{https://bertabescos.github.io/EmptyCities/}{{\tt \small https://bertabescos.github.io/EmptyCities\_SLAM/}}.

At present, we are limited to training this framework on synthetic datasets since, to our knowledge, no real-world dataset exists that provides RGB images captured under same illumination conditions at identical poses, with and without dynamic objects. 
In order to render our framework trained on synthetic data transferable to real-world data, we have fine-tuned our models with data from the Cityscapes and KITTI semantic segmentation training datasets~\cite{cordts2016cityscapes, geiger2013vision}. 
These datasets are semantically similar to the ones synthesized with CARLA. 
Nonetheless, their image statistics are different. 
We further explain this fine-tuning process in subsection~\ref{subsec:domain}.

\begin{table*}[t]
\begin{tabularx}{\linewidth}{@{}l*{8}{C}}
\toprule
\multicolumn{2}{c}{Experiment} & $G(x)$ & $G(x,m)$ & $G(x,m)$ & $G(x,m)|^w$ & $G(x,m)|^w$ & $G(x,m)|^w_{ORB}$\\
& & $D(x,y)$ & $D(x,y)$ & $D(x,y,m)$ & $D(x,y,m)|^w$ & $D(x,y,m,n)|^w$ & $D(x,y,m,n)|^w$ \\
\midrule
$L_1 (\%)$ & $Full$ $image$ & 1.98 & 2.13 & 3.03 & 2.00 & \textbf{0.96} & 1.46 \\
 & $In$ & 10.04 & 7.70 & 7.37 & 5.73 & \textbf{4.99} & 5.16 \\
 & $Out$ & 1.70 & 1.95 & 2.92 & 1.88 & \textbf{0.80} & 1.32 \\
\midrule
$Feat$ & $Full$ $image$ & 1.29 & 1.95 & 1.44 & 1.45 & 0.80 & \textbf{0.70} \\
 & $In$ & 8.75 & 7.48 & 7.92 & 5.75 & 5.84 & \textbf{5.14} \\
 & $Out$ & 1.04 & 1.77 & 1.26 & 1.33 & 0.62 & \textbf{0.53} \\
\midrule
$PSNR$ & $Full$ $image$ & 29.89 & 29.85 & 28.31 & 30.46 & \textbf{34.03} & 33.03 \\
 & $In$ & 17.39 & 20.30 & 20.58 & 22.56 & 23.25 & \textbf{23.29} \\
 & $Out$ & 32.77 & 31.23 & 29.05 & 31.35 & \textbf{36.47} & 34.74 \\
\midrule
$SSIM$ & $Full$ $image$ & 0.985 & 0.985 & 0.981 & 0.987 & \textbf{0.995} & 0.993 \\
 & $In$ & 0.151 & 0.333 & 0.488 & 0.613 & 0.646 & \textbf{0.655} \\
 & $Out$ & 0.993 & 0.989 & 0.985 & 0.990 & \textbf{0.997} & 0.995 \\
\bottomrule
\end{tabularx}
\caption{\label{tab:ablationInput} Quantitative evaluations of our contributions in the inpainting task on the test synthetic images. 
The best results for almost all the inpainting metrics ($L_1$, $PSNR$ and $SSIM$) are obtained with the generator $G(x,m)|^w$ and the discriminator $D(x, y, m, n)|^w$. 
More correct features ($Feat$ metric) are detected though when adding the features based loss $G(x,m)|^w_{ORB}$. 
$Full$ $image$ designates the per-pixel error considering the whole image. 
For $In$ and $Out$ we refer respectively to the error per pixel considering the masked and unmasked pixels.}
\end{table*}

\subsection{Inpainting}
\label{subsec:inpainting}

\begin{figure*} [h]
\centering
\subfloat{\begin{overpic}[width=.161\linewidth]{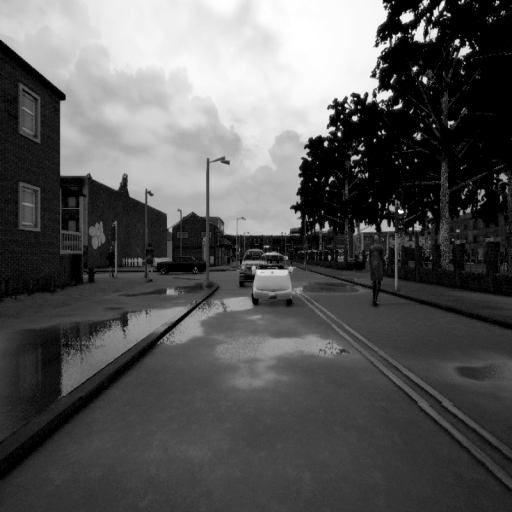}\put(1,56.0){\includegraphics[scale=0.05]{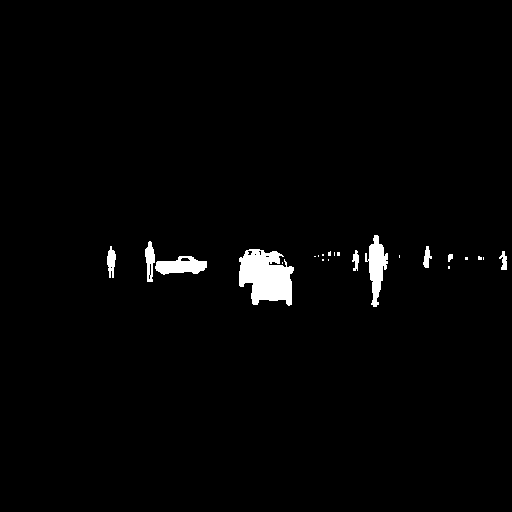}}\end{overpic}}
\hspace{\fill}
\subfloat{\includegraphics[width=.161\linewidth]{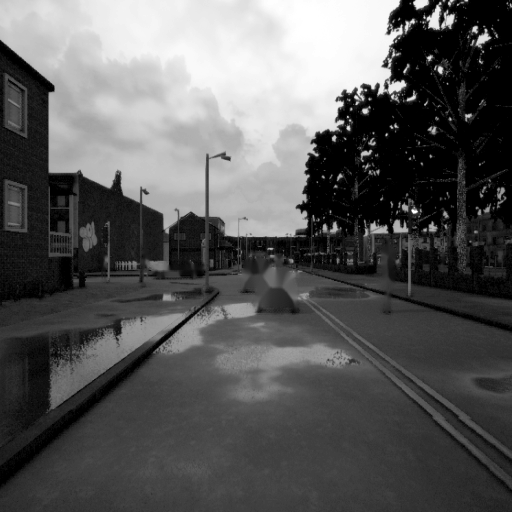}}
\hspace{\fill}
\subfloat{\includegraphics[width=.161\linewidth]{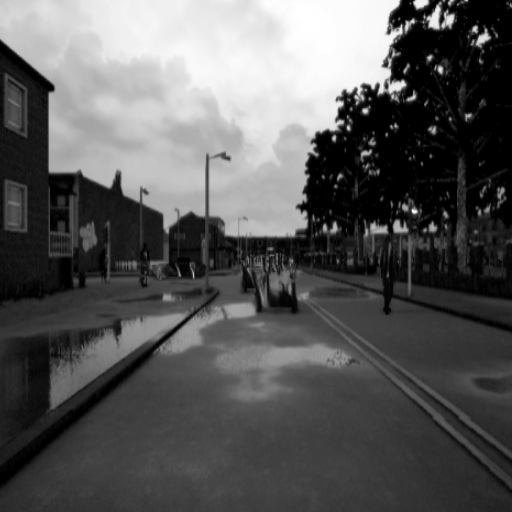}}
\hspace{\fill}
\subfloat{\includegraphics[width=.161\linewidth]{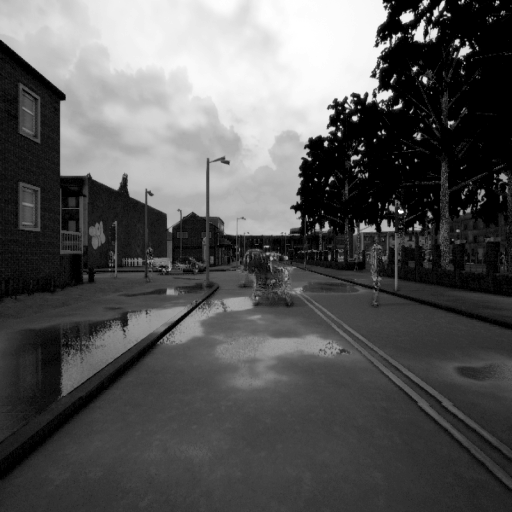}}
\hspace{\fill}
\subfloat{\includegraphics[width=.161\linewidth]{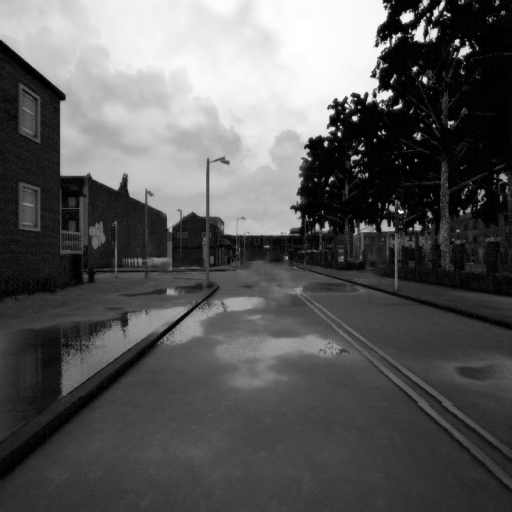}}
\hspace{\fill}
\subfloat{\includegraphics[width=.161\linewidth]{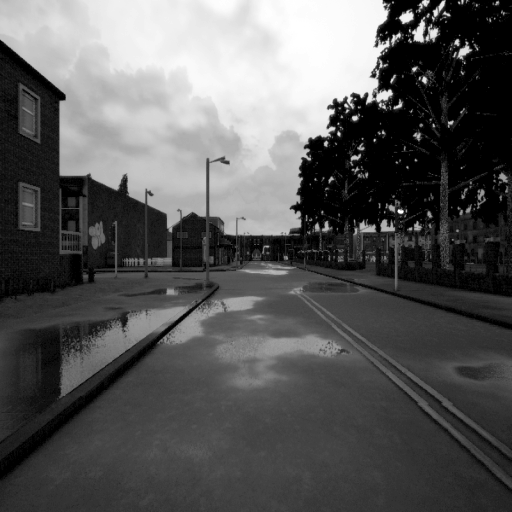}}
\\
\vspace*{-0.6\baselineskip}
\subfloat{\begin{overpic}[width=.161\linewidth]{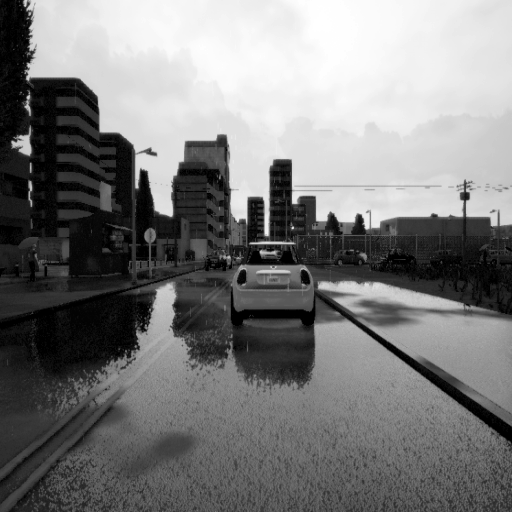}\put(1,56.0){\includegraphics[scale=0.05]{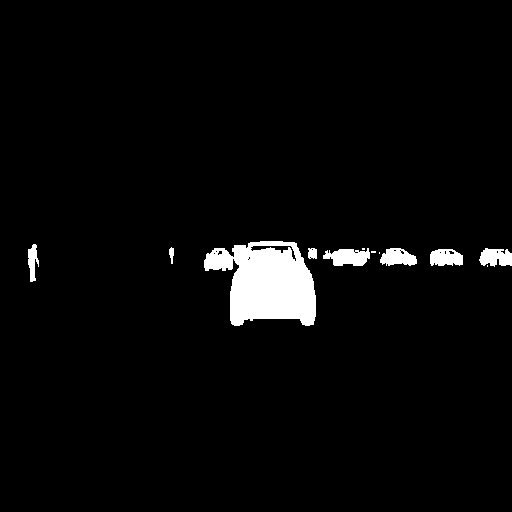}}\end{overpic}}
\hspace{\fill}
\subfloat{\includegraphics[width=.161\linewidth]{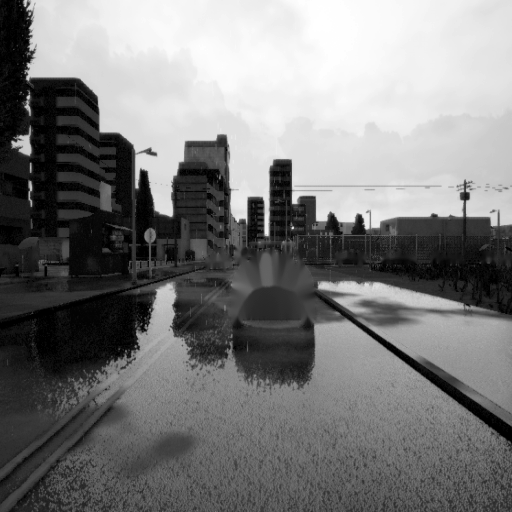}}
\hspace{\fill}
\subfloat{\includegraphics[width=.161\linewidth]{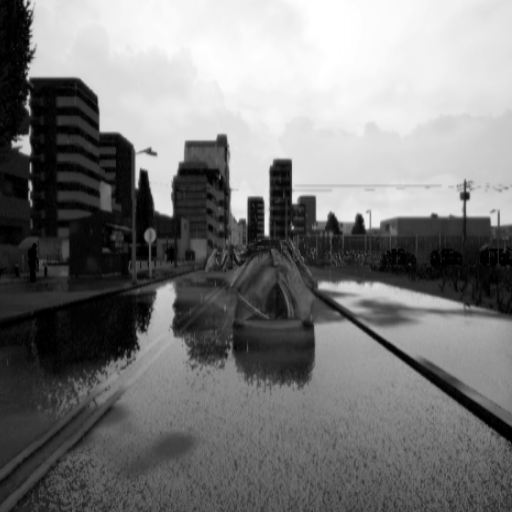}}
\hspace{\fill}
\subfloat{\includegraphics[width=.161\linewidth]{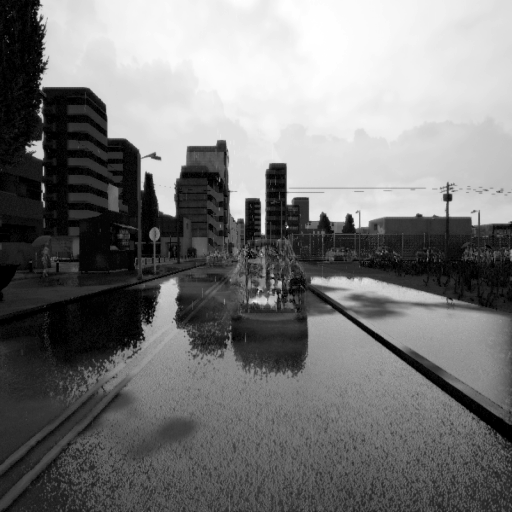}}
\hspace{\fill}
\subfloat{\includegraphics[width=.161\linewidth]{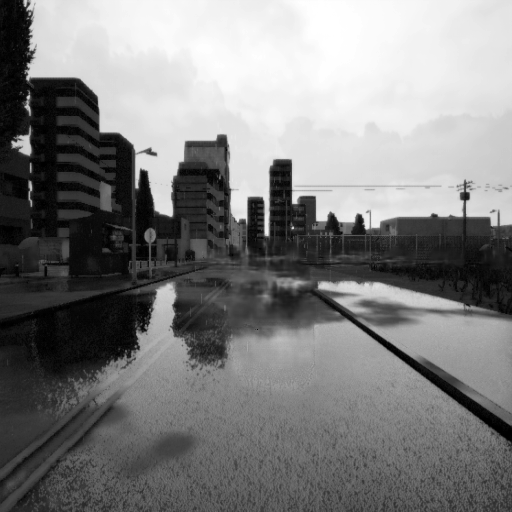}}
\hspace{\fill}
\subfloat{\includegraphics[width=.161\linewidth]{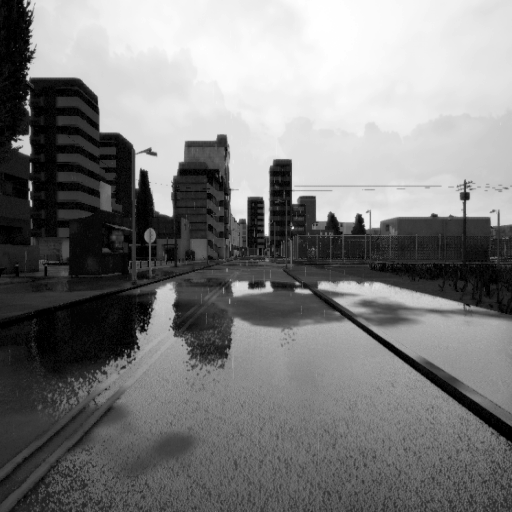}}
\\
\vspace*{-0.6\baselineskip}
\setcounter{subfigure}{0}
\subfloat[\label{fig:input} Input]{\begin{overpic}[width=.161\linewidth]{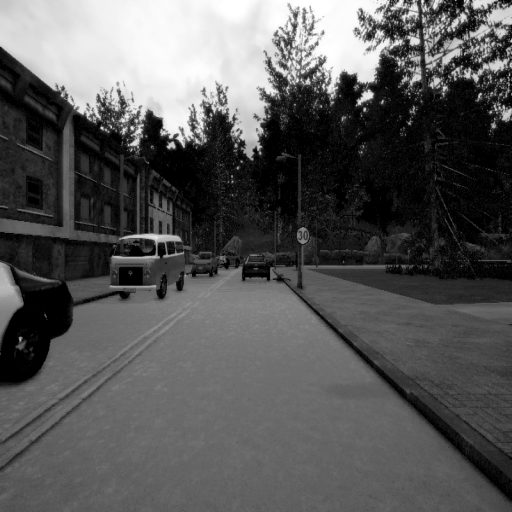}\put(1,56.0){\includegraphics[scale=0.05]{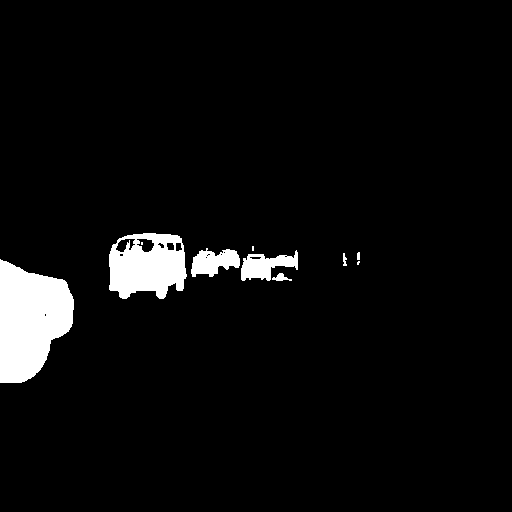}}\end{overpic}}
\hspace{\fill}
\subfloat[\label{fig:output_Geo} Geo1 \cite{telea2004image}]{\includegraphics[width=.161\linewidth]{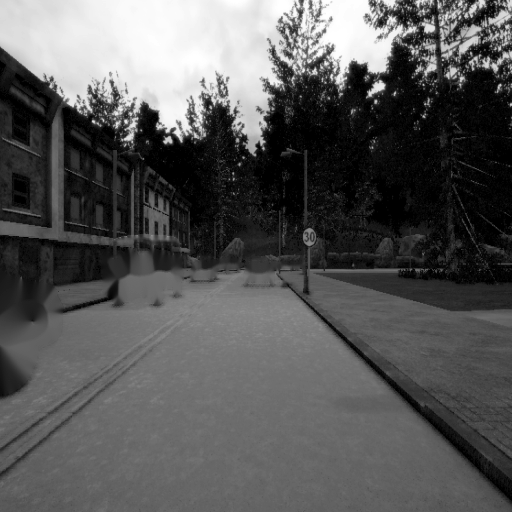}}
\hspace{\fill}
\subfloat[\label{fig:output_Lea1} Lea1 \cite{yu2018generative}]{\includegraphics[width=.161\linewidth]{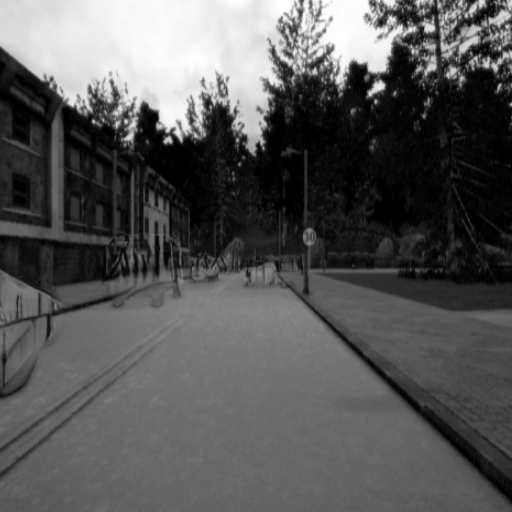}}
\hspace{\fill}
\subfloat[\label{fig:output_Lea2} Lea2 \cite{iizuka2017globally}]{\includegraphics[width=.161\linewidth]{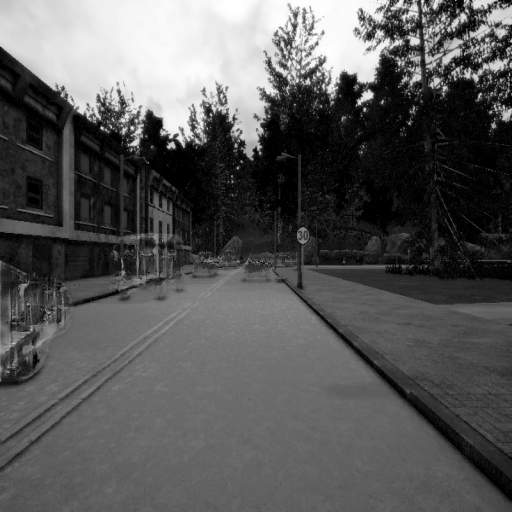}}
\hspace{\fill}
\subfloat[\label{fig:output_ours} Ours]{\includegraphics[width=.161\linewidth]{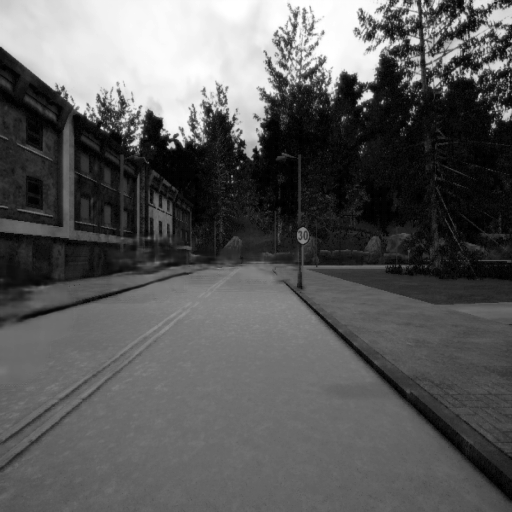}}
\hspace{\fill}
\subfloat[\label{fig:gt} Ground-truth]{\includegraphics[width=.161\linewidth]{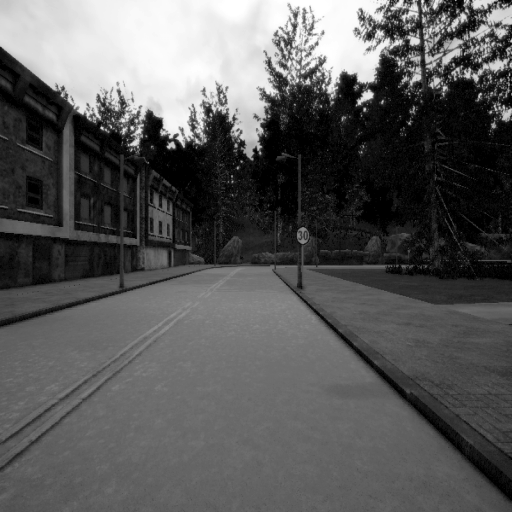}}
\caption{\label{fig:comparisonCARLA} Qualitative comparison of our method \protect\subref{fig:output_ours} against other inpainting techniques \protect\subref{fig:output_Geo}, \protect\subref{fig:output_Lea1}, \protect\subref{fig:output_Lea2} on our synthetic dataset. 
Our results are semantically and geometrically coherent, and do not show the dynamic objects' shadows, even if they are not included in the input mask.}
\end{figure*}

In this subsection we report the improvements achieved by our framework for inpainting. 
Table~\ref{tab:ablationInput} shows the ablation study of our work for the different reported inputs and losses. 
The existence of many possible solutions renders difficult to define a metric to evaluate image inpainting~\cite{yu2018generative}. 
Nevertheless, we follow previous works and report the $L_1$, $PSNR$ and $SSIM$ errors~\cite{wang2004image}, as well as a feature-based metric $Feat$. 
This last metric computes the FAST features detection as explained in subsection~\ref{subsec:ORB} for the output and ground-truth images, and compares them, similarly to Eq.~\ref{eq:det}.

Adding the dynamic/static mask as input for both the generator and discriminator helps obtaining better inpainting results within the images hole regions ($In$), at the expense of having worse quality results in the non-hole regions ($Out$).
Leveraging the unbalanced quantity of static and dynamic data within the dataset with $w$ (Eqs.~\ref{eq:wL1} and \ref{eq:wcGAN}) helps obtaining better results too. 
Providing the GAN's discriminator with the images noise makes it learn better to distinguish between real and fake images, and therefore the generator learns to produce more realistic images.
The ORB-based loss leads to slightly worse inpainting results according to $L_1$, PSNR and SSIM metrics, but renders this approach more useful for both localization and mapping tasks since more correct features are created.

\begin{figure*} [h]
\centering
\subfloat{\begin{overpic}[width=.23\linewidth]{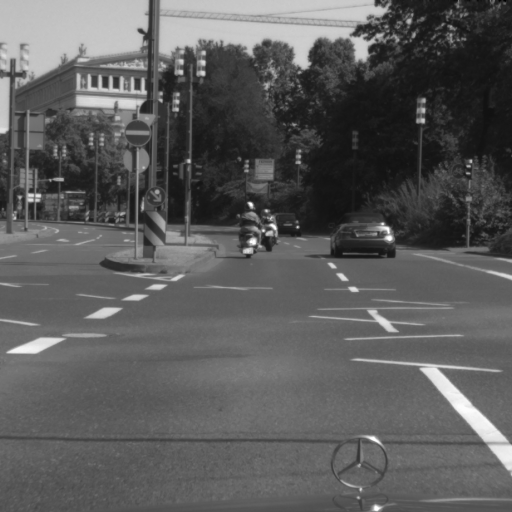}\put(1,78.0){\includegraphics[scale=0.075]{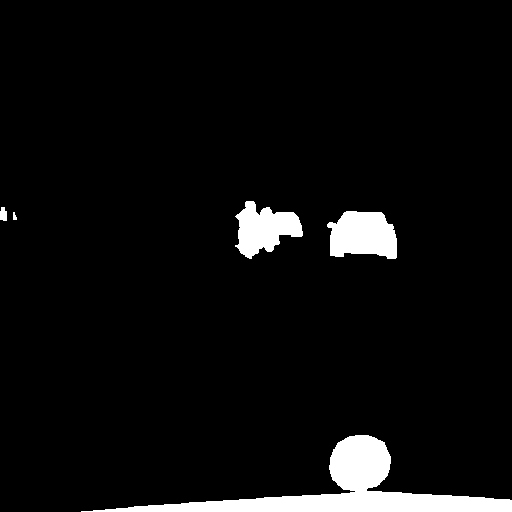}}\end{overpic}}
\hspace{\fill}
\subfloat{\includegraphics[width=.23\linewidth]{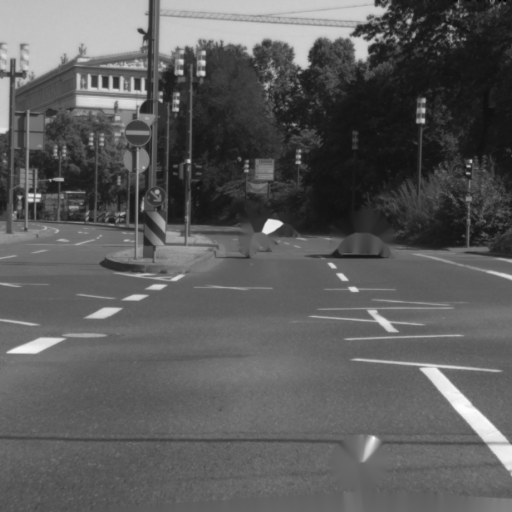}}
\hspace{\fill}
\subfloat{\includegraphics[width=.23\linewidth]{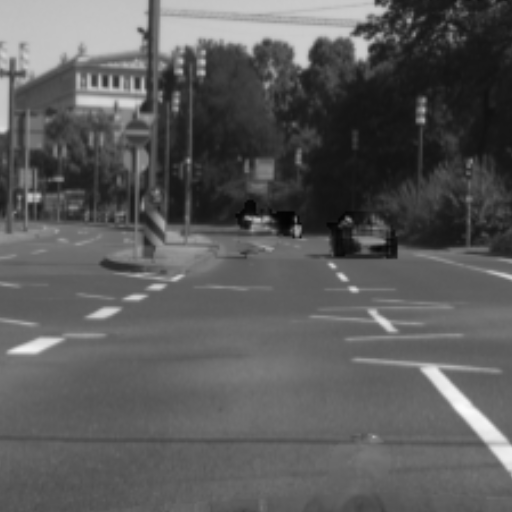}}
\hspace{\fill}
\subfloat{\includegraphics[width=.23\linewidth]{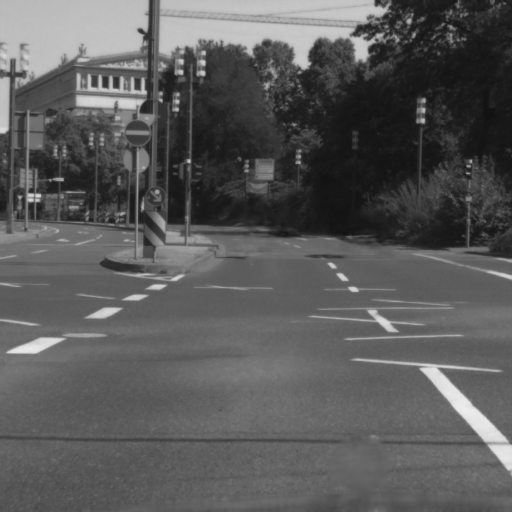}}
\\
\vspace*{-0.6\baselineskip}
\subfloat{\begin{overpic}[width=.23\linewidth]{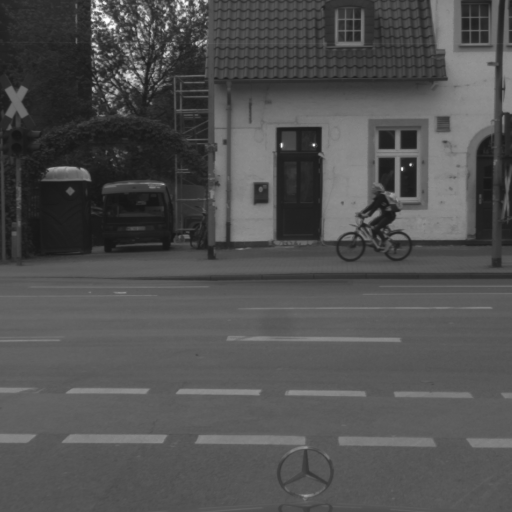}\put(1,78.0){\includegraphics[scale=0.075]{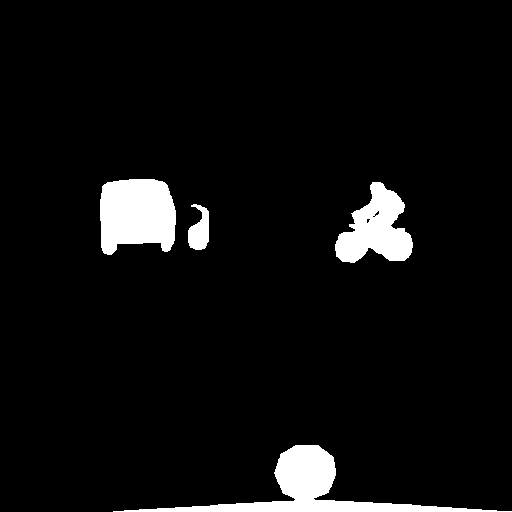}}\end{overpic}}
\hspace{\fill}
\subfloat{\includegraphics[width=.23\linewidth]{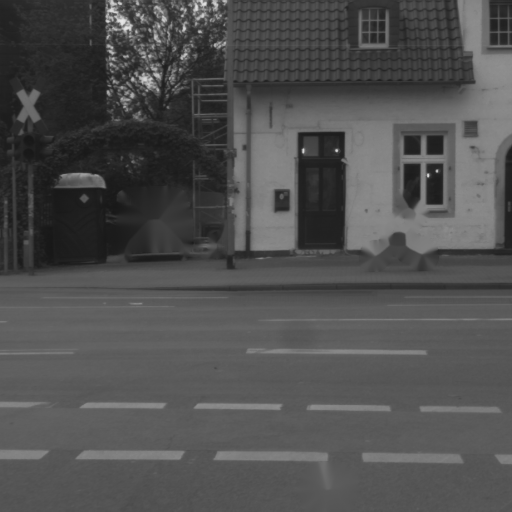}}
\hspace{\fill}
\subfloat{\includegraphics[width=.23\linewidth]{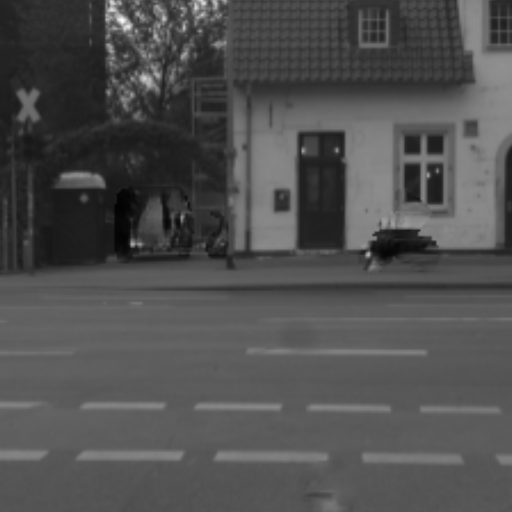}}
\hspace{\fill}
\subfloat{\includegraphics[width=.23\linewidth]{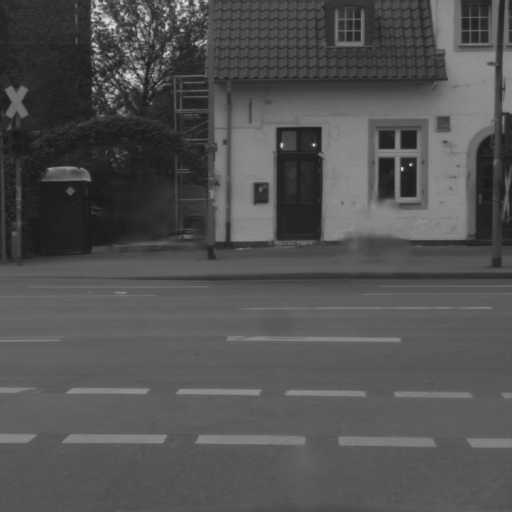}}
\\
\vspace*{-0.6\baselineskip}
\setcounter{subfigure}{0}
\subfloat[\label{fig:nsynth_input}Input]{\begin{overpic}[width=.23\linewidth]{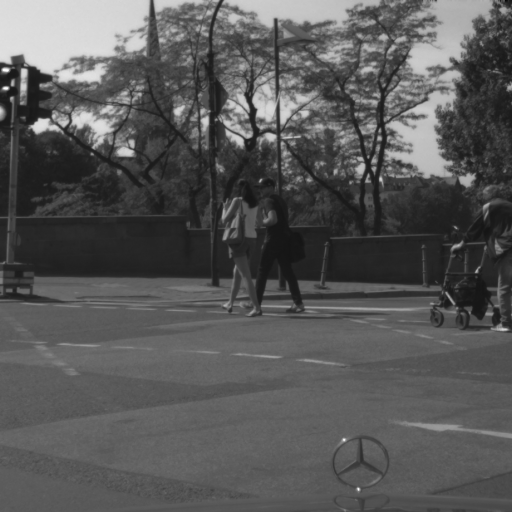}\put(1,78.0){\includegraphics[scale=0.075]{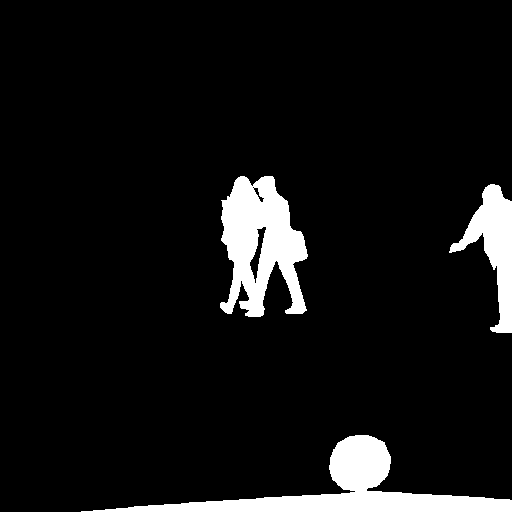}}\end{overpic}}
\hspace{\fill}
\subfloat[\label{fig:nsynth_output_Geo}Geo1~\cite{telea2004image}]{\includegraphics[width=.23\linewidth]{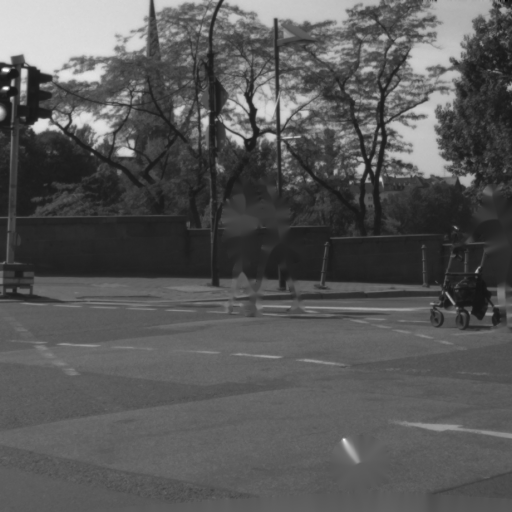}}
\hspace{\fill}
\subfloat[\label{fig:nsynth_output_Lea1}Lea1~\cite{yu2018generative}]{\includegraphics[width=.23\linewidth]{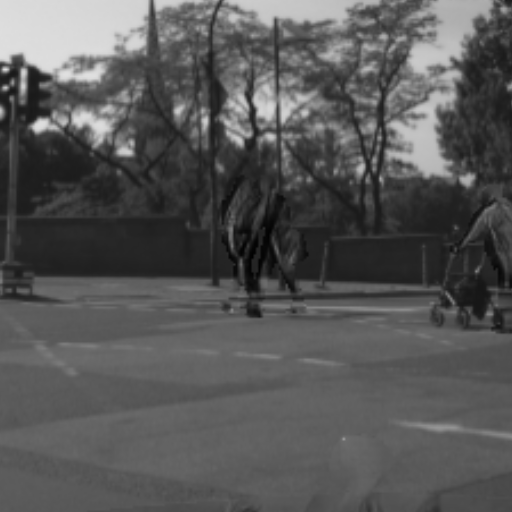}}
\hspace{\fill}
\subfloat[\label{fig:SynthToReal}Ours]{\includegraphics[width=.23\linewidth]{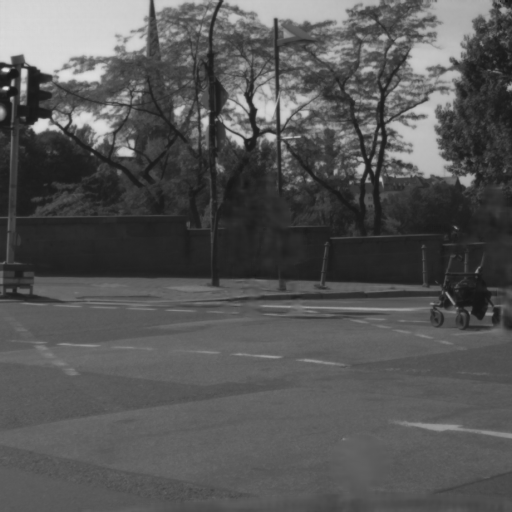}}
\caption{\label{fig:comparisonCITYSCAPES}Comparison of our method \protect\subref{fig:SynthToReal} against other image inpainting approaches \protect\subref{fig:nsynth_output_Geo}, \protect\subref{fig:nsynth_output_Lea1} on the Cityscapes validation dataset~\cite{cordts2016cityscapes}. 
\protect\subref{fig:nsynth_output_Lea1} and \protect\subref{fig:SynthToReal} show the results when real images have been incorporated into our training set together with the synthetic images with a ratio of $1/10$.}
\end{figure*}

\begin{figure*} [h]
\centering
\subfloat{\begin{overpic}[width=.23\linewidth]{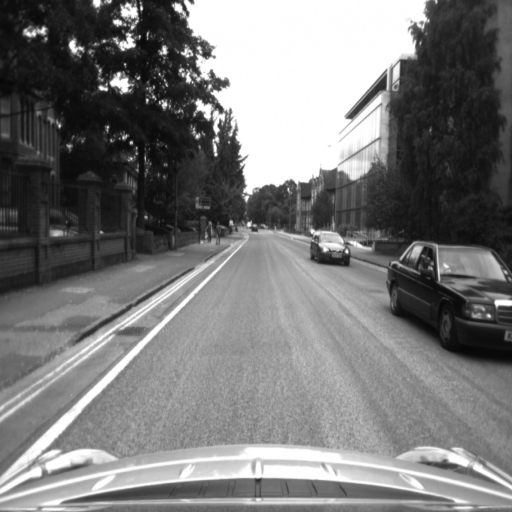}\put(1,78.0){\includegraphics[scale=0.075]{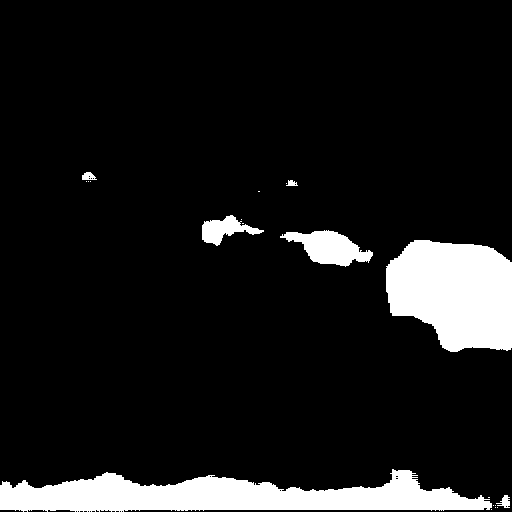}}\end{overpic}}
\hspace{\fill}
\subfloat{\includegraphics[width=.23\linewidth]{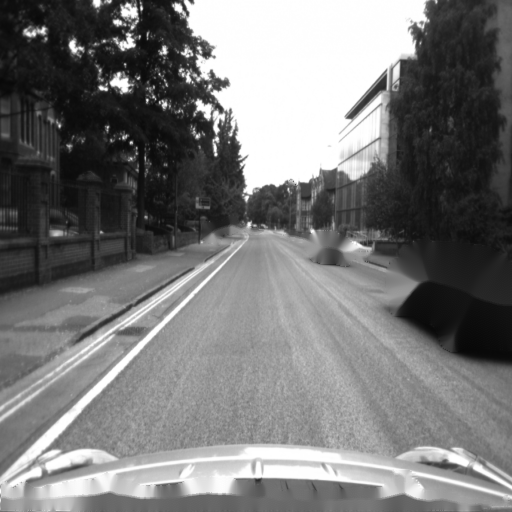}}
\hspace{\fill}
\subfloat{\includegraphics[width=.23\linewidth]{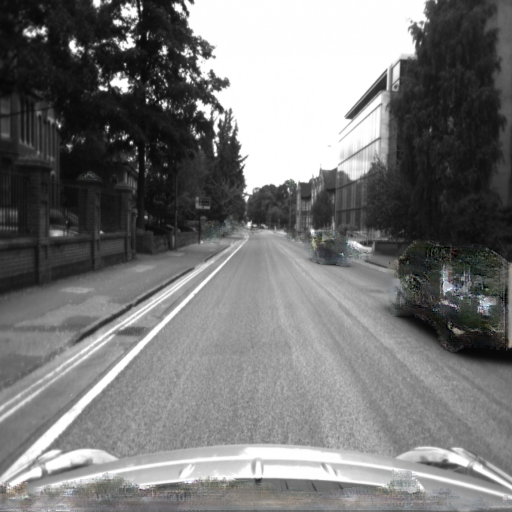}}
\hspace{\fill}
\subfloat{\includegraphics[width=.23\linewidth]{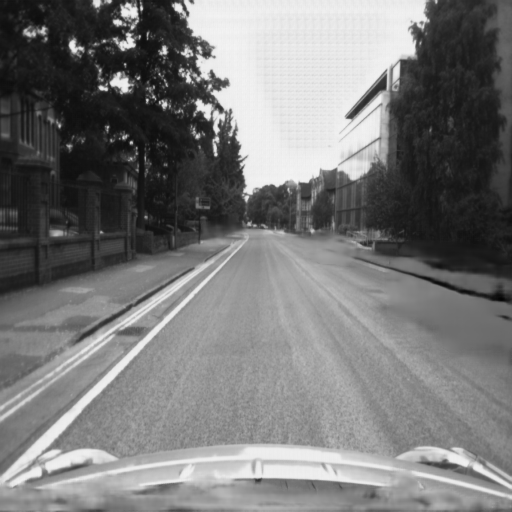}}
\\
\vspace*{-0.6\baselineskip}
\subfloat{\begin{overpic}[width=.23\linewidth]{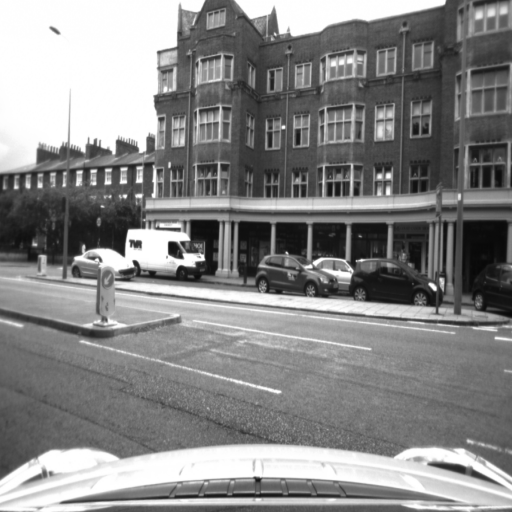}\put(1,78.0){\includegraphics[scale=0.075]{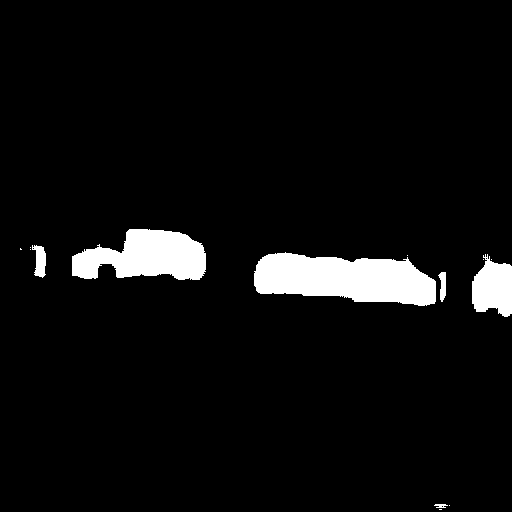}}\end{overpic}}
\hspace{\fill}
\subfloat{\includegraphics[width=.23\linewidth]{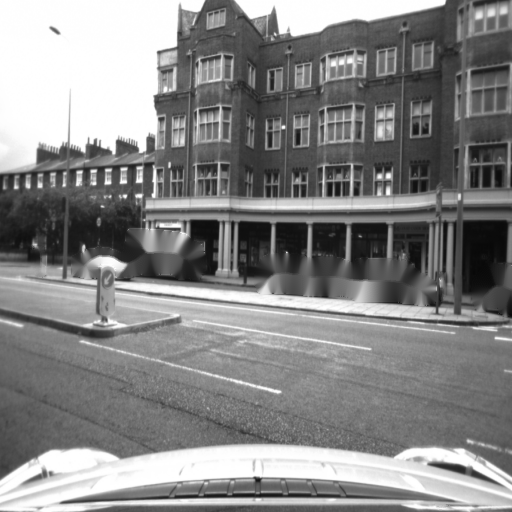}}
\hspace{\fill}
\subfloat{\includegraphics[width=.23\linewidth]{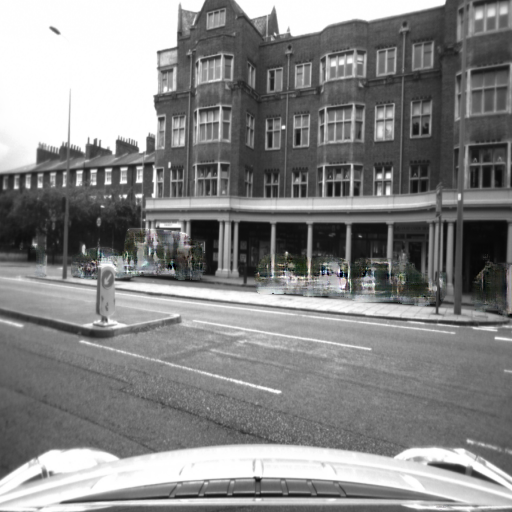}}
\hspace{\fill}
\subfloat{\includegraphics[width=.23\linewidth]{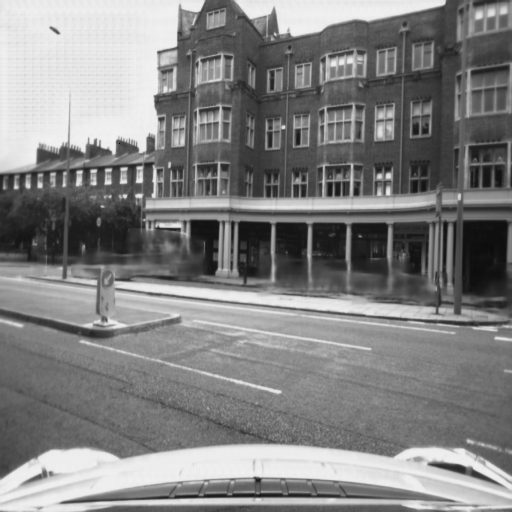}}
\\
\vspace*{-0.6\baselineskip}
\setcounter{subfigure}{0}
\subfloat[\label{fig:Oxford_in}Input]{\begin{overpic}[width=.23\linewidth]{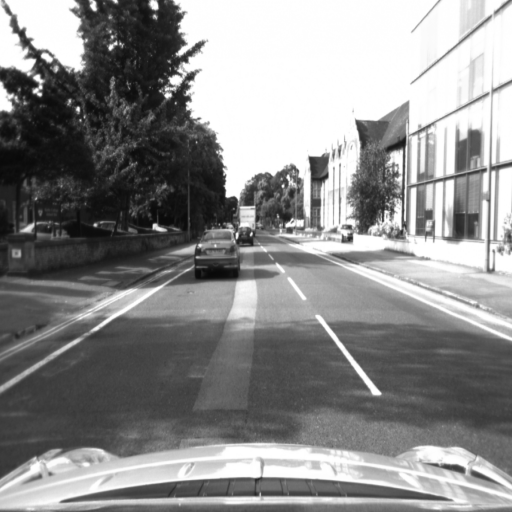}\put(1,78.0){\includegraphics[scale=0.075]{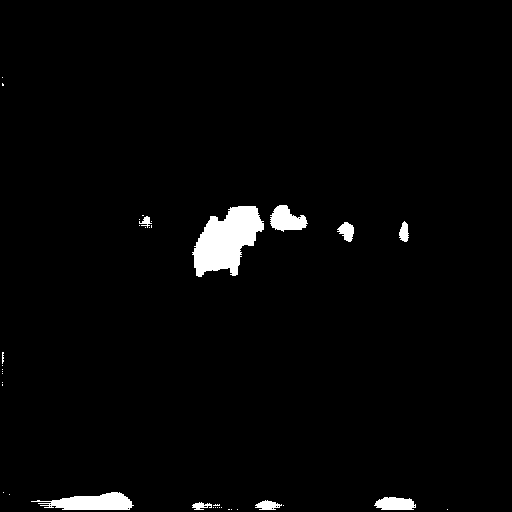}}\end{overpic}}
\hspace{\fill}
\subfloat[\label{fig:Oxford_Geo}Geo~\cite{telea2004image}]{\includegraphics[width=.23\linewidth]{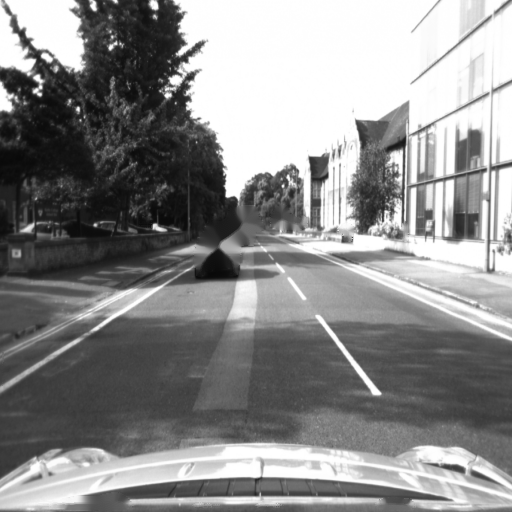}}
\hspace{\fill}
\subfloat[\label{fig:Oxford_Lea}Lea1~\cite{yu2018generative}]{\includegraphics[width=.23\linewidth]{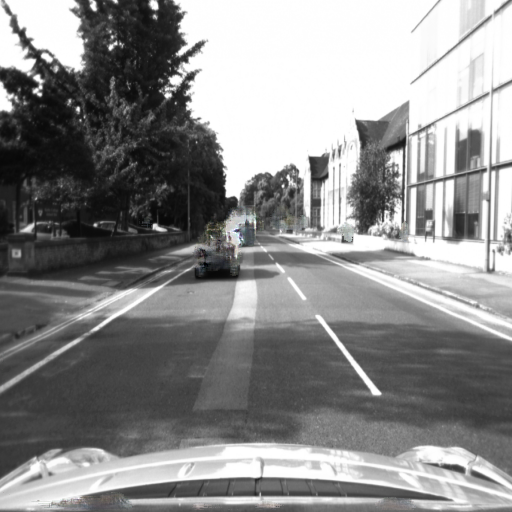}}
\hspace{\fill}
\subfloat[\label{fig:Oxford_Ours}Ours]{\includegraphics[width=.23\linewidth]{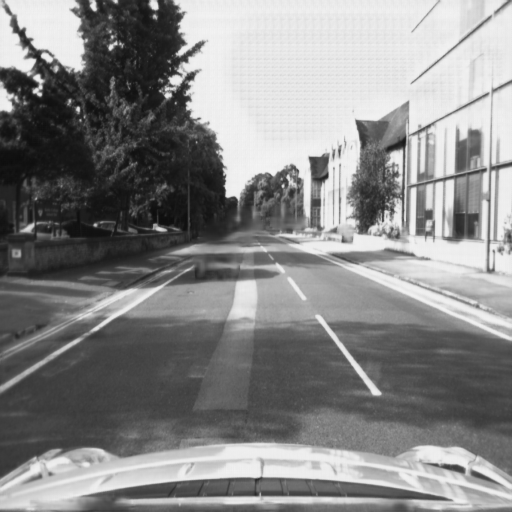}}
\caption{\label{fig:comparisonOxford}Comparison of our method \protect\subref{fig:Oxford_Ours} against other image inpainting approaches \protect\subref{fig:Oxford_Geo}, \protect\subref{fig:Oxford_Lea} on the Oxford Robotcar dataset~\cite{RobotCarDatasetIJRR}. 
\protect\subref{fig:Oxford_Lea} and \protect\subref{fig:Oxford_Ours} show the results when Cityscapes images have been incorporated into our training set together with the synthetic images with a $1/10$ ratio. 
The binary dynamic/static mask computed for every input image has been added on the top left corner of every raw.}
\end{figure*}

\textbf{Baselines for Inpainting}

We compare qualitatively and quantitatively our ``inpainting'' method with four other state-of-the-art approaches:
\begin{itemize}[leftmargin=5.0mm]
\setlength\itemsep{0.0cm}
\item{\textbf{Geo1}, \textbf{Geo2}:} two non-learning approaches~\cite{telea2004image, bertalmio2001navier}.
\item{\textbf{Lea1}, \textbf{Lea2}:} two deep learning based methods~\cite{yu2018generative, iizuka2017globally}.
\end{itemize}

{\color{black} For a fair comparison, we have trained the approach by Yu \textit{et~al.} (\textbf{Lea1}) with our same training data. 
Iizuka \textit{et~al.} (\textbf{Lea2}) do not have their training code available. We have directly used their release model~\cite{iizuka2017globally} trained on the Places2 dataset \cite{zhou2017places}. %
This dataset contains images of urban streets from a car perspective similar to ours.  A more direct comparison is not possible.}
We provide them with the same mask than to our method to generate the holes in the images. 
We evaluate qualitatively on the 3000 images from our synthetic test dataset, on the 500 validation images from the Cityscapes dataset \cite{cordts2016cityscapes} {\color{black}and on the images from the Oxford Robotcar dataset~\cite{RobotCarDatasetIJRR}}. 
We can see in Figs.~\ref{fig:comparisonCARLA}, \ref{fig:comparisonCITYSCAPES} and~\ref{fig:comparisonOxford} the qualitative comparisons on these three datasets. 
Note that results generated with both \textbf{Lea1} and \textbf{Lea2} have been generated with the color images and then converted to gray scale for visual comparison.
Visually, we see that our method obtains a more realistic output (these results are computed without the ORB loss for an inpainting oriented comparison). 
Also, it is the only one capable of removing the shadows generated by the dynamic objects even though they are not included in the dynamic/static mask (Fig.~\ref{fig:comparisonCARLA} row 2 and Fig.~\ref{fig:comparisonOxford} row 1).
The utilized masks are included in the images in Figs.~\ref{fig:input} and~\ref{fig:Oxford_in} respectively.

Table \ref{tab:stateoftheart} shows the quantitative comparison of our method against \textbf{Geo1}, \textbf{Geo2}, \textbf{Lea1} and \textbf{Lea2} on our CARLA dataset. 
It is not possible to quantitatively measure the performance of the different methods on the Cityscapes and Oxford Robotcar datasets, since ground-truth does not exist. 
By following these results, we can claim that our method outperforms both qualitatively and quantitatively the other approaches.

\begin{table} [h]
\begin{tabularx}{\linewidth}{@{}l*{5}{C}}
\toprule
 & Geo1~\cite{telea2004image} & Geo2~\cite{bertalmio2001navier} & Lea1~\cite{yu2018generative} & \mbox{Lea2~ \cite{iizuka2017globally}} & Ours \\
\midrule
$L_1 (\%)$ & 6.55 & 6.69 & 8.91 & 9.50 & \textbf{4.99} \\
$Feat$ & 6.24 & 6.33 & 10.03 & 10.63 & \textbf{5.84} \\
$PSNR$ & 21.03 & 20.90 & 18.16 & 18.23 & \textbf{23.25} \\
$SSIM$ & 0.479 & 0.467 & 0.309 & 0.316 & \textbf{0.646} \\
\bottomrule
\end{tabularx}
\caption{\label{tab:stateoftheart} {\color{black}Quantitative results of our method against other inpainting approaches in our CARLA dataset. 
For a fair comparison, we only report the different errors within the images' hole regions since the other methods are conceived to only significantly modify such parts.}}
\end{table}

As seen in Fig.~\ref{fig:comparisonOxford} row 1, the fact that our method does not perform pure inpainting but image-to-image translation with the help of a dynamic/static mask allows us to modify not only the dynamic objects themselves, but also their shadows or reflections. 
We believe that the main underlying reasoning for this is the direct supervision for image-to-image translation. 
Also, since during the training with real-world data the segmentation masks are not 100~\% accurate, the model learns that it has to modify mainly the areas of the mask and, in case a smooth representation of the world is not obtained, also its surroundings. 
We believe that, was the training performed with perfect masks that also cover the shadows, the model would not learn to handle the shadows of dynamic objects or the inaccuracies of segmentation. 
We want to highlight the importance of the inpainting robustness to inaccurate segmentation masks since, in practice, partial or missing segmentation happens frequently. 
Empty Cities cannot handle missing detections but can cope with partial segmentations covering at least the 85~\% of the object image.


We hereby report some metrics evaluating how our framework behaves with the dynamic objects's shadows. 
Thresholding the difference between the dynamic and static image, and subtracting the dynamic-objects mask yields its dynamic objects shadows and reflections mask. 
We have first generated the shadows ground truth of our CARLA dataset, and then computed the shadows masks for our inpainted images in the same way. 
The intersection over union of the estimated shadows against the ground truth is 42.8~\%. 
Following recent works in shadow detection~\cite{hosseinzadeh2018fast}, we also report our method's shadow, non-shadow and total accuracy (59.7~\%, 99.8~\% and 99.5~\% respectively).
With this framework, we can remove almost 50~\% of the shadows of the dynamic objects of our CARLA test dataset. 
Admitting this could be improved, our method's non-shadow accuracy is almost 100~\%, which means that it does not modify other objects' shadows.

\subsection{Transfer to Real Data}
\label{subsec:domain}

Models trained on synthetic data can be useful for real world vision tasks~\cite{gaidon2016virtual, peris2012towards, skinner2016high, tobin2017domain}. 
Accordingly, we provide a study of synthetic-to-real transfer learning using data from the Cityscapes dataset \cite{cordts2016cityscapes}, which offers a variety of urban real-world environments similar to the synthetic ones.

When testing our method on real data, we see qualitatively that the synthesized images show some artifacts. 
This happens because such data has different statistics than the real one, and therefore cannot be easily used. 
The combination of real and synthetic data is possible during training despite the lack of ground-truth static real images. 
In the case of the real images, the network only learns the texture and the style of the static real world by encoding its information and decoding back the original image non-hole regions. 
The synthetic data is substantially more plentiful and has information about the inpainting process. 
The rendering, however, is far from realistic. 
Thus, the chosen representation attempts to bridge the reality gap encountered when using simulated data, and to remove the need for domain adaptation. 

We provide implementation details: we have finetuned our model with real data for 25 epochs with a real/synthetic images ratio of $1/10$. 
On the one hand, for every ten images there are nine synthetic images that provide our model with information about the inpainting task. 
On the other hand, one image out of those ten is a real image from the Cityscapes train dataset. 
There is groundtruth of its semantic information but there is no groundtruth of its static representation. 
In such cases we do backpropagation of the loss derivative only on those image areas that we consider as static. 
This way, the model can learn both the inpainting task and the static real-world texture. 
Once the model is adapted to real-world data, it can be directly used in completely new real-world scenarios, \textit{e.g.}, the Oxford Robotcar dataset~\cite{RobotCarDatasetIJRR}.

\section{Experiments}

\subsection{{\color{black}Visual Odometry}}
\label{subsec:localization}

We have evaluated Empty Cities on 20 CARLA synthetic sequences and on 9 sequences from real-world new environments. 
For these VO experiments we have chosen the state-of-the-art feature-based system ORB-SLAM~\cite{mur2017orb}\footnote{ORB-SLAM acts as visual odometry in trajectories without loop closures.} and the direct method DSO~\cite{engel2018direct}. 
The former is ideal to test the influence of our ORB features loss, and the latter is useful to prove that different systems can also benefit from this approach.


\textbf{Baselines for VO in our Synthetic Dataset}

\begin{figure*} [h]
\centering
\subfloat[\label{fig:BP_ORB_In}Dynamic images]{\pgfplotstableread{%
sequence min q1 q2 q3 max 
1 0.784 0.926 1.202 1.421 4.531
3 0.055 0.084 0.109 0.160 0.604
5 0.278 0.438 1.148 6.350 7.052
6 1.015 1.160 1.441 3.134 12.650
8 0.440 0.535 0.670 0.815 1.039
9 2.432 3.863 5.292 9.641 12.631
10 1.081 1.467 1.729 2.306 3.276
11 0.330 0.463 0.553 3.681 15.172
12 2.681 9.274 9.981 10.249 10.778
13 0.586 0.630 0.785 0.941 1.663
14 0.496 0.552 0.570 0.602 1.221
15 0.436 0.484 0.504 0.538 0.627
17 0.592 0.689 0.724 0.823 0.879
18 0.492 0.658 2.099 2.881 6.575
19 2.987 3.660 4.301 4.612 5.057
20 4.518 5.233 5.460 5.677 5.959
22 10.800 12.356 12.866 14.172 19.587
23 0.585 3.221 3.741 4.006 4.216
24 0.390 0.541 0.608 0.814 2.676
25 1.520 2.165 2.396 2.594 3.086
}\datatable

\begin{tikzpicture}
\pgfplotstablegetrowsof{\datatable}
\pgfmathtruncatemacro{\rownumber}{\pgfplotsretval - 1}
\begin{axis}[
width=5.65cm,
enlargelimits = false,
boxplot/draw direction = x,
grid = major,
yticklabels = {00,02,04,06,08,10,12,14,16,18},
ytick = {1,3,...,\the\numexpr\rownumber + 1},
y tick label style = {scale = 1.0, font = \bfseries, rotate = 0,align = center},
xticklabels = {0,2,...,12},
xtick = {0,2,...,12},
x tick label style = {scale = 1.0, rotate = 0,align = center},
xmin=0, xmax=13,
cycle list/Set1-5 ]

\pgfplotsinvokeforeach{0,...,\rownumber}{

    \pgfplotstablegetelem{#1}{min}\of\datatable
    \edef\mymin{\pgfplotsretval}

    \pgfplotstablegetelem{#1}{q1}\of\datatable
    \edef\mylowerq{\pgfplotsretval}

    \pgfplotstablegetelem{#1}{q2}\of\datatable
    \edef\mymedian{\pgfplotsretval}

    \pgfplotstablegetelem{#1}{q3}\of\datatable
    \edef\myupperq{\pgfplotsretval}

    \pgfplotstablegetelem{#1}{max}\of\datatable
    \edef\mymax{\pgfplotsretval}

    \typeout{\mymin,\mylowerq,\mymedian,\myupperq,\mymax}
    \edef\temp{\noexpand\addplot+[,
        boxplot prepared={
            lower whisker=\mymin,
            upper whisker=\mymax,
            lower quartile=\mylowerq,
            upper quartile=\myupperq,
            median=\mymedian,
            every box/.style={solid,fill,opacity=0.5},
            every whisker/.style={solid },
            every median/.style={solid},
        }, 
        ]coordinates {};}
    \temp
}
\end{axis}

\end{tikzpicture}}
\subfloat[\label{fig:BP_Dyna}Dynamic images w/ mask]{\pgfplotstableread{%
sequence min q1 q2 q3 max 
1 0.344 0.429 0.471 0.697 1.020
3 0.070 0.103 0.134 0.155 0.415
5 0.213 0.305 0.604 0.722 0.941
6 0.881 1.235 1.710 2.647 3.865
8 0.436 0.656 0.743 0.813 0.959
9 2.847 5.955 9.348 11.153 12.595
10 0.837 1.460 1.628 1.861 2.202
11 0.169 0.242 0.304 0.338 0.464
12 0.686 0.765 0.987 1.400 1.546
13 0.495 0.631 0.895 1.114 2.030
14 0.541 0.577 0.620 0.712 0.777
15 0.451 0.510 0.523 0.543 0.576
17 0.499 0.568 0.731 0.923 0.998
18 0.460 0.739 1.170 2.070 12.635
19 0.727 1.034 3.027 3.615 5.366
20 3.810 4.056 4.171 4.318 4.858
22 2.566 6.472 9.206 11.298 12.635
23 0.426 0.479 0.726 0.838 1.081
24 0.242 0.334 0.367 0.390 1.383
25 1.348 1.936 2.236 2.463 3.340
}\datatable

\begin{tikzpicture}
\pgfplotstablegetrowsof{\datatable}
\pgfmathtruncatemacro{\rownumber}{\pgfplotsretval - 1}
\begin{axis}[
width=5.65cm,
enlargelimits = false,
boxplot/draw direction = x,
grid = major,
yticklabels = {,,},
ytick = {1,3,...,\the\numexpr\rownumber + 1},
xticklabels = {0,2,...,12},
xtick = {0,2,...,12},
x tick label style = {scale = 1.0, rotate = 0,align = center},
xmin=0, xmax=13,
cycle list/Set1-5 ]

\pgfplotsinvokeforeach{0,...,\rownumber}{

    \pgfplotstablegetelem{#1}{min}\of\datatable
    \edef\mymin{\pgfplotsretval}

    \pgfplotstablegetelem{#1}{q1}\of\datatable
    \edef\mylowerq{\pgfplotsretval}

    \pgfplotstablegetelem{#1}{q2}\of\datatable
    \edef\mymedian{\pgfplotsretval}

    \pgfplotstablegetelem{#1}{q3}\of\datatable
    \edef\myupperq{\pgfplotsretval}

    \pgfplotstablegetelem{#1}{max}\of\datatable
    \edef\mymax{\pgfplotsretval}

    \typeout{\mymin,\mylowerq,\mymedian,\myupperq,\mymax}
    \edef\temp{\noexpand\addplot+[,
        boxplot prepared={
            lower whisker=\mymin,
            upper whisker=\mymax,
            lower quartile=\mylowerq,
            upper quartile=\myupperq,
            median=\mymedian,
            every box/.style={solid,fill,opacity=0.5},
            every whisker/.style={solid },
            every median/.style={solid},
        }, 
        ]coordinates {};}
    \temp
}
\end{axis}

\end{tikzpicture}}
\subfloat[\label{fig:BP_ORB_Ours_Feat}Our images w/ the ORB loss]{\pgfplotstableread{%
sequence min q1 q2 q3 max 
1 0.368 0.402 0.504 0.590 0.691
3 0.059 0.063 0.074 0.095 0.221
5 0.104 0.164 0.185 0.234 0.287
6 0.603 0.823 0.966 1.153 7.601
8 0.634 0.748 0.797 0.864 0.960
9 2.824 3.304 5.599 6.800 8.610
10 0.874 1.180 1.417 1.566 2.290
11 0.179 0.210 0.214 0.326 0.352
12 0.357 0.577 0.632 0.785 1.056
13 0.226 0.381 0.512 0.527 0.826
14 0.499 0.524 0.598 0.874 0.953
15 0.449 0.468 0.522 0.551 0.552
17 0.505 0.548 0.622 0.739 0.830
18 0.610 1.411 1.759 2.634 4.681
19 1.144 1.600 2.299 3.509 4.377
20 4.039 4.923 5.241 7.306 11.654
22 4.871 5.528 6.201 8.184 8.691
23 0.205 0.306 0.459 0.542 0.808
24 0.198 0.294 0.355 0.380 0.727
25 0.953 1.153 1.182 1.620 1.803
}\datatable

\begin{tikzpicture}
\pgfplotstablegetrowsof{\datatable}
\pgfmathtruncatemacro{\rownumber}{\pgfplotsretval - 1}
\begin{axis}[
width=5.65cm,
enlargelimits = false,
grid = major,
boxplot/draw direction=x,
yticklabels = {,,},
ytick = {1,3,...,\the\numexpr\rownumber + 1},
xticklabels = {0,2,...,12},
xtick = {0,2,...,12},
x tick label style = {scale = 1.0, rotate = 0,align = center},
xmin=0, xmax=13,
cycle list/Set1-5 ]

\pgfplotsinvokeforeach{0,...,\rownumber}{

    \pgfplotstablegetelem{#1}{min}\of\datatable
    \edef\mymin{\pgfplotsretval}

    \pgfplotstablegetelem{#1}{q1}\of\datatable
    \edef\mylowerq{\pgfplotsretval}

    \pgfplotstablegetelem{#1}{q2}\of\datatable
    \edef\mymedian{\pgfplotsretval}

    \pgfplotstablegetelem{#1}{q3}\of\datatable
    \edef\myupperq{\pgfplotsretval}

    \pgfplotstablegetelem{#1}{max}\of\datatable
    \edef\mymax{\pgfplotsretval}

    \typeout{\mymin,\mylowerq,\mymedian,\myupperq,\mymax}
    \edef\temp{\noexpand\addplot+[,
        boxplot prepared={
            lower whisker=\mymin,
            upper whisker=\mymax,
            lower quartile=\mylowerq,
            upper quartile=\myupperq,
            median=\mymedian,
            every box/.style={solid,fill,opacity=0.5},
            every whisker/.style={solid },
            every median/.style={solid},
        }, 
        ]coordinates {};}
    \temp
}
\end{axis}
\end{tikzpicture}}
\subfloat[\label{fig:BP_ORB_Gt}Static images]{\pgfplotstableread{%
sequence min q1 q2 q3 max 
1 0.265 0.390 0.557 0.620 0.689
3 0.061 0.072 0.089 0.104 0.132
5 0.138 0.191 0.289 0.381 0.698
6 0.938 1.458 2.359 3.339 8.614
8 0.488 0.566 0.599 0.711 0.848
9 0.869 1.530 2.536 3.257 8.516
10 1.254 1.731 2.003 2.234 2.619
11 0.242 0.269 0.327 0.365 0.389
12 0.605 0.647 0.936 1.271 1.559
13 0.488 0.670 0.771 0.899 1.299
14 0.507 0.526 0.600 0.769 1.129
15 0.464 0.517 0.542 0.552 0.557
17 0.625 0.715 0.757 0.787 0.842
18 0.407 0.792 1.054 1.649 5.026
19 0.572 0.785 1.000 1.478 2.063
20 5.183 5.561 5.735 5.810 5.986
22 2.067 4.181 5.860 7.261 13.777
23 0.291 0.370 0.428 0.516 0.742
24 0.277 0.340 0.358 0.402 0.412
25 1.376 1.480 1.644 2.017 2.474
}\datatable

\begin{tikzpicture}
\pgfplotstablegetrowsof{\datatable}
\pgfmathtruncatemacro{\rownumber}{\pgfplotsretval - 1}
\begin{axis}[
width=5.65cm,
enlargelimits = false,
boxplot/draw direction = x,
grid = major,
yticklabels = {,,},
ytick = {1,3,...,\the\numexpr\rownumber + 1},
xticklabels = {0,2,...,12},
xtick = {0,2,...,12},
x tick label style = {scale = 1.0, rotate = 0,align = center},
xmin=0, xmax=13,
cycle list/Set1-5 ]

\pgfplotsinvokeforeach{0,...,\rownumber}{

    \pgfplotstablegetelem{#1}{min}\of\datatable
    \edef\mymin{\pgfplotsretval}

    \pgfplotstablegetelem{#1}{q1}\of\datatable
    \edef\mylowerq{\pgfplotsretval}

    \pgfplotstablegetelem{#1}{q2}\of\datatable
    \edef\mymedian{\pgfplotsretval}

    \pgfplotstablegetelem{#1}{q3}\of\datatable
    \edef\myupperq{\pgfplotsretval}

    \pgfplotstablegetelem{#1}{max}\of\datatable
    \edef\mymax{\pgfplotsretval}

    \typeout{\mymin,\mylowerq,\mymedian,\myupperq,\mymax}
    \edef\temp{\noexpand\addplot+[,
        boxplot prepared={
            lower whisker=\mymin,
            upper whisker=\mymax,
            lower quartile=\mylowerq,
            upper quartile=\myupperq,
            median=\mymedian,
            every box/.style={solid,fill,opacity=0.5},
            every whisker/.style={solid },
            every median/.style={solid},
        }, 
        ]coordinates {};}
    \temp
}
\end{axis}

\end{tikzpicture}}
\caption{\label{fig:trajORB} The vertical axis shows the different sequences from our CARLA dataset in which we have tested our model, and the horizontal axes show the ATE [m] obtained by ORB-SLAM. 
\protect\subref{fig:BP_ORB_In} shows the ORB-SLAM absolute trajectory RMSE [m] for the raw dynamic images.
\protect\subref{fig:BP_Dyna} shows the ATE computed by DynaSLAM~\cite{bescos2018dynaslam}: this system is based on ORB-SLAM and computes the dynamic objects masks in every frame not to use features belonging to them.
\protect\subref{fig:BP_ORB_Ours_Feat} shows the ORB-SLAM ATE when using our inpainted frames (the ones obtained with the ORB-features-based loss), and \protect\subref{fig:BP_ORB_Gt} shows the ORB-SLAM ATE for the ground-truth static images.}
\end{figure*}
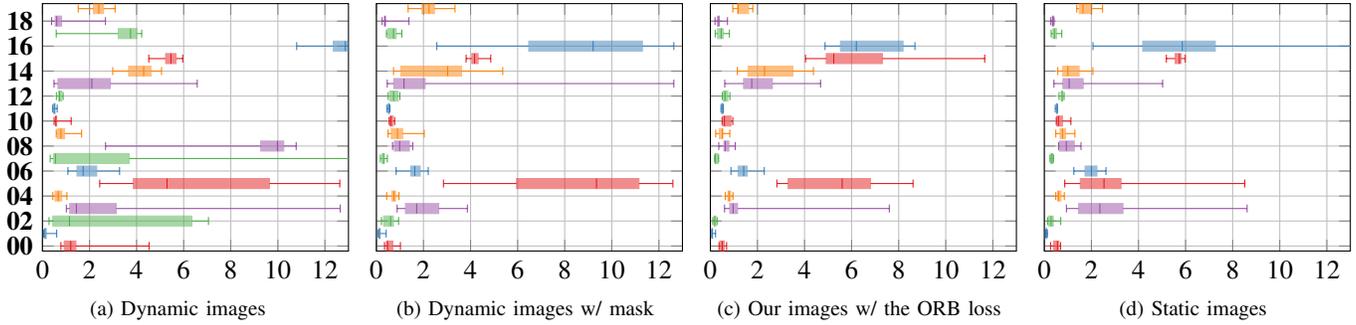


\begin{figure*} [h]
\centering
\subfloat[\label{fig:BP_Ours_wo_ORB}Our images w/o the ORB loss]{\pgfplotstableread{%
sequence min q1 q2 q3 max 
1 0.301 0.423 0.508 0.605 0.891
3 0.044 0.068 0.080 0.090 0.142
5 0.157 0.207 0.261 0.339 0.423
6 0.876 0.925 0.967 1.570 4.747
8 0.526 0.682 0.783 0.955 1.026
10 0.795 1.582 1.896 2.111 2.340
11 0.256 0.278 0.309 0.361 0.416
12 0.558 0.714 0.915 1.070 1.268
13 0.313 0.430 0.452 0.516 0.690
14 0.437 0.521 0.560 0.606 0.720
15 0.449 0.497 0.530 0.568 0.663
17 0.459 0.510 0.611 0.726 0.840
23 0.327 0.371 0.472 0.529 0.975
24 0.237 0.299 0.346 0.372 1.269
25 1.021 1.335 1.445 1.863 1.995
}\datatable

\begin{tikzpicture}
\pgfplotstablegetrowsof{\datatable}
\pgfmathtruncatemacro{\rownumber}{\pgfplotsretval - 1}
\begin{axis}[
width=6.0cm,
enlargelimits = false,
grid = major,
boxplot/draw direction=x,
yticklabels = {00,01,02,03,04,06,07,08,09,10,11,12,17,18,19},
ytick = {1,2,...,\the\numexpr\rownumber + 1},
y tick label style = {scale = 0.9, font = \bfseries, rotate = 0,align = center},
xticklabels = {0,0.5,1,1.5,2},
xtick = {0,0.5,1,1.5,2},
x tick label style = {scale = 1.0, rotate = 0,align = center},
xmin=0, xmax=2,
cycle list/Set1-5 ]

\pgfplotsinvokeforeach{0,...,\rownumber}{

    \pgfplotstablegetelem{#1}{min}\of\datatable
    \edef\mymin{\pgfplotsretval}

    \pgfplotstablegetelem{#1}{q1}\of\datatable
    \edef\mylowerq{\pgfplotsretval}

    \pgfplotstablegetelem{#1}{q2}\of\datatable
    \edef\mymedian{\pgfplotsretval}

    \pgfplotstablegetelem{#1}{q3}\of\datatable
    \edef\myupperq{\pgfplotsretval}

    \pgfplotstablegetelem{#1}{max}\of\datatable
    \edef\mymax{\pgfplotsretval}

    \typeout{\mymin,\mylowerq,\mymedian,\myupperq,\mymax}
    \edef\temp{\noexpand\addplot+[,
        boxplot prepared={
            lower whisker=\mymin,
            upper whisker=\mymax,
            lower quartile=\mylowerq,
            upper quartile=\myupperq,
            median=\mymedian,
            every box/.style={solid,fill,opacity=0.5},
            every whisker/.style={solid },
            every median/.style={solid},
        }, 
        ]coordinates {};}
    \temp
}
\end{axis}
\end{tikzpicture}}
\subfloat[\label{fig:BP_Ours_w_ORB}Our images w/ the ORB loss]{\pgfplotstableread{%
sequence min q1 q2 q3 max 
1 0.368 0.402 0.504 0.590 0.691
3 0.059 0.063 0.074 0.095 0.221
5 0.104 0.164 0.185 0.234 0.287
6 0.603 0.823 0.966 1.153 7.601
8 0.634 0.748 0.797 0.864 0.960
10 0.874 1.180 1.417 1.566 2.290
11 0.179 0.210 0.214 0.326 0.352
12 0.357 0.577 0.632 0.785 1.056
13 0.226 0.381 0.512 0.527 0.826
14 0.499 0.524 0.598 0.874 0.953
15 0.449 0.468 0.522 0.551 0.552
17 0.505 0.548 0.622 0.739 0.830
23 0.205 0.306 0.459 0.542 0.808
24 0.198 0.294 0.355 0.380 0.727
25 0.953 1.153 1.182 1.620 1.803
}\datatable

\begin{tikzpicture}
\pgfplotstablegetrowsof{\datatable}
\pgfmathtruncatemacro{\rownumber}{\pgfplotsretval - 1}
\begin{axis}[
width=6.0cm,
enlargelimits = false,
grid = major,
boxplot/draw direction=x,
yticklabels = {,,},
ytick = {1,2,...,\the\numexpr\rownumber + 1},
xticklabels = {0,0.5,1,1.5,2},
xtick = {0,0.5,1,1.5,2},
x tick label style = {scale = 1.0, rotate = 0,align = center},
xmin=0, xmax=2,
cycle list/Set1-5 ]

\pgfplotsinvokeforeach{0,...,\rownumber}{

    \pgfplotstablegetelem{#1}{min}\of\datatable
    \edef\mymin{\pgfplotsretval}

    \pgfplotstablegetelem{#1}{q1}\of\datatable
    \edef\mylowerq{\pgfplotsretval}

    \pgfplotstablegetelem{#1}{q2}\of\datatable
    \edef\mymedian{\pgfplotsretval}

    \pgfplotstablegetelem{#1}{q3}\of\datatable
    \edef\myupperq{\pgfplotsretval}

    \pgfplotstablegetelem{#1}{max}\of\datatable
    \edef\mymax{\pgfplotsretval}

    \typeout{\mymin,\mylowerq,\mymedian,\myupperq,\mymax}
    \edef\temp{\noexpand\addplot+[,
        boxplot prepared={
            lower whisker=\mymin,
            upper whisker=\mymax,
            lower quartile=\mylowerq,
            upper quartile=\myupperq,
            median=\mymedian,
            every box/.style={solid,fill,opacity=0.5},
            every whisker/.style={solid },
            every median/.style={solid},
        }, 
        ]coordinates {};}
    \temp
}
\end{axis}
\end{tikzpicture}}
\subfloat[\label{fig:BP_Ours_w_ORB_w_mask}Our images w/ the ORB loss w/ mask]{\pgfplotstableread{%
sequence min q1 q2 q3 max 
1 0.441 0.517 0.563 0.627 0.750
3 0.050 0.071 0.079 0.102 0.225
5 0.173 0.219 0.276 0.341 0.528
6 0.863 1.073 1.367 1.664 3.288
8 0.598 0.746 0.767 0.843 0.945
10 0.907 1.072 1.468 1.746 1.965
11 0.248 0.281 0.290 0.303 0.349
12 0.446 0.539 0.572 0.607 0.749
13 0.355 0.411 0.480 0.620 1.241
14 0.486 0.617 0.686 0.845 1.173
15 0.476 0.513 0.533 0.560 0.570
17 0.468 0.523 0.568 0.682 0.891
23 0.471 0.567 0.693 0.790 0.905
24 0.187 0.283 0.376 0.413 0.932
25 0.511 1.123 1.310 1.424 3.014
}\datatable

\begin{tikzpicture}
\pgfplotstablegetrowsof{\datatable}
\pgfmathtruncatemacro{\rownumber}{\pgfplotsretval - 1}
\begin{axis}[
width=6.0cm,
enlargelimits = false,
grid = major,
boxplot/draw direction=x,
yticklabels = {94\%,87\%,91\%,66\%,92\%,32\%,86\%,60\%,40\%,94\%,60\%,28\%,43\%,93\%,73\%},
ytick = {1,2,...,\the\numexpr\rownumber + 1},
xticklabels = {0,0.5,1,1.5,2},
xtick = {0,0.5,1,1.5,2},
x tick label style = {scale = 1.0, rotate = 0,align = center},
xmin=0, xmax=2,
cycle list/Set1-5,
yticklabel pos = right,
y tick label style = {scale = 0.8}]

\pgfplotsinvokeforeach{0,...,\rownumber}{

    \pgfplotstablegetelem{#1}{min}\of\datatable
    \edef\mymin{\pgfplotsretval}

    \pgfplotstablegetelem{#1}{q1}\of\datatable
    \edef\mylowerq{\pgfplotsretval}

    \pgfplotstablegetelem{#1}{q2}\of\datatable
    \edef\mymedian{\pgfplotsretval}

    \pgfplotstablegetelem{#1}{q3}\of\datatable
    \edef\myupperq{\pgfplotsretval}

    \pgfplotstablegetelem{#1}{max}\of\datatable
    \edef\mymax{\pgfplotsretval}

    \typeout{\mymin,\mylowerq,\mymedian,\myupperq,\mymax}
    \edef\temp{\noexpand\addplot+[,
        boxplot prepared={
            lower whisker=\mymin,
            upper whisker=\mymax,
            lower quartile=\mylowerq,
            upper quartile=\myupperq,
            median=\mymedian,
            every box/.style={solid,fill,opacity=0.5},
            every whisker/.style={solid },
            every median/.style={solid},
        }, 
        ]coordinates {};}
    \temp
}
\end{axis}
\end{tikzpicture}}
\caption{\label{fig:BP_ORB_Influence} The vertical axis shows the different sequences from our CARLA dataset, and the horizontal axes show the ORB-SLAM ATE [m]. 
\protect\subref{fig:BP_Ours_wo_ORB} and \protect\subref{fig:BP_Ours_w_ORB} show the ORB-SLAM absolute trajectory RMSE [m] for our images obtained with the model trained without and with the ORB loss term respectively.
\protect\subref{fig:BP_Ours_w_ORB_w_mask} shows the ATE computed by DynaSLAM for our images obtained with the model trained with the ORB loss term. 
On the right-most side one can find the percentage of keypoints extracted in the inpainted areas, w.r.t. to the ground-truth keypoints.}
\end{figure*}
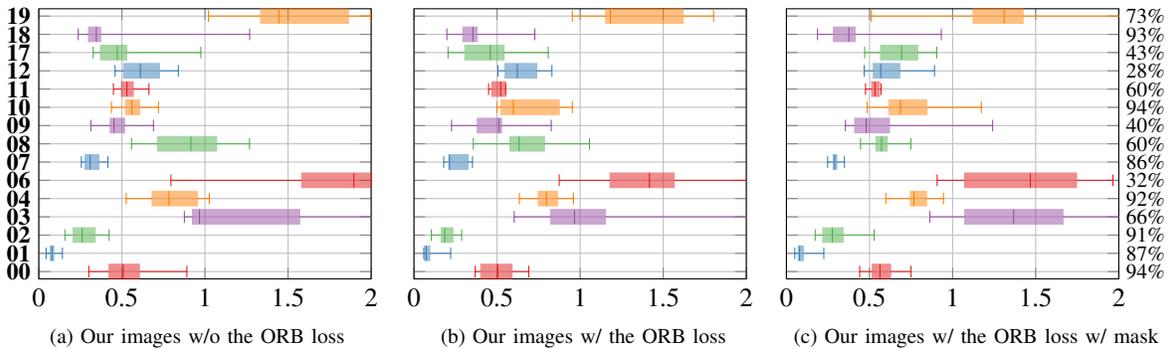

\begin{figure} [t]
\centering
\subfloat{\begin{tikzpicture}
    \node[anchor=south west,inner sep=0] (image) at (0,0) 
    {\adjincludegraphics[trim={{.02\width} {.25\height} {.40\width} {.40\height}}, clip, width=0.32\linewidth]{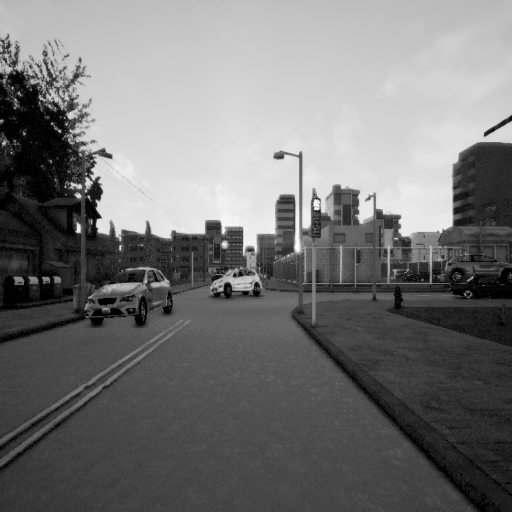}};
\end{tikzpicture}}
\hspace{\fill}
\subfloat{\begin{tikzpicture}
    \node[anchor=south west,inner sep=0] (image) at (0,0) 
    {\adjincludegraphics[trim={{.02\width} {.28\height} {.40\width} {.37\height}}, clip, width=0.32\linewidth]{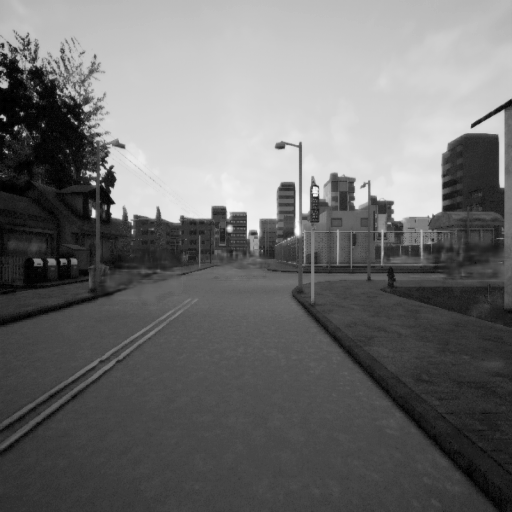}};
\end{tikzpicture}}
\hspace{\fill}
\subfloat{\begin{tikzpicture}
    \node[anchor=south west,inner sep=0] (image) at (0,0) 
    {\adjincludegraphics[trim={{.02\width} {.25\height} {.40\width} {.40\height}}, clip, width=0.32\linewidth]{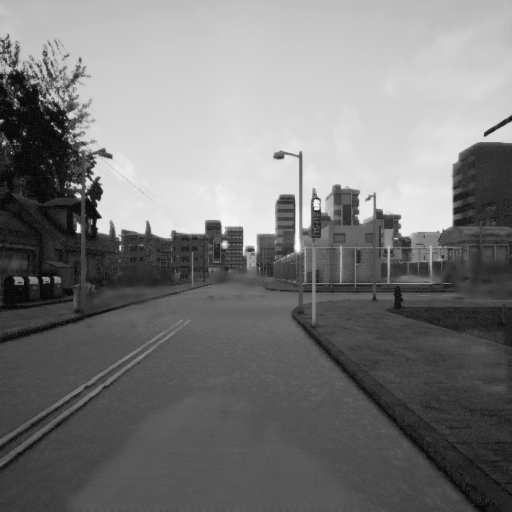}};
\end{tikzpicture}}
\\
\vspace*{-0.6\baselineskip}
\setcounter{subfigure}{0}
\subfloat[\label{fig:inputORB}Input]{\begin{tikzpicture}
    \node[anchor=south west,inner sep=0] (image) at (0,0) 
    {\adjincludegraphics[trim={{.00\width} {.25\height} {.45\width} {.40\height}}, clip, width=0.32\linewidth]{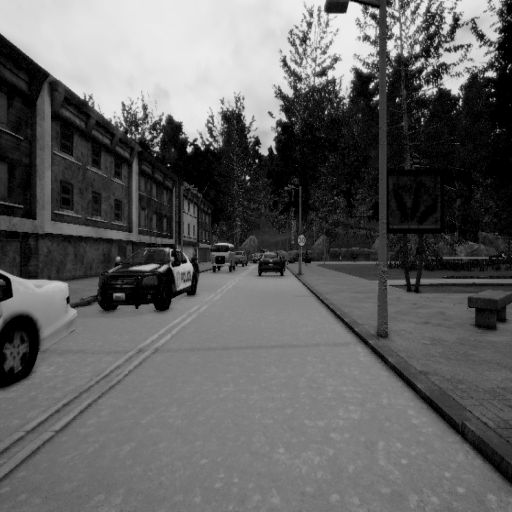}};
\end{tikzpicture}}
\hspace{\fill}
\subfloat[\label{fig:outputwoORB}Output \textbf{w/o} ORB]{\begin{tikzpicture}
    \node[anchor=south west,inner sep=0] (image) at (0,0) 
    {\adjincludegraphics[trim={{.00\width} {.25\height} {.45\width} {.40\height}}, clip, width=0.32\linewidth]{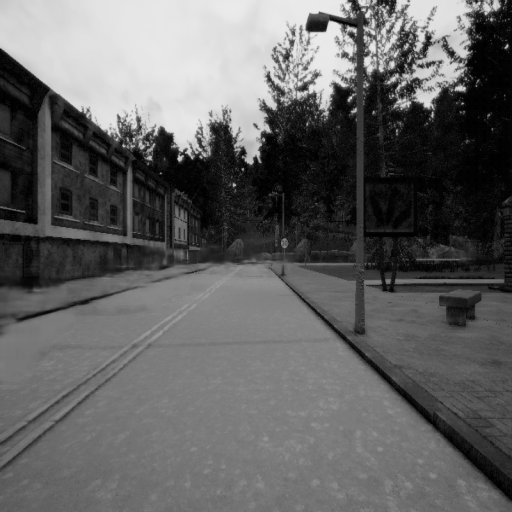}};
\end{tikzpicture}}
\hspace{\fill}
\subfloat[\label{fig:outputwORB}Output \textbf{w/} ORB]{\begin{tikzpicture}
    \node[anchor=south west,inner sep=0] (image) at (0,0) 
    {\adjincludegraphics[trim={{.00\width} {.25\height} {.45\width} {.40\height}}, clip, width=0.32\linewidth]{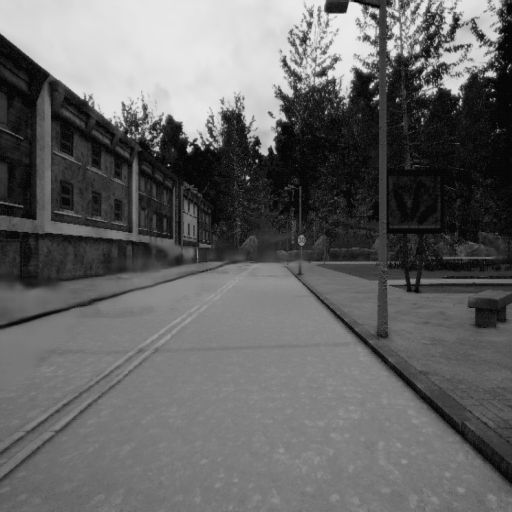}};
\end{tikzpicture}}
\caption{\label{fig:comparisonORB} Visual comparison of the improvements achieved by utilizing the ORB-based loss \protect\subref{fig:outputwORB} againt not using it \protect\subref{fig:outputwoORB}. 
{\color{black} The reconstructed curbs in \protect\subref{fig:outputwORB} are sharper and more straight than those in \protect\subref{fig:outputwoORB}}}
\end{figure}


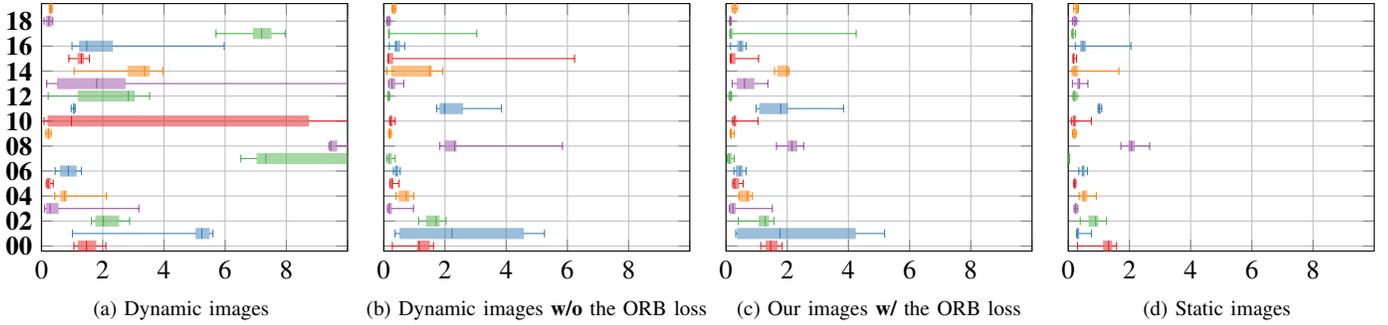
\begin{figure*} [h]
\centering
\subfloat[\label{fig:LDSO_Dyn}Dynamic images]{\pgfplotstableread{%
sequence min q1 q2 q3 max 
1 1.058 1.204 1.470 1.770 2.103
3 1.010 5.054 5.238 5.476 5.602
5 1.631 1.778 2.026 2.520 2.879
6 0.101 0.171 0.281 0.541 3.181
8 0.434 0.633 0.749 0.815 2.125
9 0.156 0.184 0.217 0.283 0.384
10 0.445 0.624 0.877 1.129 1.304
11 6.515 7.045 7.329 9.974 12.955
12 9.394 9.420 9.458 9.645 9.996
13 0.144 0.198 0.225 0.262 0.317
14 0.082 0.215 0.980 8.717 17.098
15 0.970 1.045 1.051 1.076 1.102
17 0.219 1.201 2.835 3.032 3.533
18 0.172 0.527 1.803 2.727 17.407
19 1.066 2.823 3.367 3.513 3.965
20 0.894 1.198 1.304 1.369 1.565
22 0.997 1.250 1.479 2.316 5.970
23 5.702 6.928 7.193 7.496 7.963
24 0.085 0.168 0.242 0.319 0.367
25 0.258 0.272 0.299 0.339 0.347
}\datatable

\begin{tikzpicture}
\pgfplotstablegetrowsof{\datatable}
\pgfmathtruncatemacro{\rownumber}{\pgfplotsretval - 1}
\begin{axis}[
width=5.65cm,
enlargelimits = false,
boxplot/draw direction = x,
grid = major,
yticklabels = {00,02,04,06,08,10,12,14,16,18},
ytick = {1,3,...,\the\numexpr\rownumber + 1},
y tick label style = {scale = 1.0, font = \bfseries, rotate = 0,align = center},
xticklabels = {0,2,...,8},
xtick = {0,2,...,14},
x tick label style = {scale = 1.0, rotate = 0,align = center},
xmin=0, xmax=10,
cycle list/Set1-5 ]

\pgfplotsinvokeforeach{0,...,\rownumber}{

    \pgfplotstablegetelem{#1}{min}\of\datatable
    \edef\mymin{\pgfplotsretval}

    \pgfplotstablegetelem{#1}{q1}\of\datatable
    \edef\mylowerq{\pgfplotsretval}

    \pgfplotstablegetelem{#1}{q2}\of\datatable
    \edef\mymedian{\pgfplotsretval}

    \pgfplotstablegetelem{#1}{q3}\of\datatable
    \edef\myupperq{\pgfplotsretval}

    \pgfplotstablegetelem{#1}{max}\of\datatable
    \edef\mymax{\pgfplotsretval}

    \typeout{\mymin,\mylowerq,\mymedian,\myupperq,\mymax}
    \edef\temp{\noexpand\addplot+[,
        boxplot prepared={
            lower whisker=\mymin,
            upper whisker=\mymax,
            lower quartile=\mylowerq,
            upper quartile=\myupperq,
            median=\mymedian,
            every box/.style={solid,fill,opacity=0.5},
            every whisker/.style={solid },
            every median/.style={solid},
        }, 
        ]coordinates {};}
    \temp
}
\end{axis}
\end{tikzpicture}}
\subfloat[\label{fig:LDSO_Ours}Dynamic images \textbf{w/o} the ORB loss]{\pgfplotstableread{%
sequence min q1 q2 q3 max 
1 0.279 1.108 1.157 1.487 1.624
3 0.369 0.532 2.232 4.557 5.251
5 1.143 1.398 1.709 1.808 2.038
6 0.121 0.152 0.185 0.251 0.971
8 0.405 0.505 0.736 0.826 0.985
9 0.197 0.212 0.274 0.293 0.501
10 0.316 0.383 0.410 0.464 0.537
11 0.098 0.154 0.174 0.248 0.374
12 1.832 2.013 2.319 2.360 5.839
13 0.157 0.173 0.204 0.212 0.247
14 0.189 0.213 0.227 0.277 0.373
15 1.731 1.840 1.983 2.572 3.846
17 0.140 0.148 0.154 0.170 0.196
18 0.156 0.183 0.270 0.360 0.655
19 0.105 0.255 1.501 1.563 1.929
20 0.135 0.142 0.153 0.287 6.236
22 0.180 0.365 0.405 0.520 0.685
23 0.156 0.170 0.173 0.183 3.040
24 0.102 0.122 0.154 0.188 0.213
25 0.272 0.302 0.318 0.360 0.392
}\datatable

\begin{tikzpicture}
\pgfplotstablegetrowsof{\datatable}
\pgfmathtruncatemacro{\rownumber}{\pgfplotsretval - 1}
\begin{axis}[
width=5.65cm,
enlargelimits = false,
grid = major,
boxplot/draw direction=x,
yticklabels = {,,},
ytick = {1,3,...,\the\numexpr\rownumber + 1},
xticklabels = {0,2,...,8},
xtick = {0,2,...,16},
x tick label style = {scale = 1.0, rotate = 0,align = center},
xmin=0, xmax=10,
cycle list/Set1-5 ]

\pgfplotsinvokeforeach{0,...,\rownumber}{

    \pgfplotstablegetelem{#1}{min}\of\datatable
    \edef\mymin{\pgfplotsretval}

    \pgfplotstablegetelem{#1}{q1}\of\datatable
    \edef\mylowerq{\pgfplotsretval}

    \pgfplotstablegetelem{#1}{q2}\of\datatable
    \edef\mymedian{\pgfplotsretval}

    \pgfplotstablegetelem{#1}{q3}\of\datatable
    \edef\myupperq{\pgfplotsretval}

    \pgfplotstablegetelem{#1}{max}\of\datatable
    \edef\mymax{\pgfplotsretval}

    \typeout{\mymin,\mylowerq,\mymedian,\myupperq,\mymax}
    \edef\temp{\noexpand\addplot+[,
        boxplot prepared={
            lower whisker=\mymin,
            upper whisker=\mymax,
            lower quartile=\mylowerq,
            upper quartile=\myupperq,
            median=\mymedian,
            every box/.style={solid,fill,opacity=0.5},
            every whisker/.style={solid },
            every median/.style={solid},
        }, 
        ]coordinates {};}
    \temp
}
\end{axis}
\end{tikzpicture}}
\subfloat[\label{fig:LDSO_OursFeat}Our images \textbf{w/} the ORB loss]{\pgfplotstableread{%
sequence min q1 q2 q3 max 
1 1.136 1.319 1.460 1.672 1.836
3 0.318 0.378 1.765 4.222 5.182
5 0.401 1.087 1.283 1.391 1.566
6 0.110 0.139 0.225 0.299 1.515
8 0.419 0.441 0.699 0.772 0.862
9 0.218 0.240 0.285 0.413 0.562
10 0.258 0.332 0.452 0.527 0.657
11 0.044 0.054 0.101 0.172 0.269
12 1.650 2.038 2.157 2.312 2.540
13 0.140 0.157 0.166 0.182 0.268
14 0.212 0.252 0.277 0.319 1.046
15 0.982 1.101 1.789 2.006 3.840
17 0.103 0.122 0.154 0.172 0.186
18 0.200 0.369 0.604 0.907 1.368
19 1.585 1.699 1.985 2.017 2.063
20 0.139 0.144 0.177 0.291 1.065
22 0.148 0.385 0.481 0.550 0.655
23 0.125 0.136 0.153 0.206 4.254
24 0.120 0.137 0.145 0.158 0.167
25 0.199 0.248 0.291 0.307 0.360
}\datatable

\begin{tikzpicture}
\pgfplotstablegetrowsof{\datatable}
\pgfmathtruncatemacro{\rownumber}{\pgfplotsretval - 1}
\begin{axis}[
width=5.65cm,
enlargelimits = false,
grid = major,
boxplot/draw direction=x,
yticklabels = {,,},
ytick = {1,3,...,\the\numexpr\rownumber + 1},
xticklabels = {0,2,...,8},
xtick = {0,2,...,16},
x tick label style = {scale = 1.0, rotate = 0,align = center},
xmin=0, xmax=10,
cycle list/Set1-5 ]

\pgfplotsinvokeforeach{0,...,\rownumber}{

    \pgfplotstablegetelem{#1}{min}\of\datatable
    \edef\mymin{\pgfplotsretval}

    \pgfplotstablegetelem{#1}{q1}\of\datatable
    \edef\mylowerq{\pgfplotsretval}

    \pgfplotstablegetelem{#1}{q2}\of\datatable
    \edef\mymedian{\pgfplotsretval}

    \pgfplotstablegetelem{#1}{q3}\of\datatable
    \edef\myupperq{\pgfplotsretval}

    \pgfplotstablegetelem{#1}{max}\of\datatable
    \edef\mymax{\pgfplotsretval}

    \typeout{\mymin,\mylowerq,\mymedian,\myupperq,\mymax}
    \edef\temp{\noexpand\addplot+[,
        boxplot prepared={
            lower whisker=\mymin,
            upper whisker=\mymax,
            lower quartile=\mylowerq,
            upper quartile=\myupperq,
            median=\mymedian,
            every box/.style={solid,fill,opacity=0.5},
            every whisker/.style={solid },
            every median/.style={solid},
        }, 
        ]coordinates {};}
    \temp
}
\end{axis}
\end{tikzpicture}}
\subfloat[\label{fig:LDSO_Stat}Static images]{\pgfplotstableread{%
sequence min q1 q2 q3 max 
1 0.305 1.158 1.320 1.416 1.579
3 0.271 0.284 0.310 0.343 0.759
5 0.392 0.679 0.896 0.964 1.251
6 0.191 0.226 0.241 0.279 0.316
8 0.353 0.451 0.514 0.610 0.918
9 0.180 0.201 0.211 0.220 0.253
10 0.346 0.454 0.476 0.522 0.629
11 0.029 0.031 0.035 0.036 0.041
12 1.720 1.985 2.064 2.155 2.661
13 0.140 0.178 0.196 0.224 0.259
14 0.101 0.157 0.194 0.246 0.757
15 0.974 0.993 1.015 1.042 1.095
17 0.158 0.167 0.196 0.236 0.286
18 0.136 0.298 0.359 0.376 0.644
19 0.127 0.158 0.221 0.312 1.657
20 0.158 0.162 0.171 0.196 0.272
22 0.231 0.404 0.482 0.566 2.050
23 0.117 0.137 0.152 0.170 0.240
24 0.139 0.182 0.219 0.228 0.280
25 0.180 0.243 0.287 0.309 0.331
}\datatable

\begin{tikzpicture}
\pgfplotstablegetrowsof{\datatable}
\pgfmathtruncatemacro{\rownumber}{\pgfplotsretval - 1}
\begin{axis}[
width=5.65cm,
enlargelimits = false,
grid = major,
boxplot/draw direction=x,
yticklabels = {,,},
ytick = {1,3,...,\the\numexpr\rownumber + 1},
xticklabels = {0,2,...,8},
xtick = {0,2,...,8},
x tick label style = {scale = 1.0, rotate = 0,align = center},
xmin=0, xmax=10,
cycle list/Set1-5 ]

\pgfplotsinvokeforeach{0,...,\rownumber}{

     \pgfplotstablegetelem{#1}{min}\of\datatable
    \edef\mymin{\pgfplotsretval}

    \pgfplotstablegetelem{#1}{q1}\of\datatable
    \edef\mylowerq{\pgfplotsretval}

    \pgfplotstablegetelem{#1}{q2}\of\datatable
    \edef\mymedian{\pgfplotsretval}

    \pgfplotstablegetelem{#1}{q3}\of\datatable
    \edef\myupperq{\pgfplotsretval}

    \pgfplotstablegetelem{#1}{max}\of\datatable
    \edef\mymax{\pgfplotsretval}

    \typeout{\mymin,\mylowerq,\mymedian,\myupperq,\mymax}
    \edef\temp{\noexpand\addplot+[,
        boxplot prepared={
            lower whisker=\mymin,
            upper whisker=\mymax,
            lower quartile=\mylowerq,
            upper quartile=\myupperq,
            median=\mymedian,
            every box/.style={solid,fill,opacity=0.5},
            every whisker/.style={solid },
            every median/.style={solid},
        }, 
        ]coordinates {};}
    \temp
}
\end{axis}
\end{tikzpicture}}
\caption{\label{fig:trajLDSO} The vertical axis shows the different sequences from our CARLA dataset, and the horizontal axes show the DSO ATE [m]. 
We have computed the boxplots' minimum, maximum and quartiles with the results from 10 repetitions of every test.
\protect\subref{fig:LDSO_Dyn} shows the DSO absolute trajectory RMSE [m] for the raw dynamic images. 
\protect\subref{fig:LDSO_Ours} and \protect\subref{fig:LDSO_OursFeat} show the DSO trajectory errors when using our inpainted frames without and with the ORB-features-based loss respectively. 
Eventually, \protect\subref{fig:LDSO_Stat} shows the DSO trajectory errors for the ground-truth static images.}
\end{figure*}

Fig.~\ref{fig:trajORB} displays the ORB-SLAM absolute trajectory RMSE~[m] computed for 20 CARLA sequences of approximately $100$ m long without loop closures.
Fig.~\ref{fig:BP_ORB_In} shows the results when many vehicles and pedestrians are moving independently. 
More precisely, the number of vehicles and pedestrians have been set to the maximum allowed by CARLA. 
Fig.~\ref{fig:BP_ORB_Gt} shows the same odometry results for the ground-truth static sequences. 
We can see that dynamic objects have in many sequences a big influence on ORB-SLAM's performance (sequences $02$, $07$, $08$, \textit{etc.}). 
Fig.~\ref{fig:BP_Dyna} shows the trajectory error obtained with our previous system DynaSLAM~\cite{bescos2018dynaslam}. 
This system is based on ORB-SLAM and uses the semantic segmentation network Mask \mbox{R-CNN}~\cite{he2017mask} to detect the moving objects and not extract ORB features within them. 
Even though better odometry results are obtained compared against using the raw dynamic images, this experiment shows that using static images leads to a more accurate camera tracking. 
One reason for this is that dynamic objects might occlude the nearby regions in the scene, which are the most reliable for camera pose estimation. 
Another reason might be that using dynamic objects masks does not yield anymore a homogeneous features distribution within the image. 
ORB-SLAM looks for a uniform distribution of image features. 
Pose optimization could be degraded and drifting could increase if the features do not follow such distribution. 
Finally, Fig.~\ref{fig:BP_ORB_Ours_Feat} shows the ORB-SLAM error when using our inpainted images. 
Our odometry results show that better results are usually obtained when using our inpainted images.
The inpainting is realistic enough to provide the visual odometry system with consistent features that are useful for localization.

{\color{black} We want to highlight the importance of the influence of using the ORB loss during training (Fig.~\ref{fig:BP_ORB_Influence}).
Figs.~\ref{fig:BP_Ours_wo_ORB} and~\ref{fig:BP_Ours_w_ORB} present the ATE obtained by ORB-SLAM with our inpainted images without and with the ORB loss respectively. 
The estimated errors are smaller and more constant when using this loss. 
This performance gain can be due to features in the regions that originally contained dynamic objects, and to more stable features in the static-content regions. 
Fig.~\ref{fig:BP_Ours_w_ORB_w_mask} presents the DynaSLAM ATE with the inpainted images generated with our model trained with this loss term. 
We expect these errors to be very similar to those shown in Fig.~\ref{fig:BP_Dyna}, and slightly bigger than those in Fig.~\ref{fig:BP_Ours_w_ORB}. 
This experiment shows that our model barely damages the static content of the scene, keeping the static features as they used to be. 
It also demonstrates that the hallucinated features are useful to estimate the camera SE3 pose.
To support this claim, on the right-most side of Fig.~\ref{fig:BP_ORB_Influence} one can find the percentage of keypoints extracted in the inpainted areas w.r.t. to the ground-truth keypoints.
We show in Fig.~\ref{fig:comparisonORB} two examples of the visual influence of such loss on enhancing high frequencies inpainted areas.
}



Fig.~\ref{fig:trajLDSO} shows the DSO error for the same 20 CARLA sequences for the different input images with dynamic content (Fig.~\ref{fig:LDSO_Dyn}), without dynamic content (Fig.~\ref{fig:LDSO_Stat}), and the images obtained by our framework without and with the ORB-based loss (Figs.~\ref{fig:LDSO_Ours} and \ref{fig:LDSO_OursFeat}). 
Even though direct systems are more robust to dynamic objects within the scene, utilizing our approach also yields a higher tracking accuracy. 
Despite the fact that our feature-based loss follows the ORB implementation, better results are also obtained with other visual odometry system that do not rely on ORB features. 

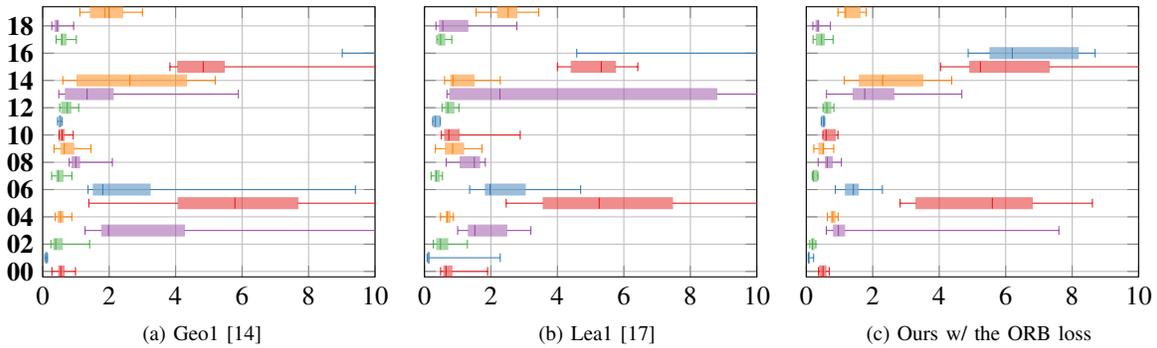
\begin{figure*} [h]
\centering
\subfloat[\label{fig:ORBGeo1}Geo1~\cite{telea2004image}]{\pgfplotstableread{%
sequence min q1 q2 q3 max 
1 0.278 0.483 0.545 0.646 0.987
3 0.065 0.097 0.117 0.125 0.147
5 0.246 0.332 0.394 0.590 1.412
6 1.275 1.776 1.981 4.264 10.665
8 0.376 0.464 0.538 0.609 0.877
9 1.388 4.075 5.784 7.676 10.173
10 1.365 1.524 1.809 3.233 9.409
11 0.269 0.420 0.470 0.608 0.879
12 0.798 0.884 0.997 1.109 2.096
13 0.341 0.549 0.650 0.933 1.452
14 0.490 0.494 0.586 0.645 0.918
15 0.447 0.488 0.532 0.539 0.586
17 0.511 0.571 0.740 0.849 1.083
18 0.485 0.683 1.335 2.122 5.889
19 0.606 1.034 2.623 4.329 5.191
20 3.828 4.067 4.831 5.464 10.401
22 9.014 11.701 13.352 14.829 15.699
23 0.404 0.552 0.578 0.699 1.013
24 0.274 0.375 0.440 0.477 0.931
25 1.118 1.436 1.877 2.404 3.003
}\datatable

\begin{tikzpicture}
\pgfplotstablegetrowsof{\datatable}
\pgfmathtruncatemacro{\rownumber}{\pgfplotsretval - 1}
\begin{axis}[
width=6.0cm,
enlargelimits = false,
grid = major,
boxplot/draw direction=x,
yticklabels = {00,02,04,06,08,10,12,14,16,18},
ytick = {1,3,...,\the\numexpr\rownumber + 1},
y tick label style = {scale = 1.0, font = \bfseries, rotate = 0,align = center},
xticklabels = {0,2,...,12},
xtick = {0,2,...,12},
x tick label style = {scale = 1.0, rotate = 0,align = center},
xmin=0, xmax=10,
cycle list/Set1-5 ]

\pgfplotsinvokeforeach{0,...,\rownumber}{

    \pgfplotstablegetelem{#1}{min}\of\datatable
    \edef\mymin{\pgfplotsretval}

    \pgfplotstablegetelem{#1}{q1}\of\datatable
    \edef\mylowerq{\pgfplotsretval}

    \pgfplotstablegetelem{#1}{q2}\of\datatable
    \edef\mymedian{\pgfplotsretval}

    \pgfplotstablegetelem{#1}{q3}\of\datatable
    \edef\myupperq{\pgfplotsretval}

    \pgfplotstablegetelem{#1}{max}\of\datatable
    \edef\mymax{\pgfplotsretval}

    \typeout{\mymin,\mylowerq,\mymedian,\myupperq,\mymax}
    \edef\temp{\noexpand\addplot+[,
        boxplot prepared={
            lower whisker=\mymin,
            upper whisker=\mymax,
            lower quartile=\mylowerq,
            upper quartile=\myupperq,
            median=\mymedian,
            every box/.style={solid,fill,opacity=0.5},
            every whisker/.style={solid },
            every median/.style={solid},
        }, 
        ]coordinates {};}
    \temp
}
\end{axis}
\end{tikzpicture}}
\subfloat[\label{fig:ORBLea1}Lea1~\cite{yu2018generative}]{\pgfplotstableread{%
sequence min q1 q2 q3 max 
1 0.486 0.584 0.648 0.823 1.899
3 0.081 0.113 0.130 0.137 2.279
5 0.265 0.369 0.480 0.699 1.291
6 0.999 1.323 1.516 2.475 3.201
8 0.482 0.653 0.686 0.769 0.872
9 2.458 3.573 5.258 7.462 10.357
10 1.359 1.834 1.972 3.031 4.698
11 0.199 0.316 0.355 0.442 0.542
12 0.653 1.075 1.499 1.658 1.830
13 0.328 0.629 0.847 1.176 1.729
14 0.508 0.610 0.733 1.042 2.881
15 0.245 0.269 0.328 0.460 0.479
17 0.529 0.636 0.706 0.883 1.037
18 0.677 0.766 2.270 8.794 10.983
19 0.604 0.790 0.856 1.489 2.282
20 4.000 4.416 5.321 5.738 6.419
22 4.581 10.941 13.363 14.194 17.014
23 0.370 0.413 0.486 0.609 0.822
24 0.345 0.455 0.553 1.301 2.780
25 1.554 2.203 2.517 2.780 3.438
}\datatable

\begin{tikzpicture}
\pgfplotstablegetrowsof{\datatable}
\pgfmathtruncatemacro{\rownumber}{\pgfplotsretval - 1}
\begin{axis}[
width=6.0cm,
enlargelimits = false,
grid = major,
boxplot/draw direction=x,
yticklabels = {,,},
ytick = {1,3,...,\the\numexpr\rownumber + 1},
xticklabels = {0,2,...,12},
xtick = {0,2,...,12},
x tick label style = {scale = 1.0, rotate = 0,align = center},
xmin=0, xmax=10,
cycle list/Set1-5 ]

\pgfplotsinvokeforeach{0,...,\rownumber}{

    \pgfplotstablegetelem{#1}{min}\of\datatable
    \edef\mymin{\pgfplotsretval}

    \pgfplotstablegetelem{#1}{q1}\of\datatable
    \edef\mylowerq{\pgfplotsretval}

    \pgfplotstablegetelem{#1}{q2}\of\datatable
    \edef\mymedian{\pgfplotsretval}

    \pgfplotstablegetelem{#1}{q3}\of\datatable
    \edef\myupperq{\pgfplotsretval}

    \pgfplotstablegetelem{#1}{max}\of\datatable
    \edef\mymax{\pgfplotsretval}

    \typeout{\mymin,\mylowerq,\mymedian,\myupperq,\mymax}
    \edef\temp{\noexpand\addplot+[,
        boxplot prepared={
            lower whisker=\mymin,
            upper whisker=\mymax,
            lower quartile=\mylowerq,
            upper quartile=\myupperq,
            median=\mymedian,
            every box/.style={solid,fill,opacity=0.5},
            every whisker/.style={solid },
            every median/.style={solid},
        }, 
        ]coordinates {};}
    \temp
}
\end{axis}
\end{tikzpicture}}
\subfloat[\label{fig:ORBOurs}Ours w/ the ORB loss]{\pgfplotstableread{%
sequence min q1 q2 q3 max 
1 0.368 0.402 0.504 0.590 0.691
3 0.059 0.063 0.074 0.095 0.221
5 0.104 0.164 0.185 0.234 0.287
6 0.603 0.823 0.966 1.153 7.601
8 0.634 0.748 0.797 0.864 0.960
9 2.824 3.304 5.599 6.800 8.610
10 0.874 1.180 1.417 1.566 2.290
11 0.179 0.210 0.214 0.326 0.352
12 0.357 0.577 0.632 0.785 1.056
13 0.226 0.381 0.512 0.527 0.826
14 0.499 0.524 0.598 0.874 0.953
15 0.449 0.468 0.522 0.551 0.552
17 0.505 0.548 0.622 0.739 0.830
18 0.610 1.411 1.759 2.634 4.681
19 1.144 1.600 2.299 3.509 4.377
20 4.039 4.923 5.241 7.306 11.654
22 4.871 5.528 6.201 8.184 8.691
23 0.205 0.306 0.459 0.542 0.808
24 0.198 0.294 0.355 0.380 0.727
25 0.953 1.153 1.182 1.620 1.803
}\datatable

\begin{tikzpicture}
\pgfplotstablegetrowsof{\datatable}
\pgfmathtruncatemacro{\rownumber}{\pgfplotsretval - 1}
\begin{axis}[
width=6.0cm,
enlargelimits = false,
grid = major,
boxplot/draw direction=x,
yticklabels = {,,},
ytick = {1,3,...,\the\numexpr\rownumber + 1},
xticklabels = {0,2,...,12},
xtick = {0,2,...,12},
x tick label style = {scale = 1.0, rotate = 0,align = center},
xmin=0, xmax=10,
cycle list/Set1-5]

\pgfplotsinvokeforeach{0,...,\rownumber}{

    \pgfplotstablegetelem{#1}{min}\of\datatable
    \edef\mymin{\pgfplotsretval}

    \pgfplotstablegetelem{#1}{q1}\of\datatable
    \edef\mylowerq{\pgfplotsretval}

    \pgfplotstablegetelem{#1}{q2}\of\datatable
    \edef\mymedian{\pgfplotsretval}

    \pgfplotstablegetelem{#1}{q3}\of\datatable
    \edef\myupperq{\pgfplotsretval}

    \pgfplotstablegetelem{#1}{max}\of\datatable
    \edef\mymax{\pgfplotsretval}

    \typeout{\mymin,\mylowerq,\mymedian,\myupperq,\mymax}
    \edef\temp{\noexpand\addplot+[,
        boxplot prepared={
            lower whisker=\mymin,
            upper whisker=\mymax,
            lower quartile=\mylowerq,
            upper quartile=\myupperq,
            median=\mymedian,
            every box/.style={solid,fill,opacity=0.5},
            every whisker/.style={solid },
            every median/.style={solid},
        }, 
        ]coordinates {};}
    \temp
}
\end{axis}
\end{tikzpicture}}
\caption{\label{fig:ORBOtherInpainting} {\color{black}The vertical axis shows the different sequences from our CARLA dataset, and the horizontal axes show the ATE [m] obtained by ORB-SLAM for 10 repetitions. 
\protect\subref{fig:ORBGeo1} and \protect\subref{fig:ORBLea1} show the absolute trajectory RMSE [m] for the images inpainted with the method by Telea~\cite{telea2004image} and Yu \textit{et~al.}~\cite{yu2018generative} respectively. 
\protect\subref{fig:ORBOurs} shows the ORB-SLAM trajectory error for our images.}}
\end{figure*}

{\color{black}\textbf{Baselines for Inpainting}

We have compared in the section~\ref{sec:image} the quality of our results with respect to the inpainting metrics against four other methods. 
We want to compare now how our approach compares to them w.r.t.  visual odometry metrics. 
Among these four other methods we have chosen two of them for this evaluation: Geo1~\cite{telea2004image}, and Lea1~\cite{yu2018generative}. 
The first choice is motivated by its performance on our inpainting test dataset: this method performs the best among the two non-learning based approaches. 
The second choice is however motivated by the fact that we have not been able to train the model by Iizuka \textit{et~al.} with our training data for a direct comparison.
The evaluation and different results can be seen in Fig.~\ref{fig:ORBOtherInpainting}. 

The images inpainted by Telea's method are usually very smooth and no features are extracted within the inpainted areas. 
The behaviour of ORB-SLAM when using such sequences (Fig.~\ref{fig:ORBGeo1}) is very similar to using DynaSLAM. 
However, the learning-based method by Yu \textit{et~al.} tends to inpaint the images with low frequency patterns found in the image static content, generating many crispy artifacts (see examples in Fig.~\ref{fig:comparisonOxford}). 
A higher ATE is observed when using such images (Fig.~\ref{fig:ORBLea1}). 
Our method seems to be more suitable for the VO task: the inpainting is neither too smooth nor generates crispy artifacts.}

\textbf{Baselines for VO in Real World New Scenarios}

For the evaluation of Empty Cities on real world environments w.r.t. visual odometry we have chosen the KITTI~\cite{geiger2013vision} and the Oxford Robotcar~\cite{RobotCarDatasetIJRR} datasets. 
Dynamic objects in the KITTI dataset do not represent a big inconvenience for camera pose estimation, as was shown in our last work~\cite{bescos2018dynaslam}. 
Most of the vehicles that appear are not moving and lay in nearby scene areas, thus their features happen to be helpful to compute the sensor odometry. 
Also, the few moving pedestrians and cars along the sequences do not represent a big region within the images. 
The Oxford Robotcar dataset has though many sequences with representative moving objects (driving cars), having also sequences with only stationary objects (parked cars). 
Note that, the reported results are not for the whole sequence since the authors provide their VO solution as ground-truth, stating that it is accurate over hundreds of metres. 
Hence, the sequences we use are between 100 and 300 m long.

To perform the KITTI experiment we have re-trained our network with $256 \times 768$ resolution images for a better adaptation. 
The CARLA camera intrinsics have also been modified to match the ones used in KITTI. 
In this case, we have shifted the previously used semantic segmentation model trained only on Cityscapes for the ERFNet model with encoder trained on ImageNet and decoder trained on Cityscapes train set, and have finetuned it with the KITTI semantic segmentation training dataset. 
The generator and discriminator have also been finetuned with such data as explained in subsection~\ref{subsec:domain}.

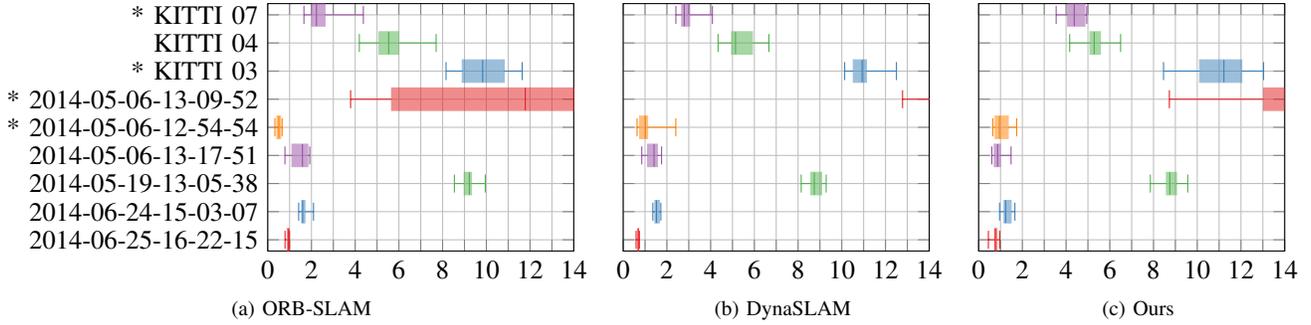
\begin{figure*} [h]
\centering
\subfloat[\label{fig:oxford_in}ORB-SLAM]{\pgfplotstableread{%
sequence min q1 q2 q3 max 
1 0.791 0.890 0.954 0.974 0.995
2 1.416 1.557 1.597 1.699 2.094
3 8.539 8.986 9.224 9.313 9.951
4 0.787 1.108 1.589 1.834 1.922
5 0.327 0.416 0.509 0.594 0.657
6 3.791 5.664 11.782 17.998 29.487
7 8.157 8.899 9.832 10.819 11.640
8 4.190 5.091 5.532 5.983 7.703
9 1.656 2.010 2.229 2.618 4.374
}\datatable

\begin{tikzpicture}
\pgfplotstablegetrowsof{\datatable}
\pgfmathtruncatemacro{\rownumber}{\pgfplotsretval - 1}
\begin{axis}[
width=5.65cm,
enlargelimits = false,
boxplot/draw direction = x,
grid = major,
yticklabels = {2014-06-25-16-22-15,
            2014-06-24-15-03-07,
            2014-05-19-13-05-38, 
            2014-05-06-13-17-51,
            * 2014-05-06-12-54-54,
            * 2014-05-06-13-09-52,
            * KITTI 03,
            KITTI 04,
            * KITTI 07},
ytick = {1,2,3,...,\the\numexpr\rownumber + 1},
xticklabels = {0,,2,,4,,6,,8,,10,,12,,14},
xtick = {0,1,2,...,14},
x tick label style = {scale = 1.0, rotate = 0,align = center},
xmin=0, xmax=14,
cycle list/Set1-5 ]

\pgfplotsinvokeforeach{0,...,\rownumber}{

    \pgfplotstablegetelem{#1}{min}\of\datatable
    \edef\mymin{\pgfplotsretval}

    \pgfplotstablegetelem{#1}{q1}\of\datatable
    \edef\mylowerq{\pgfplotsretval}

    \pgfplotstablegetelem{#1}{q2}\of\datatable
    \edef\mymedian{\pgfplotsretval}

    \pgfplotstablegetelem{#1}{q3}\of\datatable
    \edef\myupperq{\pgfplotsretval}

    \pgfplotstablegetelem{#1}{max}\of\datatable
    \edef\mymax{\pgfplotsretval}

    \typeout{\mymin,\mylowerq,\mymedian,\myupperq,\mymax}
    \edef\temp{\noexpand\addplot+[,
        boxplot prepared={
            lower whisker=\mymin,
            upper whisker=\mymax,
            lower quartile=\mylowerq,
            upper quartile=\myupperq,
            median=\mymedian,
            every box/.style={solid,fill,opacity=0.5},
            every whisker/.style={solid },
            every median/.style={solid},
        }, 
        ]coordinates {};}
    \temp
}
\end{axis}

\end{tikzpicture}}
\subfloat[\label{fig:oxford_mask}DynaSLAM]{\pgfplotstableread{%
sequence min q1 q2 q3 max 
1 0.591 0.675 0.688 0.709 0.740
2 1.357 1.450 1.520 1.652 1.723
3 8.143 8.586 8.742 9.081 9.285
4 0.850 1.118 1.412 1.563 1.756
5 0.635 0.751 1.016 1.118 2.407
6 12.781 22.945 24.652 26.558 33.552
7 10.132 10.526 10.929 11.126 12.500
8 4.340 4.960 5.139 5.914 6.667
9 2.410 2.676 2.793 3.043 4.082
}\datatable

\begin{tikzpicture}
\pgfplotstablegetrowsof{\datatable}
\pgfmathtruncatemacro{\rownumber}{\pgfplotsretval - 1}
\begin{axis}[
width=5.65cm,
enlargelimits = false,
boxplot/draw direction = x,
grid = major,
yticklabels = {,,},
ytick = {1,2,...,\the\numexpr\rownumber + 1},
xticklabels = {0,,2,,4,,6,,8,,10,,12,,14},
xtick = {0,1,2,...,14},
x tick label style = {scale = 1.0, rotate = 0,align = center},
xmin=0, xmax=14,
cycle list/Set1-5 ]

\pgfplotsinvokeforeach{0,...,\rownumber}{

    \pgfplotstablegetelem{#1}{min}\of\datatable
    \edef\mymin{\pgfplotsretval}

    \pgfplotstablegetelem{#1}{q1}\of\datatable
    \edef\mylowerq{\pgfplotsretval}

    \pgfplotstablegetelem{#1}{q2}\of\datatable
    \edef\mymedian{\pgfplotsretval}

    \pgfplotstablegetelem{#1}{q3}\of\datatable
    \edef\myupperq{\pgfplotsretval}

    \pgfplotstablegetelem{#1}{max}\of\datatable
    \edef\mymax{\pgfplotsretval}

    \typeout{\mymin,\mylowerq,\mymedian,\myupperq,\mymax}
    \edef\temp{\noexpand\addplot+[,
        boxplot prepared={
            lower whisker=\mymin,
            upper whisker=\mymax,
            lower quartile=\mylowerq,
            upper quartile=\myupperq,
            median=\mymedian,
            every box/.style={solid,fill,opacity=0.5},
            every whisker/.style={solid },
            every median/.style={solid},
        }, 
        ]coordinates {};}
    \temp
}
\end{axis}

\end{tikzpicture}}
\subfloat[\label{fig:oxford_ours}Ours]{\pgfplotstableread{%
sequence min q1 q2 q3 max 
1 0.445 0.745 0.757 0.832 0.968
2 0.957 1.163 1.234 1.486 1.658
3 7.856 8.591 8.745 9.051 9.566
4 0.604 0.713 0.867 0.990 1.483
5 0.651 0.754 0.948 1.353 1.738
6 8.725 13.032 22.426 31.970 37.080
7 8.458 10.109 11.220 12.037 13.035
8 4.165 5.092 5.287 5.573 6.494
9 3.546 4.071 4.374 4.861 4.947
}\datatable

\begin{tikzpicture}
\pgfplotstablegetrowsof{\datatable}
\pgfmathtruncatemacro{\rownumber}{\pgfplotsretval - 1}
\begin{axis}[
width=5.65cm,
enlargelimits = false,
boxplot/draw direction = x,
grid = major,
yticklabels = {,,},
ytick = {1,2,...,\the\numexpr\rownumber + 1},
xticklabels = {0,,2,,4,,6,,8,,10,,12,,14},
xtick = {0,1,2,...,14},
x tick label style = {scale = 1.0, rotate = 0,align = center},
xmin=0, xmax=14,
cycle list/Set1-5 ]

\pgfplotsinvokeforeach{0,...,\rownumber}{

    \pgfplotstablegetelem{#1}{min}\of\datatable
    \edef\mymin{\pgfplotsretval}

    \pgfplotstablegetelem{#1}{q1}\of\datatable
    \edef\mylowerq{\pgfplotsretval}

    \pgfplotstablegetelem{#1}{q2}\of\datatable
    \edef\mymedian{\pgfplotsretval}

    \pgfplotstablegetelem{#1}{q3}\of\datatable
    \edef\myupperq{\pgfplotsretval}

    \pgfplotstablegetelem{#1}{max}\of\datatable
    \edef\mymax{\pgfplotsretval}

    \typeout{\mymin,\mylowerq,\mymedian,\myupperq,\mymax}
    \edef\temp{\noexpand\addplot+[,
        boxplot prepared={
            lower whisker=\mymin,
            upper whisker=\mymax,
            lower quartile=\mylowerq,
            upper quartile=\myupperq,
            median=\mymedian,
            every box/.style={solid,fill,opacity=0.5},
            every whisker/.style={solid },
            every median/.style={solid},
        }, 
        ]coordinates {};}
    \temp
}
\end{axis}

\end{tikzpicture}}
\caption{\label{fig:oxfordATE} {\color{black} The vertical axis show the KITTI and Oxford Robotcar datasets sequences in which we have tested our model, and the horizontal axes show the boxplots of the absolute trajectory RMSE [m] with the results from 10 repetitions of every test. 
\protect\subref{fig:oxford_in} and \protect\subref{fig:oxford_mask} show the ORB-SLAM and DynaSLAM trajectory errors respectively for the raw dynamic images. 
\protect\subref{fig:oxford_ours} shows ORB-SLAM results when our framework is employed. 
The asterisk at the beginning of some sequences names means that most of the observed vehicles are either not moving or parked.}}
\end{figure*}

Fig.~\ref{fig:oxfordATE} shows the evaluation of our method performance with the sequences from the Oxford Robotcar and the KITTI datasets. 
The asterisk at the beginning of some sequences names means that most of the observed vehicles are either not moving or parked. 
The other sequences though present many moving vehicles. 
In the former type of sequences (*) the highest accuracy should be observed in the case in which the raw images are used (Fig.~\ref{fig:oxford_in}). 
Removing the features from such vehicles as in Fig.~\ref{fig:oxford_mask} leads to a lower accuracy since the most nearby features are no longer used. 
Inpainting the static scene behind these vehicles (Fig.~\ref{fig:oxford_ours}) would still remove nearby features but would create new static features a little bit further. 
That is, the visual odometry accuracy should be lower than in Fig.~\ref{fig:oxford_in} but a little bit higher or similar to Fig.~\ref{fig:oxford_mask}. 
This is the case of our performance on the Oxford Robotcar sequences and the KITTI sequence 03. 
However, DynaSLAM achieves a result in the KITTI sequence $07$ better the proposed Empty Cities. 
After the first half of the sequence, there are a few consecutive frames in which a truck covers almost 75~\% of the image. 
The task of inpainting becomes especially difficult, thus worsening the estimation of the camera's trajectory.
Regarding the performance of the second type of sequences, removing the features from moving vehicles and pedestrians should lead to a lower ATE (Fig.~\ref{fig:oxford_mask} compared to Fig.~\ref{fig:oxford_in}). 
Empty Cities adds in these sequences an important number of features for pose estimation that usually leads to a slightly better trajectory estimation (Fig.~\ref{fig:oxford_ours}). 
Note that the results given in here for the KITTI sequences might not match the ones reported by ORB-SLAM and DynaSLAM respectively because of the utilized images resolution ($256 \times 768$).

\subsection{Visual Place Recognition}
Visual place recognition (VPR) is an important task for visual SLAM. 
Such algorithms are useful when revisiting places to perform loop closure and correct the accumulated drift along long trajectories. 
Bags of visual words (BoW) is the approach that is widely used to perform such task, as can be seen in ORB-SLAM~\cite{mur2017orb} and LDSO~\cite{gao2018ldso}. 
Lately, thanks to the boost of deep learning, learnt global image descriptors are also used for VPR~\cite{Arandjelovic16}. 
In our previous work~\cite{bescos2019empty} we showed preliminary results proving the benefits of our solution for VPR by using descriptors from an off-the-shelf CNN~\cite{olid2018single}. 

\textbf{Baseline for VPR in our Synthetic Dataset}

In this subsection we show a VPR experiment performed with the bag of words work by Mur and Tardos~\cite{mur2014fast, galvez2012bags}. 
It is ideal to test our model since is based on ORB features. 
We also show an experiment with one of the strongest learning-based baselines NetVLAD~\cite{Arandjelovic16}, which is trained for the specific task of VPR. 
This comparison can provide a broader understanding of how and when end-to-end task-specific learning becomes more or less suitable than an explicit use of semantics-based visual description, which forms the primary pitch of this paper.

We have generated two CARLA sequences with loop closures with and without dynamic objects.
Two images are defined as the same place if they are less than 10 meters apart.

\begin{figure*}
    \centering
    \subfloat[\label{fig:PR_BoW}Place recognition results for the BoW work by Mur and Tardos~\cite{mur2014fast, galvez2012bags}.]{
    \begin{tikzpicture}[scale=0.51]
    \begin{axis}[
    xlabel={Recall [\%]},
    xlabel style={font=\Large, at={(0.5,0.1)}},
    ylabel={Precision [\%]},
    ylabel style={font=\Large, at={(-0.1,0.5)}}, 
    xmin=0, xmax=90,
    ymin=30, ymax=100,
    enlarge y limits=false,
    legend pos= south west,
    legend style={font=\small},
    ]
    \addplot[
    color=blue,
    mark=*,
    ]
    coordinates {
       (25.6245, 6.5389) (24.3167, 6.4843) (23.2792, 9.9238) (22.0152, 13.7797) (16.9578, 42.207899999999995) (11.5363, 69.6118) (7.5666, 93.7364) (3.752, 99.9333) (1.916, 100.0) (0.8498, 100.0) 
    };
    
    \addplot[
    color=red,
    mark=otimes,
    ]
    coordinates {
        (25.6245, 3.6457) (23.781, 6.908799999999999) (21.1241, 17.6983) (15.7977, 43.4766) (10.6065, 76.214) (6.8395, 93.053) (4.7344, 98.3619) (2.4166, 100.0) (1.0563, 100.0) (0.3942, 100.0) 
    };
    
    \addplot[
    color=black,
    mark=triangle,
    ]
    coordinates {
        (97.6116, 6.5493999999999994) (84.959, 6.0795) (74.6854, 8.238199999999999) (93.8263, 13.791500000000001) (85.0162, 33.2389) (69.0503, 57.2264) (51.4874, 82.94930000000001) (28.7376, 99.5212) (11.8564, 100.0) (10.8982, 100.0) 
    };
    
    \addplot[
    color=magenta,
    mark=star,
    ]
    coordinates {
        (26.0151, 8.3942) (27.6182, 8.6621) (26.5883, 14.3123) (23.835, 26.837400000000002) (19.2645, 52.2038) (14.5163, 82.2584) (11.0121, 97.71979999999999) (7.0899, 99.9017) (4.9348, 100.0) (3.7233, 100.0) 
    };
    
    \addplot[
    color=green,
    mark=square,
    ]
    coordinates {
        (24.761, 6.5382) (25.622, 4.0411) (19.7525, 8.4307) (23.9124, 18.2816) (20.5171, 36.857) (15.3246, 63.8858) (10.7003, 89.8865) (5.195, 99.9037) (3.0474, 100.0) (1.667, 100.0) 
    };
    
    \addplot[
    color=cyan,
    mark=diamond,
    ]
    coordinates {
        (24.0151, 6.3942) (25.6145, 6.644799999999999) (24.9712, 6.0814) (18.3771, 16.819799999999997) (17.3095, 40.838) (12.749, 73.76) (6.225, 92.06) (3.0824, 99.9189) (2.5043, 100.0) (1.3266, 100.0)
    };
    
    \end{axis}
    \end{tikzpicture}
    \hfill
    \begin{tikzpicture}[scale=0.51]
    \begin{axis}[
    xlabel={Recall [\%]},
    xlabel style={font=\Large, at={(0.0,0.1)}},
    ylabel={Precision [\%]},
    ylabel style={font=\Large, at={(-0.1,0.5)}},
    xmin=10, xmax=55,
    ymin=10, ymax=100,
    enlarge y limits=false,
    legend pos = south west,
    legend style={font=\small},
    ]
    \addplot[
    color=blue,
    mark=*,
    ]
    coordinates {
        (50.8011, 7.3269) (50.7706, 7.349600000000001) (49.128, 7.0578) (45.2849, 20.9102) (38.9069, 43.4591) (30.359, 75.1695) (22.4498, 93.9728) (12.8777, 99.76429999999999) (7.828, 100.0) (4.8976, 100.0)
    };
    \addlegendentry{Dynamic}
    
    \addplot[
    color=red,
    mark=otimes,
    ]
    coordinates {
        (50.8011, 1.8966) (50.7199, 7.5245999999999995) (47.5056, 15.6011) (41.8779, 28.0304) (33.5023, 48.2054) (24.853, 87.8495) (18.4851, 97.4345) (10.9714, 99.9077) (6.6721, 100.0) (4.1979, 100.0) 
    };
    \addlegendentry{With mask}
    
    \addplot[
    color=black,
    mark=triangle,
    ]
    coordinates {
        (50.8011, 6.830799999999999) (50.8011, 3.9915) (49.7566, 11.350399999999999) (47.3839, 21.0495) (42.8716, 42.884699999999995) (35.5607, 72.6388) (28.26, 87.8625) (17.39, 99.8835) (11.6305, 100.0) (7.7063, 100.0) 
    };
    \addlegendentry{Static}
    
    \addplot[
    color=magenta,
    mark=star,
    ]
    coordinates {
        (50.8011, 4.0417000000000005) (50.8011, 6.518300000000001) (49.4322, 7.907400000000001) (46.3598, 16.0387) (41.1478, 40.3741) (33.9079, 74.16279999999999) (27.0432, 94.70880000000001) (17.0554, 100.0) (11.4987, 100.0) (7.9497, 100.0) 
    };
    \addlegendentry{Ours}
    
    \addplot[
    color=green,
    mark=square,
    ]
    coordinates {
        (50.8011, 7.3269) (50.7909, 6.997000000000001) (49.4727, 7.045800000000001) (46.3496, 20.4702) (40.6814, 41.9008) (33.2488, 73.4378) (25.36, 92.974) (14.8651, 99.6601) (9.349, 100.0) (6.013, 100.0) 
    };
    \addlegendentry{Geo1}
    
    \addplot[
    color=cyan,
    mark=diamond,
    ]
    coordinates {
        (50.8011, 7.5159) (50.8011, 7.2283) (49.1685, 11.067499999999999) (46.0556, 13.597999999999999) (40.144, 40.2788) (32.0422, 71.7367) (24.2344, 92.8155) (13.9728, 99.92750000000001) (8.7406, 100.0) (5.506, 100.0) 
    };
    \addlegendentry{Lea1}
    
    \end{axis}
    \end{tikzpicture}}
    \hfill
    \subfloat[\label{fig:PR_NetVLAD}Place recognition results for the learning-based method NetVLAD~\cite{Arandjelovic16}.]{
    \input{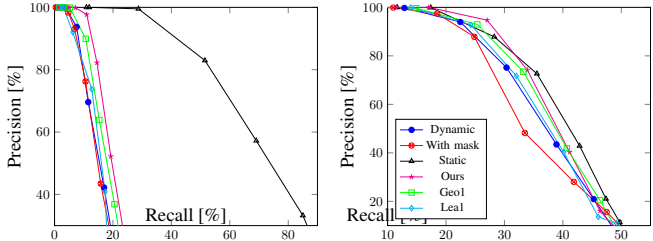}
    \hfill
    \input{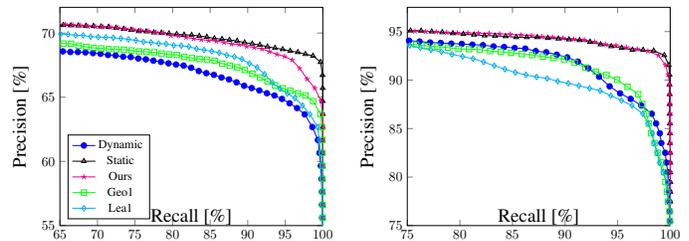}}
    \caption{\protect\subref{fig:PR_BoW} and \protect\subref{fig:PR_NetVLAD} show the precision and recall curves for the VPR results with BoW and NetVLAD respectively. 
    We report the results for the dynamic, the inpainted and ground-truth static sequences, as well as the results obtained when masking out the dynamic objects as in DynaSLAM (only in \protect\subref{fig:PR_BoW}). 
    The precision and recall curves for the inpainting methods Geo1~\cite{telea2004image} and Lea1~\cite{yu2018generative} are also presented.}
    \label{fig:PRcurves_ours}
\end{figure*}

The precision-recall curves for the BoW experiments are depicted in Fig.~\ref{fig:PR_BoW}. 
We have extracted the visual words of every frame along the trajectories and have tried to match every two images as a function of the number of common visual words. 
For both trajectories the results obtained with the dynamic images with and without masks are similar. 
This is congruent with the idea that the database of visual words mostly contains static and long-term stable words.
The first trajectory is a good example of how the VPR recall drops fast in presence of dynamic objects: a place is better represented with words from the whole static image. 
It leads to less false positives and less false negatives mostly. 
Even though our method slightly brings closer in the bag-of-words space images from the same place with different dynamic objects, we would have expected our results to be closer to those of the ground-truth static images. 
Our intuition behind these results is that the synthetic features, despite being useful for feature matching, do not fully fall on any visual words in the BoW space.

\begin{figure}
    \centering
    \subfloat[\label{fig:PRIn1}Ref]{\includegraphics[width=.24\linewidth]{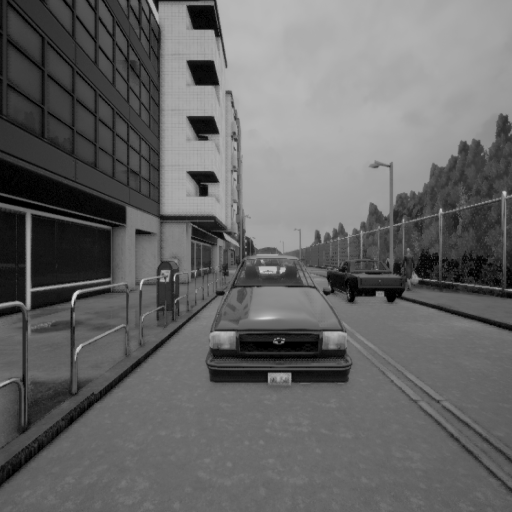}}
    \hspace*{0.001\linewidth}
    \subfloat[\label{fig:PRIn2}Query]{\includegraphics[width=.24\linewidth]{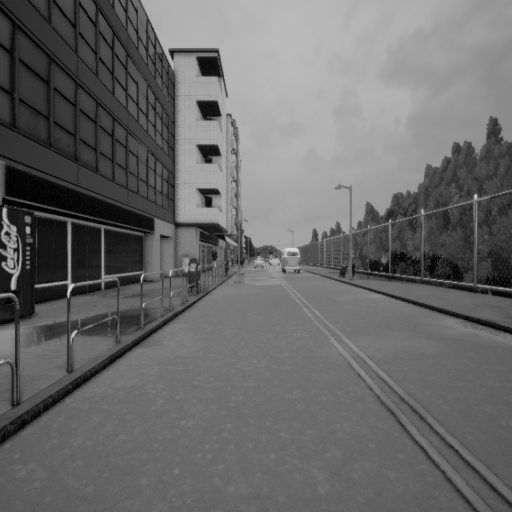}}
    \hspace*{0.01\linewidth}
    \subfloat[\label{fig:PROut1}Empty~Ref]{\includegraphics[width=.24\linewidth]{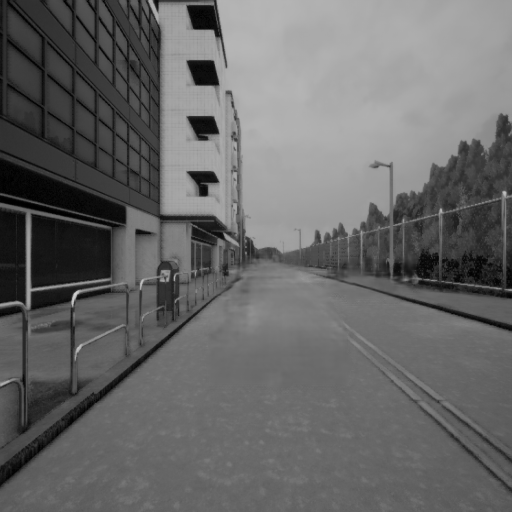}}
    \hspace*{0.001\linewidth}
    \subfloat[\label{fig:PROut2}Empty~Query]{\includegraphics[width=.24\linewidth]{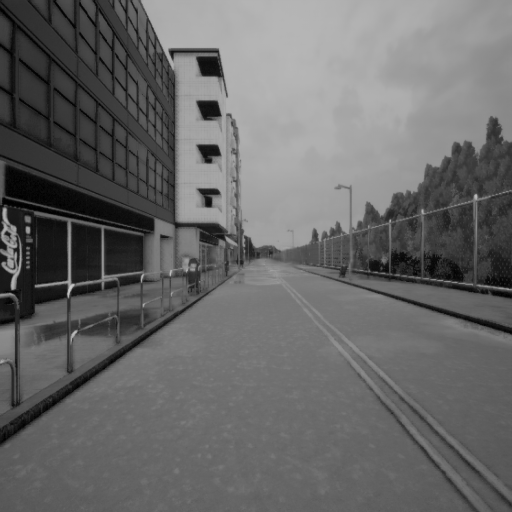}}
    \caption{\label{fig:PRCARLABoW}\protect\subref{fig:PRIn1} and \protect\subref{fig:PRIn2} show the same location with different viewpoints and objects setups. 
    NetVLAD~\cite{Arandjelovic16} fails to match them, but it succeeds when our framework is previously employed (\protect\subref{fig:PROut1} and \protect\subref{fig:PROut2}).}
\end{figure}

The precision-recall curves for the NetVLAD experiments are shown in Fig.~\ref{fig:PR_NetVLAD}. 
We have extracted the learnt descriptors of every frame along the trajectories and tried to match every two images as a function of their descriptors' Euclidean distance\footnote{with the Python-Tensorflow implementation by Cieslewski \textit{et~al.}~\cite{cieslewski2018data}}. 
Despite NetVLAD's incredible performance on VPR face to illumination, view point and clutter changes, we show that their execution is seen slightly degraded by dynamic objects. 
Empty Cities brings closer together the descriptors of the same place, and pulls apart the descriptors of different places with the same dynamic objects, leading to higher precision and recall. 
That is, the hidden semantic representations of our model match the ones learnt by NetVLAD. 
We can conclude that our method brings more relevant improvements in VPR if it is used in conjunction with a learning-based method. 
Finally, Fig.~\ref{fig:PRCARLABoW} shows a case in which NetVLAD fails at matching two dynamic frames of the same place, but manages to match them when dynamic objects are inpainted with our framework.

\textbf{Baseline for Inpainting}

We compare our approach to the inpainting methods (Geo1~\cite{telea2004image} and Lea1~\cite{yu2018generative}) w.r.t. the VPR metrics in Fig.~\ref{fig:PRcurves_ours}. 
For the BoW curves the conclusion is similar to what we have seen in the previous experiment. 
Even if the extracted synthetic features were useful for matching, they seem to be of less help for place recognition with BoW. 
Few synthetic ORB features match any existing visual word in the BoW space. 
Our method though creates more useful visual words than Lea1 and Geo1. 
As for the results with NetVLAD, the use of the geometric method Geo1 brings little improvement. 
However, the learning-based method Lea1 decreases NetVLAD’s performance on the second trajectory. 
NetVLAD can –up to some extent– ignore the dynamic classes clues, but cannot ignore the transformed regions. 
That is, the hidden semantic representations of these inpainted regions do not always match the static scene representation learnt by NetVLAD.

\begin{figure*}[t]
    \centering
    \subfloat{\includegraphics[width=0.10\linewidth]{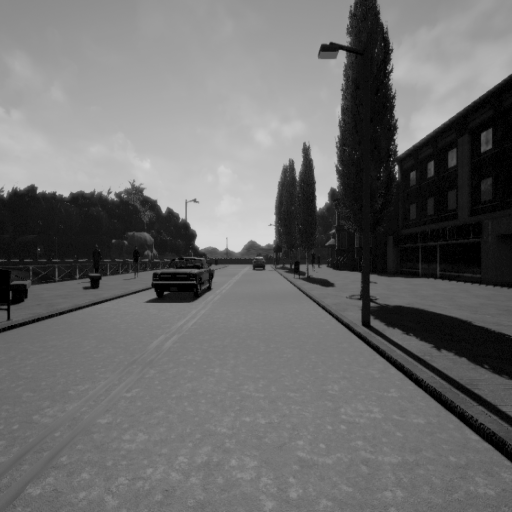}
    \hspace{\fill}
    \includegraphics[width=0.10\linewidth]{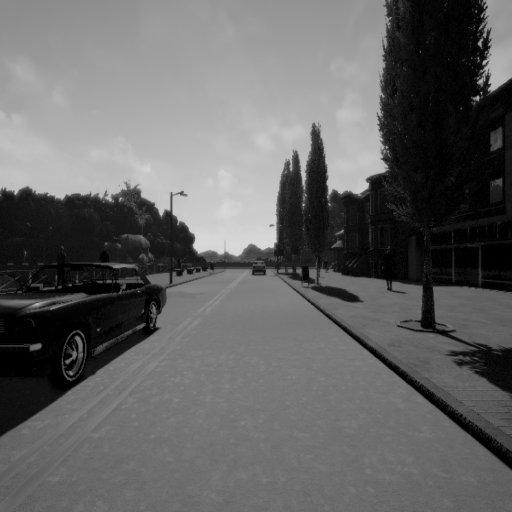}
    \hspace{\fill}
    \includegraphics[width=0.10\linewidth]{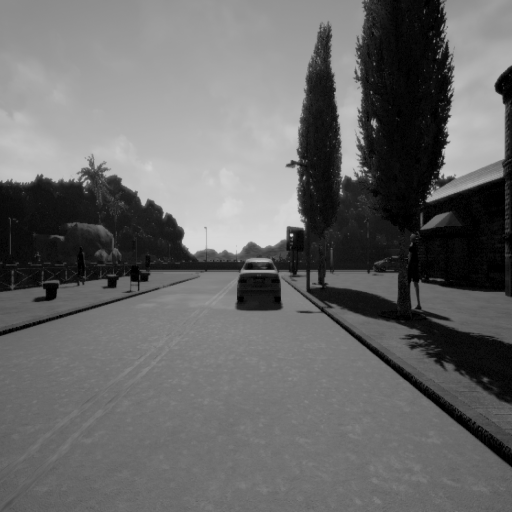}}
    \hspace{\fill}
    \subfloat{\includegraphics[width=0.10\linewidth]{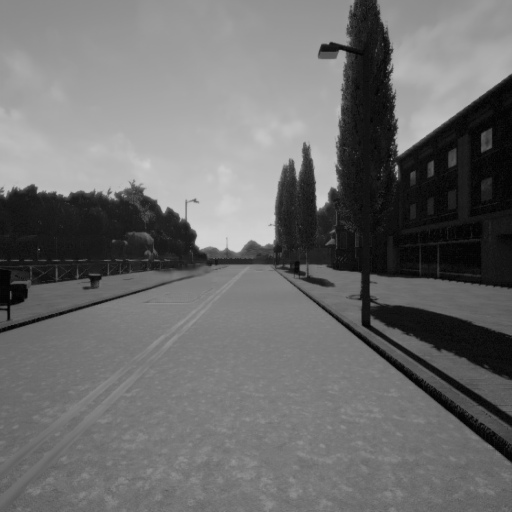}
    \hspace{\fill}
    \includegraphics[width=0.10\linewidth]{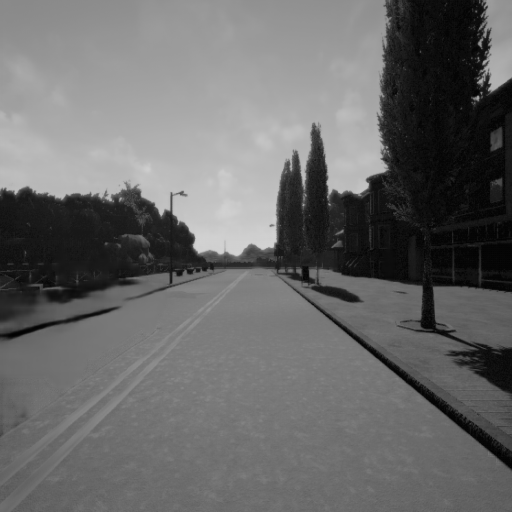}
    \hspace{\fill}
    \includegraphics[width=0.10\linewidth]{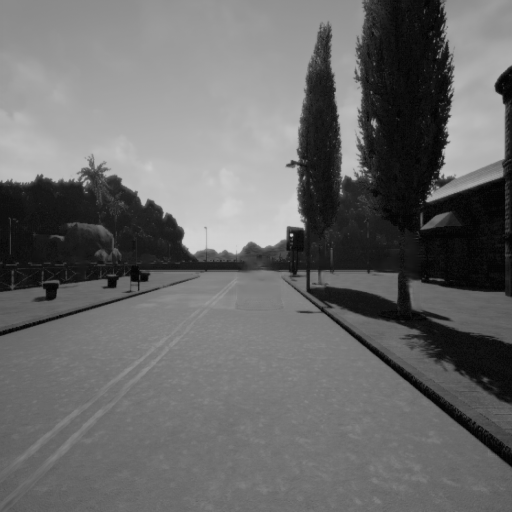}}
    \hspace{\fill}
    \subfloat{\includegraphics[width=0.10\linewidth]{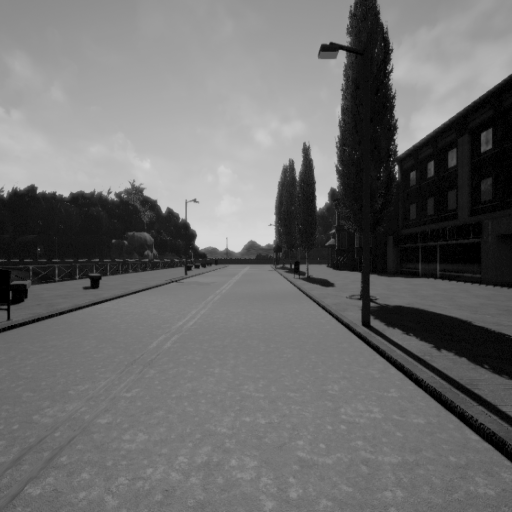}
    \hspace{\fill}
    \includegraphics[width=0.10\linewidth]{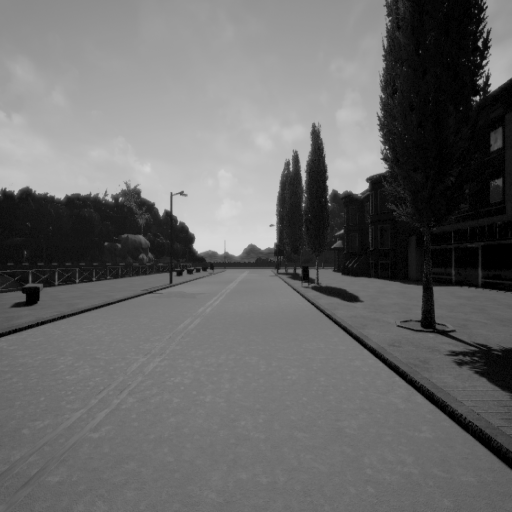}
    \hspace{\fill}
    \includegraphics[width=0.10\linewidth]{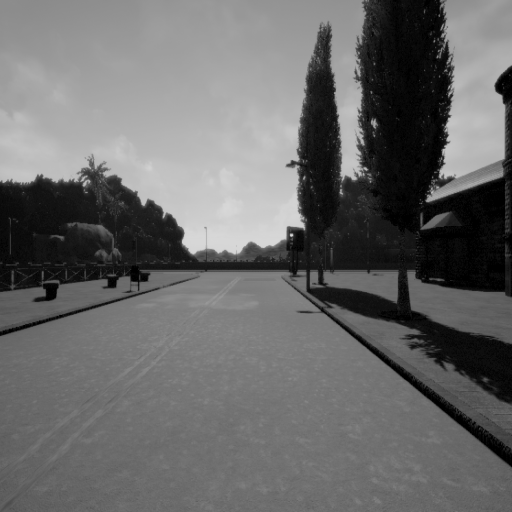}}
    \\
    \setcounter{subfigure}{0}
    \subfloat[\label{fig:DynMap} Dynamic map]{
    \begin{tikzpicture}
        \node[anchor=south west,inner sep=0] (image) at (0,0) {\includegraphics[width=0.32\linewidth]{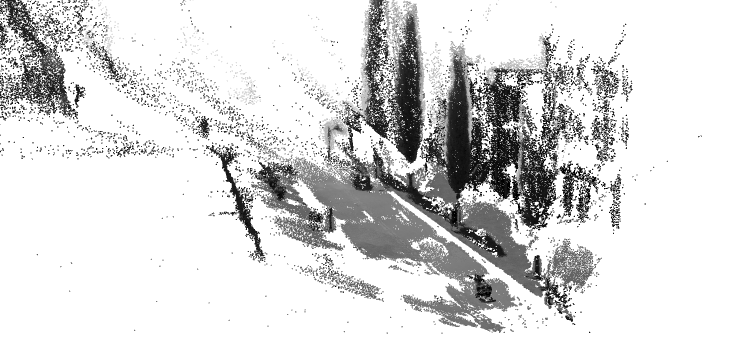}};
        \begin{scope}[x={(image.south east)},y={(image.north west)}]
            \draw[red,ultra thick,rounded corners] (0.4,0.35) rectangle (0.6,0.65);
            \draw[red,ultra thick,rounded corners] (0.50,0.03) rectangle (0.75,0.25);
        \end{scope}
    \end{tikzpicture}}
    \hspace{\fill}
    \subfloat[\label{fig:OurMap} Our map]{\includegraphics[width=0.32\linewidth]{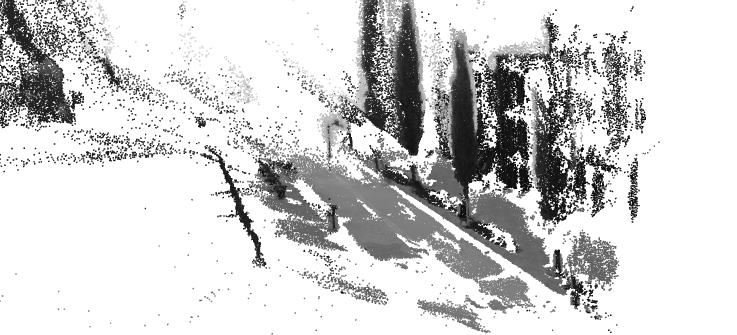}}
    \hspace{\fill}
    \subfloat[\label{fig:StaMap} Ground-truth map]{\includegraphics[width=0.32\linewidth]{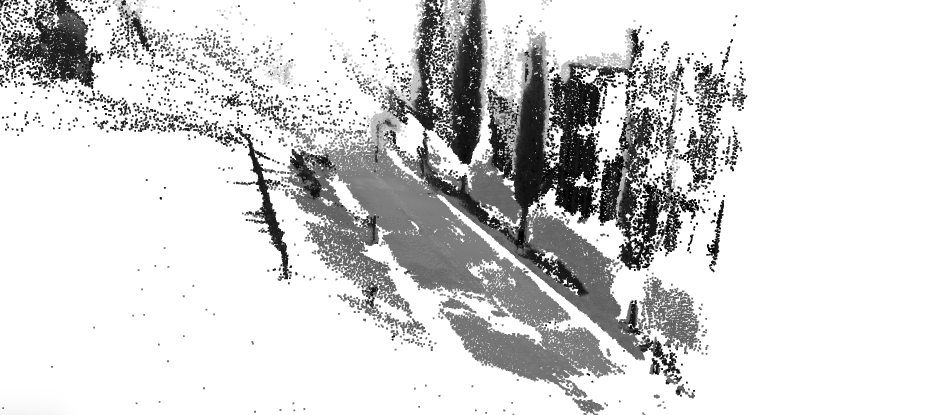}}
    \caption{Dense maps of a CARLA city environment with the COLMAP Multi-View-Stereo software. 
    The upper row shows some of the sequence images used to build such maps. 
    \protect\subref{fig:DynMap} shows the case in which the original dynamic images have been used. 
    \protect\subref{fig:OurMap} shows the resulting map with the images previously processed by our framework, and \protect\subref{fig:StaMap} uses the ground-truth static images to create the resulting map. 
    All maps are computed with the ground-truth camera poses.}
    \label{fig:DenseMap}
\end{figure*}

\subsection{Mapping}
Another important application of our framework is the creation of high-detail road maps.
Our inpainting framework allows to create long-term and reusable maps that neither show dynamic objects nor the empty spaces left by them.  
To this end we use the MVS and SfM software COLMAP~\cite{schoenberger2016sfm, schoenberger2016mvs}.

\textbf{Baseline for Mapping in our Synthetic Dataset}

Fig.~\ref{fig:DenseMap} shows the dense map of a simulated city environment (CARLA) with the original dynamic sequence, with the images processed by our framework and with the ground-truth static images. 
The map seen in Fig.~\ref{fig:DynMap} is not useful for future use since it shows dynamic objects that might not be there any more.
Fig.~\ref{fig:StaMap} shows the map built with the ground-truth static images, and Fig.~\ref{fig:OurMap} shows the map computed with our generated images. 
The areas which have been consistently inpainted along the frames are mapped, even if they have never been seen. 
When inpainting fails or is not consistent along the sequence, the mapping photometric and geometric epipolar constraints are not met and such areas cannot be reconstructed.
This idea makes our framework suitable to build stable maps.

To give a quantitative experiment on the validity of our maps, we choose the standard Iterative Closest Point (ICP) algorithm~\cite{besl1992method} based on the Euclidean fitness score. 
That is, having the point cloud built with the ground-truth static images as a fix reference, for each of its points the algorithm searches for the closest point in the target point cloud and calculates the distance based on the result of the algorithm search. 
To have a baseline for our experiment, we compute the point cloud with the dynamic images and with the CARLA segmentation. 
That is, the pixels belonging to the dynamic objects have not been used in the multi-view-stereo pipeline. 
The results can be seen in Table~\ref{tab:mapping_table}. 
Since the map has no scale, the ICP Euclidean fitness score is also up-to-scale. 
Even though the similarity score is improved when masking out dynamic objects, the map built with our images has triangulated more 3D points from the inpainted regions, and such points have a low error.

\begin{table} [h]
\begin{tabularx}{\linewidth}{@{}l*{5}{C}}
\toprule
& Dynamic & Dynamic w/ masks & Ours & Geo1~\cite{telea2004image} & Lea1~\cite{yu2018generative} \\
\midrule
ICP score & 63.21 & 49.02 & \textbf{33.34} & 82.23 & 70.89 \\
\bottomrule
\end{tabularx}
\caption{\label{tab:mapping_table} {\color{black}Quantitative mapping results for Figs.~\ref{fig:DenseMap} and~\ref{fig:DenseMap_SoTA}. 
We report the Euclidean fitness score given by the ICP algorithm. 
This score has been computed between the different point clouds w.r.t. the point cloud built with the ground-truth static images. 
Note that, since the maps do not have scale, the reported scores are up-to-scale.}}
\end{table}

\begin{figure*}[t]
    \centering
    \subfloat[\label{fig:OurMapSoTA} Our map]{
    \begin{tikzpicture}
        \node[anchor=south west,inner sep=0] (image) at (0,0) {\includegraphics[width=0.32\linewidth]{Images/our_map02.png}};
        \begin{scope}[x={(image.south east)},y={(image.north west)}]
            \draw[red,ultra thick,rounded corners] (0.4,0.35) rectangle (0.6,0.65);
            \draw[red,ultra thick,rounded corners] (0.50,0.03) rectangle (0.75,0.25);
        \end{scope}
    \end{tikzpicture}}
    \hspace{\fill}
    \subfloat[\label{fig:Geo1Map} Geo1's map]{
    \begin{tikzpicture}
        \node[anchor=south west,inner sep=0] (image) at (0,0)
        {\includegraphics[width=0.32\linewidth]{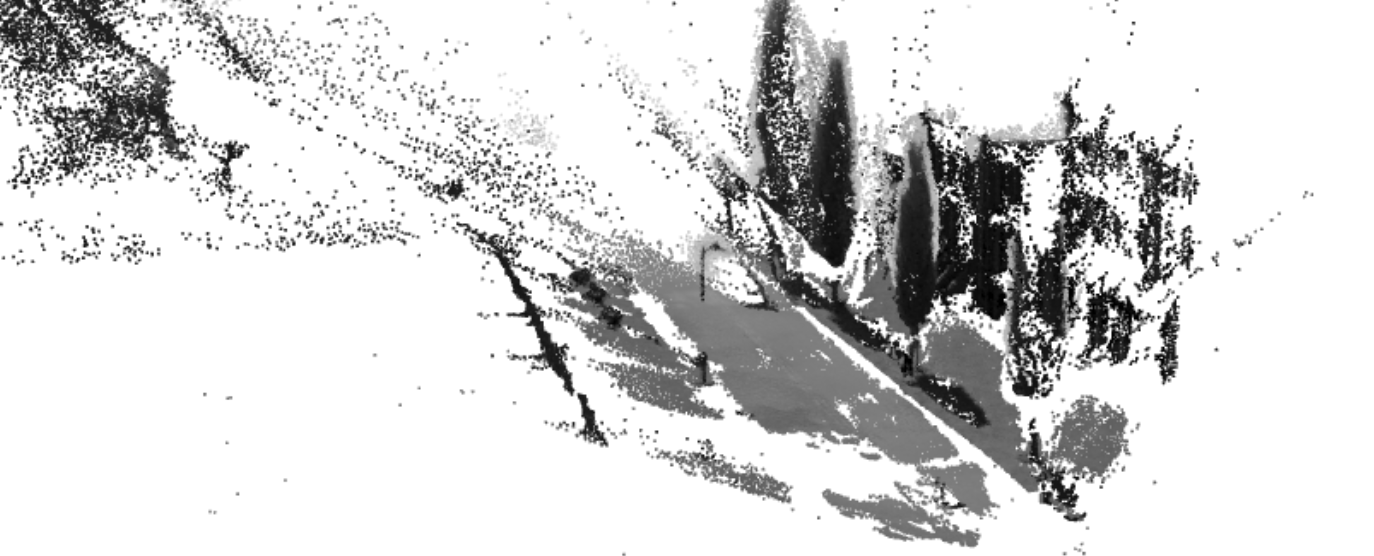}};
        \begin{scope}[x={(image.south east)},y={(image.north west)}]
            \draw[red,ultra thick,rounded corners] (0.4,0.35) rectangle (0.6,0.65);
            \draw[red,ultra thick,rounded corners] (0.50,0.03) rectangle (0.75,0.25);
        \end{scope}
    \end{tikzpicture}}
    \hspace{\fill}
    \subfloat[\label{fig:Lea1Map} Lea1's map]{
    \begin{tikzpicture}
        \node[anchor=south west,inner sep=0] (image) at (0,0)
        {\includegraphics[width=0.32\linewidth]{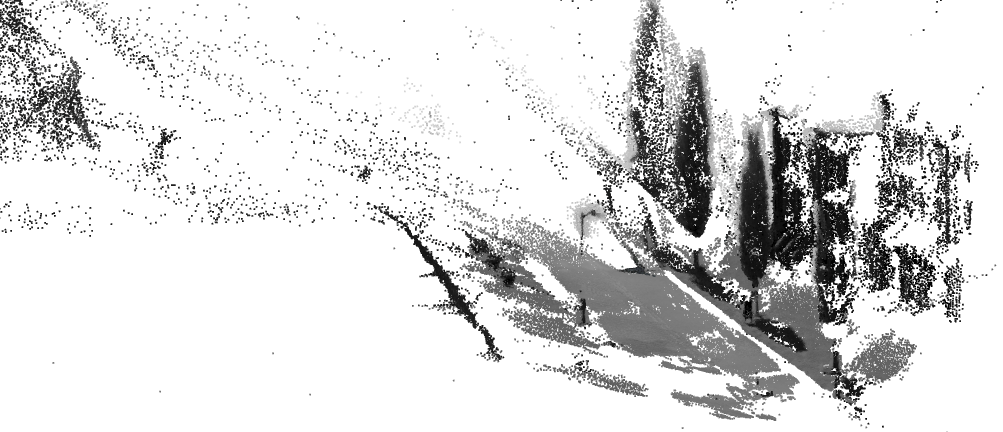}};
        \begin{scope}[x={(image.south east)},y={(image.north west)}]
            \draw[red,ultra thick,rounded corners] (0.5,0.30) rectangle (0.75,0.60);
            \draw[red,ultra thick,rounded corners] (0.55,0.03) rectangle (0.80,0.25);
        \end{scope}
    \end{tikzpicture}}
    \caption{{\color{black}Dense maps of the same CARLA city environment than in Fig.~\ref{fig:DenseMap} with the COLMAP Multi-View-Stereo software. 
    \protect\subref{fig:OurMapSoTA} shows the case in which our inpainted images have been used. 
    \protect\subref{fig:Geo1Map} and \protect\subref{fig:Lea1Map} show the resulting maps with the images previously processed by the frameworks of Telea~\cite{telea2004image} and Yu \textit{et~al.}~\cite{yu2018generative} respectively. 
    All maps are computed with the ground-truth camera poses.}}
    \label{fig:DenseMap_SoTA}
\end{figure*}

{\color{black}\textbf{Baseline for Inpainting}

We also compare our approach with the two other state-of-the-art inpainting methods (Geo1~\cite{telea2004image}  and  Lea1~\cite{yu2018generative}) w.r.t. the map quality. 
The qualitative results are depicted in Fig.~\ref{fig:DenseMap_SoTA}, and the ICP scores are shown in Table~\ref{tab:mapping_table}. 
It can be seen that with both other inpainting methods the shadows of the dynamic objects are reconstructed, as well as their prolongation into the inpainted image regions. 
This leads to an ICP score that is higher than that of the map built with dynamic objects.}

\textbf{Baseline for Mapping in Real World Environments}

\begin{figure*}
    \centering
    \subfloat{\begin{overpic}[width=.24\linewidth]{Images/kitti04_000003.png}\put(1,27.5){\includegraphics[scale=0.05]{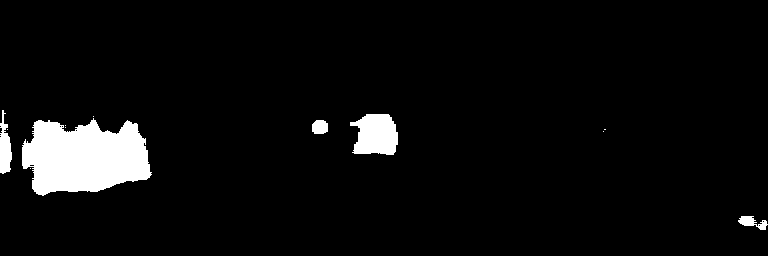}}\end{overpic}}
    \hspace{\fill}
    \subfloat{\begin{overpic}[width=.24\linewidth]{Images/kitti04_000020.png}\put(1,27.5){\includegraphics[scale=0.05]{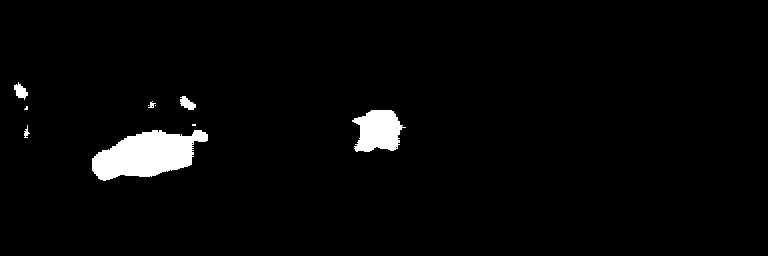}}\end{overpic}}
    \hspace{\fill}
    \includegraphics[width=0.24\linewidth]{Images/kitti04out_000003.png}
    \hspace{\fill}
    \includegraphics[width=0.24\linewidth]{Images/kitti04out_000020.png}
    \\
    \setcounter{subfigure}{0}
    \subfloat[\label{fig:KITTIDynMap} Dynamic map]{
    \begin{tikzpicture}
        \node[anchor=south west,inner sep=0] (image) at (0,0) {\includegraphics[width=0.49\linewidth]{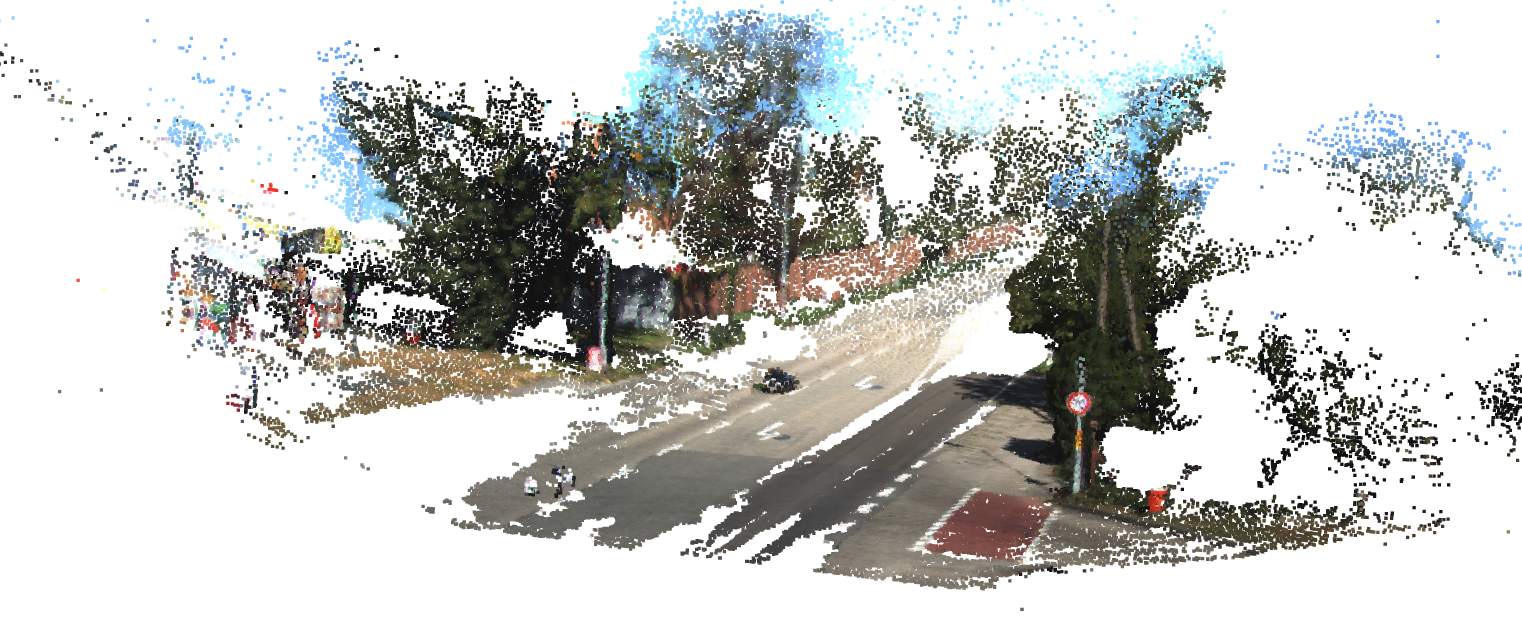}};
        \begin{scope}[x={(image.south east)},y={(image.north west)}]
            \draw[red,ultra thick,rounded corners] (0.42,0.35) rectangle (0.6,0.5);
            \draw[red,ultra thick,rounded corners] (0.3,0.15) rectangle (0.43,0.35);
        \end{scope}
    \end{tikzpicture}}
    \hspace{\fill}
    \subfloat[\label{fig:KITTIOurMap} Our map]{\includegraphics[width=0.49\linewidth]{Images/kitti_stat_map.png}}
    \caption{Dense maps built with images from the KITTI dataset~\cite{geiger2013vision}. 
    The upper row shows some of the sequence images used to build each map. 
    \protect\subref{fig:KITTIDynMap} shows the case in which the original dynamic images have been used, and \protect\subref{fig:KITTIOurMap} shows the resulting map with the images previously processed by our framework. 
    Both maps are computed with the camera poses that ORB-SLAM estimates for the different sequences.}
    \label{fig:KITTI_map}
\end{figure*}

Fig.~\ref{fig:KITTI_map} shows an example of the computed dense maps for both types of inputs with the sequence $04$ from the KITTI dataset. 
These maps have been computed with the camera poses given by ORB-SLAM using respectively the dynamic and inpainted images.
Our framework, when being able to inpaint a coherent context along the sequence, allows us to densely reconstruct unseen areas. 
When the inpainting task is not coherent along the sequence, the epipolar constraints are not met and therefore such areas cannot be reconstructed. 
Note that this map is shown in RGB only for visualization purposes\footnote{Since our framework offers enough flexibility, we have re-trained our model with RGB images just as explained in section~\ref{sec:system}. 
The only difference is that since the features have to be extracted in gray-scale images, we add a convolutional layer to convert the RGB output images to gray scale.}. 
We do recommend to use the gray-scale images for both localization and mapping purposes since a lower reconstruction error is usually achieved.

\subsection{Timing Analysis}

Reporting our framework efficiency is crucial to judge its suitability for autonomous driving and robotic tasks in general. 
The end-to-end pipeline runs at 100~fps on a nVidia Titan Xp 12GB with images of a $512\times512$ resolution. 
Out of the 10~ms it takes to process one frame, 8~ms are invested into obtaining its semantic segmentation, and 2~ms are used for the inpainting task. 
Other than to deal with dynamic objects, the semantic segmentation may be needed for many other tasks involved in autonomous navigation. 
In such cases, our framework would only add 2 extra ms per frame. 
Based on our analysis, we consider that the inpainting task is not a bottleneck.

\section{Failure Modes and Future Work}

{\color{black}
The aim of this section is to provide the reader with an understanding on the benefits and limitations of our proposal, and on how to integrate it on a VO, SLAM or MVS pipeline. 

Having a scene static representation leads to better visual odometry results than just excluding the features belonging to moving objects. 
The presented inpainting approach has though some weaknesses: the bigger the image dynamic region is, the lower inpainting quality the resulting image has. 
Empty Cities would be suitable in setups in which approximately less than 15~\% of the camera field of view is covered by dynamic objects. 
In such setups, the reconstructed L1 error is acceptable and usually lies between 1 and 10~\%. 
The L1 error goes above 10~\% when more than 15~\% of the image pixels are covered~\footnote{As a practical example, the Oxford Robotcar sequence $2014$-$05$-$06$-$12$-$54$-$54$ has 56~\% of images with less than 5~\% of covered pixels, 30~\% with a percentage of dynamic pixels between 5 and 10~\%, 12~\% between 10 and 15~\% and 2~\% between 15 and 20~\%. 
This sequence is not Manhattan at 11 am, but shows cars parked at both sides of the road and cars driving nearby.}.
Work remains to be done to tackle extreme situations. 
Also, developing a system that processes the sequence as a whole, rather than binning it into independent frames, would result in a more consistent image inpainting along time.

Our system processes the image streams outside of the application pipeline and hence can be used naturally as a front end to many existing systems.
We hereby want to discuss other application-dependent possibilities to boost its performance.

\textbf{Visual Odometry}: Removing the features that belong to stationary objects certainly damages VO. 
However, \textit{e.g.}, a car can from one frame to another change from static to moving. 
Had we a movement detector, we would use the static objects' features and the inpainted ones behind moving objects.

\textbf{Place Recognition}: Using the features of stationary dynamic objects damages the performance of place recognition algorithms. 
For example, two frames of the same place with a different setup of parked cars can be incorrectly tagged as a different place.
Also, two frames of different places with the same parked cars setup can be wrongly matched as the same place. 
Only the features of objects that remain stable in the long term (buildings, sidewalks, \textit{etc.}) would benefit VPR.

\textbf{Mapping}: A map containing information about dynamic objects would be useless for future reuse. 
Only the information belonging to objects that remain stable in the long term, as well as the most likely static representation of the static scene behind dynamic objects should be included in the map.

Ideally, one would use a movement detector to identify the status of the different observed instances and also to allow the discovery of new dynamic classes on the fly~\cite{zhou2018dynamic}. 
The features belonging to static instances would be used for visual odometry, and the corresponding inpainted features would be used for place recognition and mapping. 
This approach would bring the highest accuracy results but would entail a series of modifications to the existing pipeline.
Our method would currently pose problems in the case in which one wanted to inpaint the static scene behind a car that is currently moving, and this static scene contained parked cars. 
Our model would fail to reconstruct the unseen parts of the parked cars. 
It would be interesting for future work to include such scenarios in our training data. 
Our suggestion is, for now, inpainting all instances, regardless of their current dynamic status. }

\section{Conclusions}

We have presented an end-to-end deep learning framework that translates images that contain dynamic objects within a city environment, such as vehicles or pedestrians, into a realistic image with only static content. 
These images are suitable for visual {\color{black}odometry, place recognition} and mapping tasks thanks to a new loss based on steganalysis techniques and ORB features maps, descriptors and orientation.
We motivate this extra complexity by showing quantitatively that the systems ORB-SLAM and DSO obtain a higher accuracy when utilizing the images synthesized with this loss. 
Also, mapping systems can benefit from this approach since not only they would not map dynamic objects but also would they map the plausible static scene behind them.
Finally, an architectural nicety is that our system processes the image streams outside of the localization pipeline, either offline or online, and hence can be used naturally as a front end to many existing systems.

\section{Acknowledgements}

This work is supported by the Spanish Ministry of Economy and Competitiveness (project PID2019-108398GB-I00 and FPI grant BES-2016-077836), the Arag\'on regional government (Grupos DGA T45-17R, T45-20R), the EU H2020 research project under grant agreement No 688652 and the Swiss State Secretariat for Education, Research and Innovation (SERI) No 15.0284 and NVIDIA, through the donation of a Titan X GPU.

\begin{IEEEbiography}[{\includegraphics[width=1in,height=1.25in,clip,keepaspectratio]{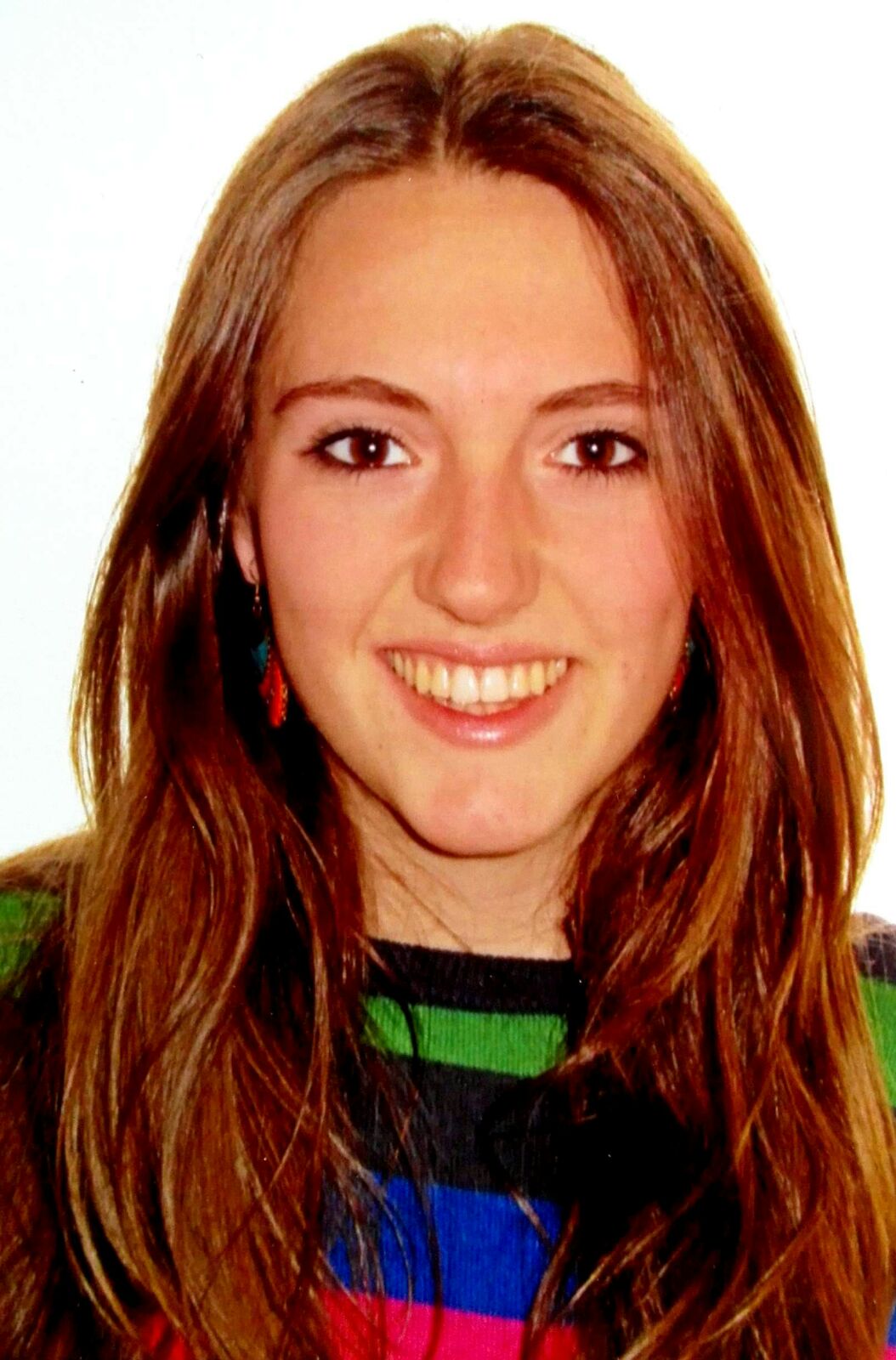}}]{Berta Bescos}

was born in Zaragoza, Spain, in 1993. 
She received a Bachelor's and a M.S. Degree in Industrial Engineering with mention in Robotics and Computer Vision from the University of Zaragoza, where she is currently pursuing her Ph.D. degree at the I3A Robotics, Perception and Real Time Group. 
Her research interests lie in the intersection between perception and learning for robotics. 
More concisely, her PhD. topic involves dealing with dynamic objects in SLAM for a better scene understanding.

\end{IEEEbiography}

\begin{IEEEbiography}[{\includegraphics[width=1in,height=1.25in,clip,keepaspectratio]{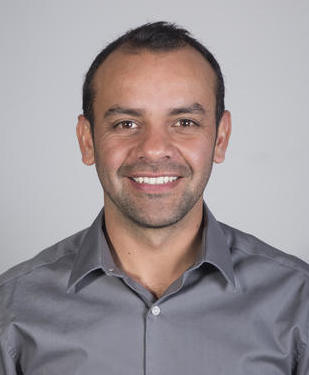}}]{Cesar Cadena}

received the Ph.D. degree in Computer Science from University of Zaragoza, Zaragoza, Spain, in 2011. He is a Senior Researcher at ETH Zurich, Switzerland. His main research interest lies in the interception of perception and learning in robotics. He is particularly interested on how to provide machines the capability of understanding this ever changing world through the sensory information they can gather. He has work intensively on Robotic Scene Understanding, both geometry and semantics, covering Semantic Mapping, Data Association and Place Recognition tasks, Simultaneous Localization and Mapping problems, as well as persistent mapping in dynamic environments.

\end{IEEEbiography}

\begin{IEEEbiography}[{\includegraphics[width=1in,height=1.25in,clip,keepaspectratio]{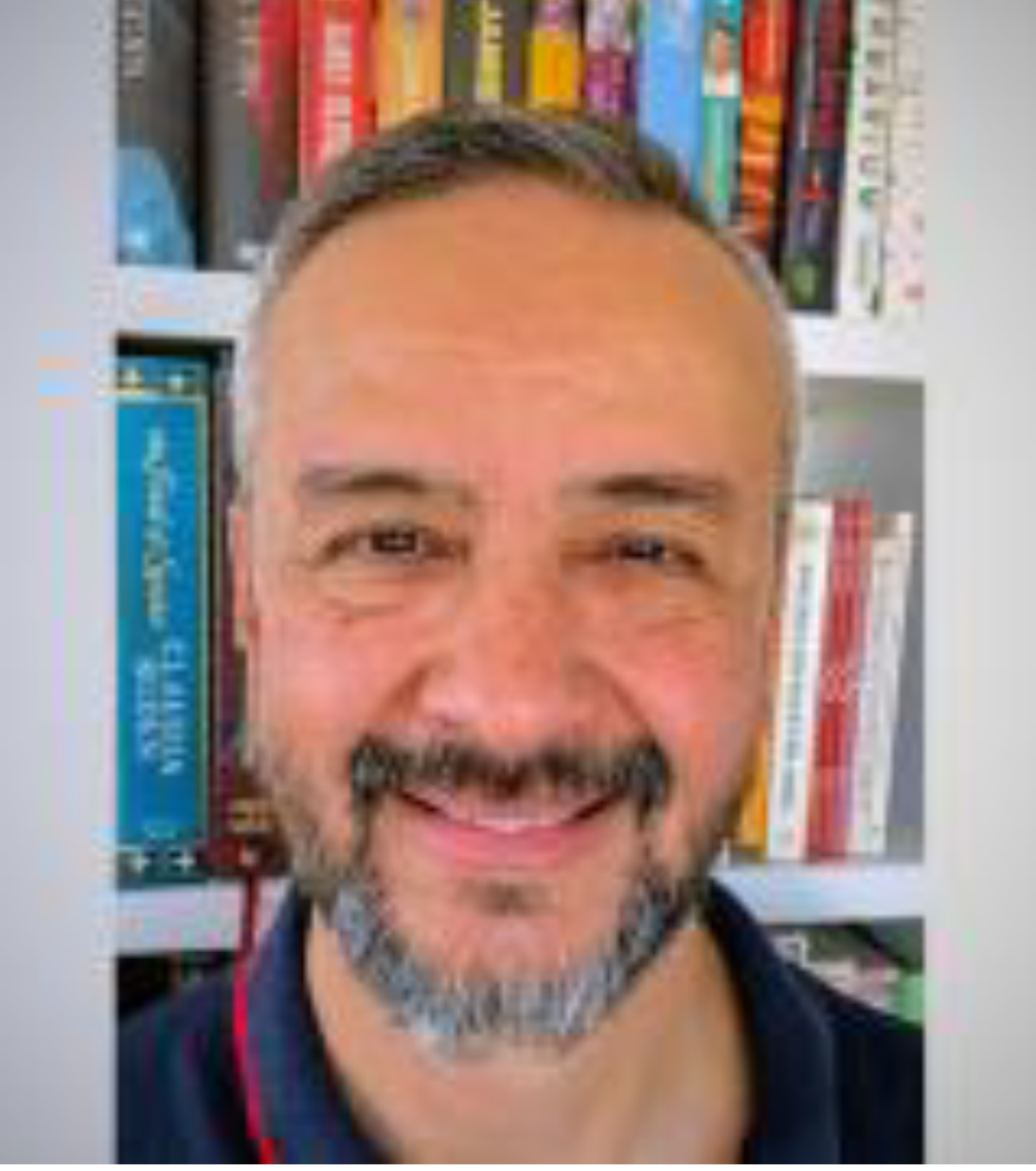}}]{José Neira} was born in Bogotá, Colombia, in 1963.
He received the M.S. degree in computer science
from the Universidad de los Andes, Colombia, in
1986, and the Ph.D. degree in computer science
from the University of Zaragoza, Spain, in 1993.
Since 2010, José is Full Professor at the Departamento
de Informática e Ingeniería de Sistemas, University
of Zaragoza, where he is in charge of courses
in compiler theory, computer vision, machine learning and mobile
robotics. His current research interests are centered around 
Robust, Life-long
Simultaneous  Localization  and Mapping.  José  also coordinates  the university's Master Program in Robotics, Graphics and Computer Vision.

\end{IEEEbiography}

\addtolength{\textheight}{-1cm} 

\bibliographystyle{ieeetr}
\bibliography{ref}

\begin{thebibliography}{10}

\bibitem{murTRO2015}
R.~Mur-Artal, J.~M.~M. Montiel, and J.~D. Tard\'os, ``{ORB-SLAM}: a versatile
  and accurate monocular {SLAM} system,'' {\em IEEE Transactions on Robotics},
  vol.~31, no.~5, pp.~1147--1163, 2015.

\bibitem{engel2018direct}
J.~Engel, V.~Koltun, and D.~Cremers, ``Direct sparse odometry,'' {\em IEEE
  transactions on pattern analysis and machine intelligence}, vol.~40, no.~3,
  pp.~611--625, 2018.

\bibitem{forster2014svo}
C.~Forster, M.~Pizzoli, and D.~Scaramuzza, ``{SVO: Fast semi-direct monocular
  visual odometry},'' in {\em 2014 IEEE international conference on robotics
  and automation (ICRA)}, pp.~15--22, IEEE, 2014.

\bibitem{agudo2015sequential}
A.~Agudo, F.~Moreno-Noguer, B.~Calvo, and J.~M.~M. Montiel, ``Sequential
  non-rigid structure from motion using physical priors,'' {\em IEEE
  transactions on pattern analysis and machine intelligence}, vol.~38, no.~5,
  pp.~979--994, 2015.

\bibitem{lamarca2019defslam}
J.~Lamarca, S.~Parashar, A.~Bartoli, and J.~Montiel, ``{DefSLAM: Tracking and
  Mapping of Deforming Scenes from Monocular Sequences},'' {\em arXiv preprint
  arXiv:1908.08918}, 2019.

\bibitem{bescos2018dynaslam}
B.~Bescos, J.~M. F{\'a}cil, J.~Civera, and J.~Neira, ``{DynaSLAM: Mapping,
  Tracking and Inpainting in Dynamic Scenes},'' {\em IEEE Robotics and
  Automation Letters}, vol.~3, pp.~4076 -- 4083, 2018.

\bibitem{alcantarilla2012combining}
P.~F. Alcantarilla, J.~J. Yebes, J.~Almaz{\'a}n, and L.~M. Bergasa, ``{On
  combining visual SLAM and dense scene flow to increase the robustness of
  localization and mapping in dynamic environments},'' in {\em Robotics and
  Automation (ICRA), 2012 IEEE International Conference on}, pp.~1290--1297,
  IEEE, 2012.

\bibitem{wang2014motion}
Y.~Wang and S.~Huang, ``{Motion segmentation based robust RGB-D SLAM},'' in
  {\em Intelligent Control and Automation (WCICA), 2014 11th World Congress
  on}, pp.~3122--3127, IEEE, 2014.

\bibitem{tan2013robust}
W.~Tan, H.~Liu, Z.~Dong, G.~Zhang, and H.~Bao, ``{Robust monocular SLAM in
  dynamic environments},'' in {\em ISMAR}, pp.~209--218, 2013.

\bibitem{he2017mask}
K.~He, G.~Gkioxari, P.~Doll{\'a}r, and R.~Girshick, ``{Mask R-CNN},'' in {\em
  Computer Vision (ICCV), 2017 IEEE International Conference on},
  pp.~2980--2988, IEEE, 2017.

\bibitem{romera2018erfnet}
E.~Romera, J.~M. Alvarez, L.~M. Bergasa, and R.~Arroyo, ``{ERFNet: Efficient
  Residual Factorized ConvNet for Real-Time Semantic Segmentation},'' {\em IEEE
  Transactions on Intelligent Transportation Systems}, vol.~19, no.~1,
  pp.~263--272, 2018.

\bibitem{barnes2017driven}
D.~Barnes, W.~Maddern, G.~Pascoe, and I.~Posner, ``Driven to distraction:
  Self-supervised distractor learning for robust monocular visual odometry in
  urban environments,'' in {\em 2018 IEEE International Conference on Robotics
  and Automation (ICRA)}, pp.~1894--1900, IEEE, 2018.

\bibitem{zhou2018dynamic}
G.~Zhou, B.~Bescos, M.~Dymczyk, M.~Pfeiffer, J.~Neira, and R.~Siegwart,
  ``Dynamic objects segmentation for visual localization in urban
  environments,'' {\em Intelligent Robots and Systems (IROS), workshop From
  freezing to jostling robots: Current challenges and new paradigms for safe
  robot navigation in dense crowds}, 2018.

\bibitem{telea2004image}
A.~Telea, ``An image inpainting technique based on the fast marching method,''
  {\em Journal of graphics tools}, vol.~9, no.~1, pp.~23--34, 2004.

\bibitem{bertalmio2001navier}
M.~Bertalmio, A.~L. Bertozzi, and G.~Sapiro, ``Navier-stokes, fluid dynamics,
  and image and video inpainting,'' in {\em Proceedings of the 2001 IEEE
  Computer Society Conference on Computer Vision and Pattern Recognition. CVPR
  2001}, vol.~1, pp.~I--I, IEEE, 2001.

\bibitem{liu2018image}
G.~Liu, F.~A. Reda, K.~J. Shih, T.-C. Wang, A.~Tao, and B.~Catanzaro, ``Image
  inpainting for irregular holes using partial convolutions,'' in {\em The
  European Conference on Computer Vision (ECCV)}, 2018.

\bibitem{yu2018generative}
J.~Yu, Z.~Lin, J.~Yang, X.~Shen, X.~Lu, and T.~S. Huang, ``{Generative Image
  Inpainting with Contextual Attention},'' {\em Proceedings of the IEEE
  Conference on Computer Vision and Pattern Recognition}, 2018.

\bibitem{iizuka2017globally}
S.~Iizuka, E.~Simo-Serra, and H.~Ishikawa, ``{Globally and locally consistent
  image completion},'' {\em ACM Transactions on Graphics (TOG)}, vol.~36,
  no.~4, p.~107, 2017.

\bibitem{pathak2016context}
D.~Pathak, P.~Krahenbuhl, J.~Donahue, T.~Darrell, and A.~A. Efros, ``Context
  encoders: Feature learning by inpainting,'' in {\em Proceedings of the IEEE
  Conference on Computer Vision and Pattern Recognition}, pp.~2536--2544, 2016.

\bibitem{isola2017image}
P.~Isola, J.-Y. Zhu, T.~Zhou, and A.~A. Efros, ``{Image-to-Image Translation
  with Conditional Adversarial Networks},'' {\em Computer Vision and Pattern
  Recognition}, 2017.

\bibitem{bescos2019empty}
B.~Bescos, R.~Siegwart, J.~Neira, and C.~Cadena, ``{Empty Cities: Image
  Inpainting for a Dynamic-Object-Invariant Space},'' {\em IEEE International
  Conference on Robotics and Automation}, 2019.

\bibitem{rublee2011orb}
E.~Rublee, V.~Rabaud, K.~Konolige, and G.~Bradski, ``Orb: An efficient
  alternative to sift or surf,'' in {\em Computer Vision (ICCV), 2011 IEEE
  international conference on}, pp.~2564--2571, IEEE, 2011.

\bibitem{mur2017orb}
R.~Mur-Artal and J.~D. Tard{\'o}s, ``{ORB-SLAM2: An open-source slam system for
  monocular, stereo, and RGB-D cameras},'' {\em IEEE T-RO}, 2017.

\bibitem{klein2007parallel}
G.~Klein and D.~Murray, ``{Parallel tracking and mapping for small AR
  workspaces},'' in {\em ISMAR}, pp.~225--234, 2007.

\bibitem{runz2018maskfusion}
M.~Runz, M.~Buffier, and L.~Agapito, ``Maskfusion: Real-time recognition,
  tracking and reconstruction of multiple moving objects,'' in {\em 2018 IEEE
  International Symposium on Mixed and Augmented Reality (ISMAR)}, pp.~10--20,
  IEEE, 2018.

\bibitem{scona2018staticfusion}
R.~Scona, M.~Jaimez, Y.~R. Petillot, M.~Fallon, and D.~Cremers,
  ``{StaticFusion: Background Reconstruction for Dense RGB-D SLAM in Dynamic
  Environments},'' in {\em Robotics and Automation (ICRA), 2018 IEEE
  International Conference on}, Institute of Electrical and Electronics
  Engineers, 2018.

\bibitem{granados2012background}
M.~Granados, K.~I. Kim, J.~Tompkin, J.~Kautz, and C.~Theobalt, ``Background
  inpainting for videos with dynamic objects and a free-moving camera,'' in
  {\em European Conference on Computer Vision}, pp.~682--695, Springer, 2012.

\bibitem{uittenbogaard2018moving}
R.~Uittenbogaard, D.~Gavrila, C.~Sebastian, and J.~Vijverberg, ``Moving object
  detection and image inpainting in street-view imagery,'' {\em Master Thesis},
  2018.

\bibitem{efros2001image}
A.~A. Efros and W.~T. Freeman, ``Image quilting for texture synthesis and
  transfer,'' in {\em Proceedings of the 28th annual conference on Computer
  graphics and interactive techniques}, pp.~341--346, ACM, 2001.

\bibitem{yang2017high}
C.~Yang, X.~Lu, Z.~Lin, E.~Shechtman, O.~Wang, and H.~Li, ``High-resolution
  image inpainting using multi-scale neural patch synthesis,'' in {\em The IEEE
  Conference on Computer Vision and Pattern Recognition (CVPR)}, vol.~1, p.~3,
  2017.

\bibitem{song2017image}
Y.~Song, C.~Yang, Z.~L. Lin, H.~Li, Q.~Huang, and C.-C.~J. Kuo, ``Image
  inpainting using multi-scale feature image translation,'' {\em CoRR},
  vol.~abs/1711.08590, 2017.

\bibitem{ulyanov2017deep}
D.~Ulyanov, A.~Vedaldi, and V.~Lempitsky, ``Deep image prior,'' in {\em
  Proceedings of the IEEE Conference on Computer Vision and Pattern
  Recognition}, pp.~9446--9454, 2018.

\bibitem{goodfellow2014generative}
I.~Goodfellow, J.~Pouget-Abadie, M.~Mirza, B.~Xu, D.~Warde-Farley, S.~Ozair,
  A.~Courville, and Y.~Bengio, ``Generative adversarial nets,'' in {\em
  Advances in neural information processing systems}, pp.~2672--2680, 2014.

\bibitem{gauthier2014conditional}
J.~Gauthier, ``Conditional generative adversarial nets for convolutional face
  generation,'' {\em Class Project for Stanford CS231N: Convolutional Neural
  Networks for Visual Recognition, Winter semester}, vol.~2014, no.~5, p.~2,
  2014.

\bibitem{ronneberger2015u}
O.~Ronneberger, P.~Fischer, and T.~Brox, ``U-net: Convolutional networks for
  biomedical image segmentation,'' in {\em International Conference on Medical
  image computing and computer-assisted intervention}, pp.~234--241, Springer,
  2015.

\bibitem{guerrero2018white}
R.~Guerrero, C.~Qin, O.~Oktay, C.~Bowles, L.~Chen, R.~Joules, R.~Wolz,
  M.~Vald{\'e}s-Hern{\'a}ndez, D.~Dickie, J.~Wardlaw, {\em et~al.}, ``White
  matter hyperintensity and stroke lesion segmentation and differentiation
  using convolutional neural networks,'' {\em NeuroImage: Clinical}, vol.~17,
  pp.~918--934, 2018.

\bibitem{hinton2006reducing}
G.~E. Hinton and R.~R. Salakhutdinov, ``Reducing the dimensionality of data
  with neural networks,'' {\em science}, vol.~313, no.~5786, pp.~504--507,
  2006.

\bibitem{kingma2013auto}
D.~P. Kingma and M.~Welling, ``Auto-encoding variational bayes,'' {\em
  International Conference on Learning Representations}, 2013.

\bibitem{fridrich2012rich}
J.~Fridrich and J.~Kodovsky, ``Rich models for steganalysis of digital
  images,'' {\em IEEE Transactions on Information Forensics and Security},
  vol.~7, no.~3, pp.~868--882, 2012.

\bibitem{zhou2018learning}
P.~Zhou, X.~Han, V.~I. Morariu, and L.~S. Davis, ``Learning rich features for
  image manipulation detection,'' {\em Proceedings of the IEEE conference on
  Computer Vision and Pattern Recognition}, 2018.

\bibitem{galvez2012bags}
D.~G{\'a}lvez-L{\'o}pez and J.~D. Tardos, ``Bags of binary words for fast place
  recognition in image sequences,'' {\em IEEE Transactions on Robotics},
  vol.~28, no.~5, pp.~1188--1197, 2012.

\bibitem{rosten2006machine}
E.~Rosten and T.~Drummond, ``Machine learning for high-speed corner
  detection,'' in {\em European conference on computer vision}, pp.~430--443,
  Springer, 2006.

\bibitem{porav2018adversarial}
H.~Porav, W.~Maddern, and P.~Newman, ``Adversarial training for adverse
  conditions: Robust metric localisation using appearance transfer,'' {\em IEEE
  International Conference on Robotics and Automation}, 2018.

\bibitem{bay2008speeded}
H.~Bay, A.~Ess, T.~Tuytelaars, and L.~Van~Gool, ``Speeded-up robust features
  (surf),'' {\em Computer vision and image understanding}, vol.~110, no.~3,
  pp.~346--359, 2008.

\bibitem{erfnet2017er}
E.~Romera, J.~M. Alvarez, L.~M. Bergasa, and R.~Arroyo, ``{ERFNet}.''
  \url{https://github.com/Eromera/erfnet}, 2017.

\bibitem{cordts2016cityscapes}
M.~Cordts, M.~Omran, S.~Ramos, T.~Rehfeld, M.~Enzweiler, R.~Benenson,
  U.~Franke, S.~Roth, and B.~Schiele, ``The cityscapes dataset for semantic
  urban scene understanding,'' in {\em Proceedings of the IEEE conference on
  computer vision and pattern recognition}, pp.~3213--3223, 2016.

\bibitem{Dosovitskiy17}
A.~Dosovitskiy, G.~Ros, F.~Codevilla, A.~Lopez, and V.~Koltun, ``{CARLA}: {An}
  open urban driving simulator,'' in {\em Proceedings of the 1st Annual
  Conference on Robot Learning}, pp.~1--16, 2017.

\bibitem{geiger2013vision}
A.~Geiger, P.~Lenz, C.~Stiller, and R.~Urtasun, ``{Vision meets robotics: The
  KITTI dataset},'' {\em IJRR}, vol.~32, no.~11, pp.~1231--1237, 2013.

\bibitem{wang2004image}
Z.~Wang, A.~C. Bovik, H.~R. Sheikh, E.~P. Simoncelli, {\em et~al.}, ``Image
  quality assessment: from error visibility to structural similarity,'' {\em
  IEEE transactions on image processing}, vol.~13, no.~4, pp.~600--612, 2004.

\bibitem{RobotCarDatasetIJRR}
W.~Maddern, G.~Pascoe, C.~Linegar, and P.~Newman, ``{1 Year, 1000km: The Oxford
  RobotCar Dataset},'' {\em The International Journal of Robotics Research
  (IJRR)}, vol.~36, no.~1, pp.~3--15, 2017.

\bibitem{zhou2017places}
B.~Zhou, A.~Lapedriza, A.~Khosla, A.~Oliva, and A.~Torralba, ``Places: A 10
  million image database for scene recognition,'' {\em IEEE Transactions on
  Pattern Analysis and Machine Intelligence}, 2017.

\bibitem{hosseinzadeh2018fast}
S.~Hosseinzadeh, M.~Shakeri, and H.~Zhang, ``Fast shadow detection from a
  single image using a patched convolutional neural network,'' in {\em 2018
  IEEE/RSJ International Conference on Intelligent Robots and Systems (IROS)},
  pp.~3124--3129, IEEE, 2018.

\bibitem{gaidon2016virtual}
A.~Gaidon, Q.~Wang, Y.~Cabon, and E.~Vig, ``Virtual worlds as proxy for
  multi-object tracking analysis,'' {\em Proceedings of the IEEE Conference on
  Computer Vision and Pattern Recognition}, 2016.

\bibitem{peris2012towards}
M.~Peris, S.~Martull, A.~Maki, Y.~Ohkawa, and K.~Fukui, ``Towards a simulation
  driven stereo vision system,'' in {\em Pattern Recognition (ICPR), 2012 21st
  International Conference on}, pp.~1038--1042, IEEE, 2012.

\bibitem{skinner2016high}
J.~Skinner, S.~Garg, N.~S{\"u}nderhauf, P.~Corke, B.~Upcroft, and M.~Milford,
  ``High-fidelity simulation for evaluating robotic vision performance,'' in
  {\em Intelligent Robots and Systems (IROS), 2016 IEEE/RSJ International
  Conference on}, pp.~2737--2744, IEEE, 2016.

\bibitem{tobin2017domain}
J.~Tobin, R.~Fong, A.~Ray, J.~Schneider, W.~Zaremba, and P.~Abbeel, ``Domain
  randomization for transferring deep neural networks from simulation to the
  real world,'' in {\em Intelligent Robots and Systems (IROS), 2017 IEEE/RSJ
  International Conference on}, pp.~23--30, IEEE, 2017.

\bibitem{gao2018ldso}
X.~Gao, R.~Wang, N.~Demmel, and D.~Cremers, ``{LDSO: Direct Sparse Odometry
  with Loop Closure},'' {\em International Conference on Intelligent Robots and
  Systems}, 2018.

\bibitem{Arandjelovic16}
R.~Arandjelovi\'c, P.~Gronat, A.~Torii, T.~Pajdla, and J.~Sivic, ``{NetVLAD}:
  {CNN} architecture for weakly supervised place recognition,'' in {\em IEEE
  Conference on Computer Vision and Pattern Recognition}, 2016.

\bibitem{olid2018single}
D.~Olid, J.~M. F{\'a}cil, and J.~Civera, ``Single-view place recognition under
  seasonal changes,'' {\em Intelligent Robots and Operative Systems (IROS),
  workshop Planning, Perception and Navigation for Intelligent Vehicles}, 2018.

\bibitem{mur2014fast}
R.~Mur-Artal and J.~D. Tard{\'o}s, ``Fast relocalisation and loop closing in
  keyframe-based slam,'' in {\em 2014 IEEE International Conference on Robotics
  and Automation (ICRA)}, pp.~846--853, IEEE, 2014.

\bibitem{cieslewski2018data}
T.~Cieslewski, S.~Choudhary, and D.~Scaramuzza, ``Data-efficient decentralized
  visual slam,'' in {\em 2018 IEEE International Conference on Robotics and
  Automation (ICRA)}, pp.~2466--2473, IEEE, 2018.

\bibitem{schoenberger2016sfm}
J.~L. Sch\"{o}nberger and J.-M. Frahm, ``Structure-from-motion revisited,'' in
  {\em Conference on Computer Vision and Pattern Recognition (CVPR)}, 2016.

\bibitem{schoenberger2016mvs}
J.~L. Sch\"{o}nberger, E.~Zheng, M.~Pollefeys, and J.-M. Frahm, ``Pixelwise
  view selection for unstructured multi-view stereo,'' in {\em European
  Conference on Computer Vision (ECCV)}, 2016.

\bibitem{besl1992method}
P.~J. Besl and N.~D. McKay, ``Method for registration of 3-d shapes,'' in {\em
  Sensor fusion IV: control paradigms and data structures}, vol.~1611,
  pp.~586--606, International Society for Optics and Photonics, 1992.

\end{thebibliography}

\end{document}